\documentclass{article}

\usepackage[final]{neurips_2021}

\usepackage[utf8]{inputenc} %
\usepackage[T1]{fontenc}    %
\usepackage{hyperref}       %
\usepackage{url}            %
\usepackage{booktabs}       %
\usepackage{amsfonts}       %
\usepackage{nicefrac}       %
\usepackage{microtype}      %
\usepackage{xcolor}         %

\usepackage{algorithm}

\usepackage{caption}

\usepackage{microtype}
\usepackage{graphicx}
\usepackage{subfigure}
\usepackage{booktabs} %

\usepackage{amsmath}
\usepackage{amssymb}
\usepackage{bbm}
\usepackage{amsthm}
\usepackage{threeparttable}
\usepackage{array}
\usepackage{graphicx}
\usepackage{makecell}
\usepackage[capitalise]{cleveref}

\usepackage{xcolor}
\usepackage{csquotes}
\usepackage{comment}
\usepackage{wrapfig}
\usepackage{varwidth}
\usepackage{capt-of}
\usepackage{multicol}

\definecolor{ourspecialtextcolor}{rgb}{0.528, 0.471, 0.701} %

\usepackage{algorithm}
\usepackage[noend]{algpseudocode}
\DeclareCaptionSubType*{algorithm}

\algrenewcommand{\algorithmiccomment}[1]{\bgroup\hfill//~#1\egroup}
\algrenewcommand{\Return}{\State\textbf{return}\ }
\algnewcommand{\Save}{\State\textbf{save}\ }
\algnewcommand{\Load}{\State\textbf{load}\ }

\newtheorem{theorem}{Theorem}
\newtheorem{fact}{Fact}
\newtheorem{lemma}{Lemma}
\newtheorem{example}{Example}

\newtheorem{proposition}{Proposition}

\newtheorem{mirrortheorem}{Theorem}
\newtheorem{mirrorfact}{Fact}
\newtheorem{mirrorlemma}{Lemma}

\newtheorem{mirrorproposition}{Proposition}

\DeclareMathOperator{\fMAP}{\mathtt{MAP}}

\DeclareMathOperator{\GammaDist}{Gamma}
\DeclareMathOperator{\SoGDist}{SoG}
\DeclareMathOperator{\GumbelDist}{Gumbel}
\DeclareMathOperator{\ExpDist}{Exp}

\usepackage{bm}
\DeclareMathOperator*{\argmax}{arg\,max}
\DeclareMathOperator*{\argmin}{arg\,min}

\usepackage{mathtools}

\usepackage{multirow}

\usepackage{xspace}

\usepackage{placeins}

\newcommand{\RSet}{\mathbb{R}}

\newcommand{\bc}{\bm{c}}
\newcommand{\bz}{\bm{z}}
\newcommand{\bx}{\bm{x}}
\newcommand{\by}{\bm{y}}

\newcommand{\bu}{\bm{u}}
\newcommand{\bv}{\bm{v}}
\newcommand{\bomega}{\bm{\omega}}
\newcommand{\bepsilon}{\bm{\epsilon}}

\newcommand{\hbx}{\hat{\mathbf{z}}}

\newcommand{\ttheta}{\tilde{\theta}}

\newcommand{\bmu}{\bm{\mu}}
\newcommand{\btheta}{\bm{\theta}}

\newcommand{\diff}[2]{\partial_{#1}#2}
\newcommand{\grad}[2]{\nabla_{#1}#2}

\newcommand{\inputspace}{\mathcal{X}}
\newcommand{\outputspace}{\mathcal{Y}}
\newcommand{\latentspace}{\mathcal{Z}}
\newcommand{\distparamspace}{\Theta}
\newcommand{\latentprobdist}{p(\bz; \btheta)}

\newcommand{\mleobj}{\mathcal{L}}
\newcommand{\mlegradient}{\widehat{\nabla}_{\btheta}\mleobj}

\newcommand{\noisedist}{\rho(\epsilon)}
\newcommand{\bnoisedist}{\rho(\bepsilon)}

\newcommand{\dimz}{m}

\newcommand{\exx}{\hat{\bx}}
\newcommand{\exy}{\hat{\by}}
\newcommand{\exz}{\hat{\bz}}
\newcommand{\exc}{\hat{\bc}}
\newcommand{\extheta}{\hat{\btheta}}

\usepackage{xspace}
\newcommand*{\eg}{e.g.\@\xspace}
\newcommand*{\ie}{i.e.\@\xspace}

\newcommand{\std}[1]{ \pm #1}

\newcommand{\imle}{\textsc{I-MLE}\@\xspace}

\newcommand{\norm}[1]{\left\lVert#1\right\rVert}

\usepackage{etoolbox}

\newbool{proofsIMLE}

\newbool{printOld}

\newbool{printChecklist}

\newbool{printApdx}
\booltrue{printApdx}

\newbool{extraspace}

\begin{document}

\author{%
  Mathias Niepert \\
  NEC Laboratories Europe\\
  \texttt{mathias.niepert@neclab.eu} \\
  \And
  Pasquale Minervini \\
  University College London \\
  \texttt{p.minervini@ucl.ac.uk} \\
   \And
  Luca Franceschi \\
  Istituto Italiano di Tecnologia \\ 
  University College London \\
  \texttt{ucablfr@ucl.ac.uk} \\
}

\title{Implicit MLE: Backpropagating Through Discrete\\Exponential Family Distributions}

\maketitle

\vspace{-5mm}
\begin{abstract}
Combining discrete probability distributions and combinatorial optimization problems with neural network components has numerous applications but poses several challenges.
We propose Implicit Maximum Likelihood Estimation (\imle), a framework for end-to-end learning of models combining discrete exponential family distributions and differentiable neural components.
\imle is widely applicable as it only requires the ability to compute the most probable states and does not rely on smooth relaxations.
The framework encompasses several approaches such as perturbation-based implicit differentiation and recent methods to differentiate through black-box combinatorial solvers.
We introduce a novel class of noise distributions for approximating marginals via perturb-and-MAP.
Moreover, we show that \imle simplifies to maximum likelihood estimation when used in some recently studied learning settings that involve combinatorial solvers.
Experiments on several datasets suggest that \imle is competitive with and often outperforms existing approaches which rely on problem-specific relaxations.
\end{abstract}

\section{Introduction}

While deep neural networks excel at perceptual tasks, they tend to generalize poorly whenever the problem at hand requires %
some level of symbolic manipulation or reasoning, or exhibit some (known) algorithmic structure.
Logic, relations, and explanations, as well as decision processes, frequently find natural abstractions in discrete structures, ill-captured by the continuous mappings of standard neural nets.
Several application domains, ranging from relational and explainable ML to discrete decision-making~\citep{mivsic2020data}, could benefit from general-purpose learning algorithms whose inductive biases are more amenable to integrating symbolic and neural computation.
Motivated by these considerations, there is a growing interest in end-to-end learnable models incorporating discrete components that allow, e.g., to sample from discrete latent distributions~\citep{jang2016categorical,paulus2020gradient} or solve combinatorial optimization problems~\citep{poganvcic2019differentiation,Mandi_Guns:2020}. Discrete energy-based models (EBMs)~\citep{lecun2006tutorial} and discrete world models~\citep{hafner2020mastering} are additional examples of neural network-based models that require the ability to backpropagate through discrete probability distributions.  

For complex discrete distributions, it is intractable to compute the exact gradients of the expected loss.
For combinatorial optimization problems, the loss is discontinuous, and the gradients are zero 
almost everywhere.
The standard approach revolves around problem-specific smooth relaxations, which allow one to fall back to (stochastic) backpropagation.
These strategies, however, require tailor-made relaxations, presuppose access to the constraints and are, therefore, not always feasible nor tractable for large state spaces.
Moreover, reverting to discrete outputs at test time may cause unexpected behavior. In other situations, discrete outputs are required at training time because one has to make one of a number of discrete choices, such as accessing discrete memory or deciding on an action in a game.

With this paper, we take a step towards the vision of general-purpose algorithms for hybrid learning systems.
Specifically, we consider settings where the discrete component(s), embedded in a larger computational graph, are discrete random variables from the constrained exponential family\footnote{This includes integer linear programs via a natural link that we outline in \cref{ex:ILP}.}.
Grounded in concepts from Maximum Likelihood Estimation (MLE) and perturbation-based implicit differentiation, we propose Implicit Maximum Likelihood Estimation (\imle).
To approximate the gradients of the discrete distributions' parameters, \imle computes, at each update step, a \emph{target distribution} $q$ that depends on the loss incurred from the discrete output in the forward pass.
In the backward pass, we approximate maximum likelihood gradients by treating $q$ as the empirical distribution.
We propose ways to derive target distributions and introduce a novel family of noise perturbations well-suited for approximating marginals via perturb-and-MAP.
\imle is general-purpose as it only requires the ability to compute the most probable states and not faithful samples or probabilistic inference.
In summary, we make the following contributions:
\begin{enumerate}
    \item We propose implicit maximum likelihood estimation (\imle) as a  framework for computing gradients with respect to the parameters of discrete exponential family distributions;
    \item We show that this framework is useful  for backpropagating gradients through \emph{both} discrete probability distributions and discrete combinatorial optimization problems;
    \item \imle requires two ingredients: a family of target distribution $q$ and a method to sample from complex discrete distributions. We propose two
    families of target distributions and a family of noise-distributions for Gumbel-max (perturb-and-MAP) based sampling.  
    \item We show that \imle simplifies to \emph{explicit} maximum-likelihood learning when used in some recently studied learning settings involving combinatorial optimization solvers.
    \item Extensive experimental results suggest that \imle is flexible and competitive compared to the straight-through and relaxation-based estimators.
\end{enumerate}
Instances of the \imle framework can be easily integrated into modern deep learning pipelines, allowing one to readily utilize several types of discrete layers with minimal effort. We provide implementations and Python notebooks at  \url{https://github.com/nec-research/tf-imle}

\section{Problem Statement and Motivation}
\label{sec-ps}

\begin{wrapfigure}[8]{r}{0.48\textwidth}
\centering
\vspace{-6mm}
\includegraphics[clip,width=0.47\textwidth]{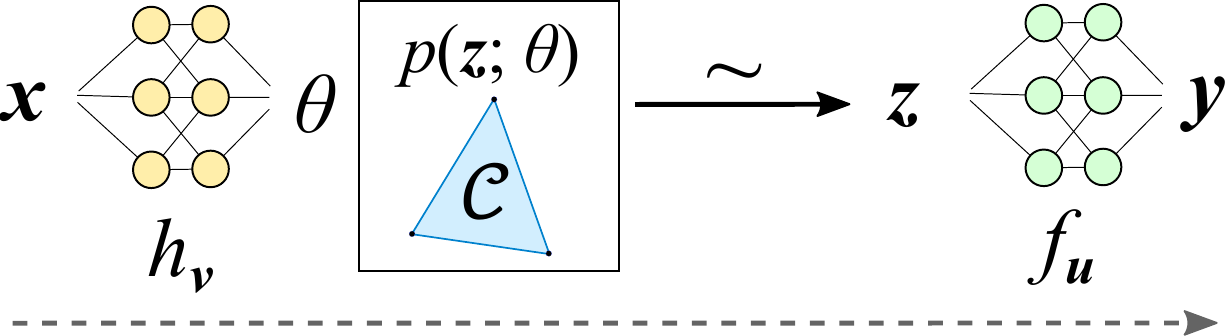}%
\caption{\small \label{setting-ilustration-main} Illustration of the addressed learning problem. $\bz$ is the discrete (latent) structure. 
}
\end{wrapfigure}

We consider models described by the equations
\begin{equation} \label{eq:hybridm}
 \btheta = h_{\bv}(\bx), \quad \bz\sim \latentprobdist, \quad  \by=f_{\bu}(\bz),
\end{equation}
where $\bx\in\inputspace$ and $\by\in\outputspace$ denote feature inputs and target outputs,
$h_{\bv}:\inputspace\to\distparamspace$ and $f_{\bu}:\latentspace\to\outputspace$ are smooth parameterized maps,
and $\latentprobdist$ is a discrete probability distribution.

Given a set of examples $\mathcal{D}=\{(\exx_j, \exy_j)\}_{j=1}^N$, we are concerned with learning the parameters $\bomega = (\bv, \bu)$ of \eqref{eq:hybridm} by finding approximate solutions of 
$\min_{\bomega}  \sum_j L(\exx_j, \exy_j; \bomega)/N$.
The training error $L$ is typically 
defined as:
\begin{equation} \label{eq:loss_cs}
L(\exx, \exy; \bomega) = \mathbb{E}_{\exz\sim p(\bz; \extheta)}\left[\ell(f_{\bu}(\exz), \exy)\right] \quad \text{with} \quad \extheta = h_{\bv}(\exx),
\end{equation}
where $\ell:\outputspace\times\outputspace\to\RSet^+$ is a point-wise loss function. \cref{setting-ilustration-main} illustrates the setting.
For example, an interesting instance 
of \eqref{eq:hybridm} and \eqref{eq:loss_cs} arises in  
\emph{learning to explain} user reviews~\citep{chen2018learning} where the task is to infer a target sentiment score (\eg w.r.t. the quality of a product) from a review while also providing a concise explanation of the predicted score by selecting a subset of exactly $k$ words (cf. \cref{ex-topk}). In \cref{sec-exps}, we present experiments precisely in this setting.
As anticipated in the introduction, we
restrict the discussion to instances in which 
$\latentprobdist$ belongs
to the (constrained) \emph{discrete exponential family}, which we now formally introduce.

Let $\bm{Z}$ be a vector of discrete random variables over a state space $\latentspace$ and let $\mathcal{C} \subseteq \latentspace$ be the set of states  that satisfy a given set of \emph{linear constraints}.\footnote{For the sake of simplicity we assume $\latentspace \subseteq \{0, 1\}^{\dimz}$. The set $\mathcal{C}$ is the integral polytope spanned by the given, problem-specific,
linear constraints.}
Let $\btheta \in \distparamspace \subseteq \mathbb{R}^m$ be a real-valued parameter vector.
The probability mass function (PMF) of a discrete constrained exponential family r.v. is:
\begin{equation} \label{def-constrained-exp-family}
p(\bz; \btheta) = \left\lbrace
\begin{array}{ll}
     \exp\left(\langle\bz,\btheta\rangle/\tau - A(\btheta) \right) & \text{if } \bz \in \mathcal{C}, \\
     0 & \text{otherwise.} %
\end{array}
\right.
\end{equation}
Here, $\langle\cdot, \cdot\rangle$ is the inner product and $\tau$ the temperature, which, if not mentioned otherwise, is assumed to be $1$.
$A(\btheta)$ is the log-partition function defined as $A(\btheta) = \log\left(\sum_{\bz \in \mathcal{C}} \exp \left(\langle\bz,\btheta\rangle/\tau  \right)\right)$.
We call $\langle \bz, \btheta\rangle$ the \emph{weight} of the state $\bz$.
The \emph{marginals} (expected value, mean) of 
the 
r.v.s 
$\mathbf{Z}$ are defined as 
$\bmu(\btheta) \coloneqq \mathop{\mathbb{E}}_{\exz \sim \latentprobdist}[\exz]$. 
Finally, the most probable or Maximum A-Posteriori (MAP) states are
defined as $\fMAP(\btheta) \coloneqq \argmax_{\bz \in \mathcal{C}} \;\langle\bz,\btheta\rangle$.
The family of probability distributions we define here captures a broad range of settings and subsumes probability distributions such as positive Markov random fields and statistical relational formalisms~\citep{wainwright2008graphical,raedt2016statistical}.
We now discuss some examples which we will use in the experiments. 
Crucially, in \cref{ex:ILP} we establish the link between the constrained exponential family and  
integer linear programming (ILP) 
identifying the ILP cost coefficients with the 
distribution's parameters $\btheta$.

\begin{example}[Categorical Variables] \label{ex-catvars}
An $m$-way (one-hot) categorical variable 
corresponds to $p(\bz; \btheta) = \exp\left( \langle \bz, \btheta\rangle - A(\btheta) \right)$, subject to the constraint $\langle \bz, \mathbf{1} \rangle = 1$, where $\mathbf{1}$ is a vector of ones.
\end{example}

As %
$\mathcal{C} = \{\mathbf{e}_i\}_{i=1}^m$, where $\mathbf{e}_i$ is the $i$-th vector of the canonical base, the parameters of the above distribution coincide with the weights, which are often called \emph{logits} in this context.
The marginals $\bmu$ coincide with the PMF and can be expressed through a closed-form smooth function of $\btheta$: the softmax. 
This facilitates a natural relaxation that involves using $\bmu(\btheta)$ in place of $\bz$~\citep{jang2016categorical}. 
The convenient properties 
of the categorical distribution, however, quickly disappear even for slightly more complex distributions, as the following example shows. 
\begin{example}[$k$-subset Selection]  \label{ex-topk}
Assume we want to sample binary $m$-dimensional vectors with $k$ ones.
This amounts to replacing the constraint in \cref{ex-catvars} by the constraint
$\langle \bz, \mathbf{1}\rangle = k$.

\end{example}
Here, a closed-form expression for the marginals does not exist:
sampling from this distribution requires computing 
the $\binom{m}{k}=O(m^k)$ weights (if $k\leq m/2$). Computing  
MAP states instead
takes time linear in $m$.

\begin{example}[%
Integer Linear Programs%
]
\label{ex:ILP}
Consider the combinatorial optimization problem given by the integer linear program $\argmin_{\bz \in \mathcal{C}} \langle \bz, \bc \rangle$, where $\mathcal{C}$ is an integral polytope and $\bc\in\mathbb{R}^m$ is a vector of  cost coefficients, and let $\bz^*(\bc)$ be the set of its solutions.
We can associate to the ILP the family (indexed by $\tau > 0$) of probability distributions  $\latentprobdist$ from  \eqref{def-constrained-exp-family},  with $\mathcal{C}$ the ILP polytope and $\btheta=-\bc$. 
Then, for every $\tau>0$, the solutions of the ILP correspond to the MAP states: $\fMAP(\btheta) =  \argmax_{\bz \in \mathcal{C}} \langle\bz, \btheta \rangle = \bz^*(\bc)$ and for $\tau\to 0$ one has that $\Pr(\mathbf{Z}\in \bz^*(\bc))\to 1$.
\end{example}

Many problems of practical interest can be expressed as ILPs, such as finding shortest paths, %
planning and scheduling problems, and inference in propositional logic.

\section{The Implicit Maximum Likelihood Estimator} \label{section-imle-framework}

In this section, we develop and motivate a family of general-purpose gradient estimators for \cref{eq:loss_cs} 
that respect the structure of $\mathcal{C}$
.~\footnote{
The derivations are adaptable to other types of losses defined over the outputs of \cref{eq:hybridm}. 
}
Let $(\exx, \exy)\in\mathcal{D}$ be a training example and $\exz \sim p(\bz; h_{\bv}(\exx))$. 
The gradient 
of $L$ w.r.t. $\bu$ is given by 
$
\grad{\bu}{L
(\exx, \exy; \bomega)} 
= \mathbb{E}_{\exz} [
    \diff{\bu}{f_{\bu}(\exz)^{\intercal}
    \grad{\by}{\ell(\by, \exy)}
    }
]
$ with $\by = f_{\bu}(\exz)$, 
which may be estimated by drawing one or more samples from $p$.
Regarding $\grad{\bv}{L}$, one has
\begin{equation}
\label{eq-chain-rule-for-w}
    \grad{\bv}{L
(\exx, \exy; \bomega)} = 
    \diff{\bv}{h_{\bv}(\exx)^\intercal}
    \nabla_{\btheta}L(\exx, \exy; \bomega),
\end{equation}
where the major challenge is to compute $\nabla_{\btheta}L$.
A 
standard approach is to employ the score function estimator (SFE) which typically suffers from high variance. Whenever a pathwise derivative estimator (PDE) is available it is usually the preferred choice \citep{schulman2015gradient}.
In our setting, however, the PDE is not readily applicable
since $\bz$ is discrete and, therefore, every
(exact) reparameterization path would be discontinuous. 
Various authors developed (biased) adaptations of the PDE for discrete r.v.s (see \cref{sec:rw}).
These involve either smooth approximations of $\latentprobdist$  
or approximations of the derivative of the reparameterization map. %
\emph{Our proposal departs from these two routes and instead involves the formulation 
of an implicit maximum likelihood estimation problem.}
In a nutshell, I-MLE is a (biased) estimator that replaces 
$\nabla_{\btheta}L$ in \cref{eq-chain-rule-for-w}  
with $\mlegradient$, where $\mleobj$ is an implicitly defined MLE objective and $\widehat{\nabla}$ is an estimator of the gradient.

We now focus on deriving the (implicit) MLE objective $\mleobj$. Let us assume we can, for any given $\exy$, construct an exponential family distribution $q(\bz; \btheta')$ that, ideally, is such that
\begin{equation}
\label{eq-inequiality-pq}
    \mathbb{E}_{\exz\sim q(\bz; \btheta')}\left[\ell(f_{\bu}(\exz), \exy)\right] \leq \mathbb{E}_{\exz\sim p(\bz; \btheta)}\left[\ell(f_{\bu}(\exz), \exy)\right].
\end{equation}
We will call $q$ the \emph{target distribution}. 
The idea is that, by making $p$ more similar to $q$ we can (iteratively) reduce the model loss $L(\exx, \exy; \bomega)$.
To this purpose, we define $\mleobj$ as the MLE objective\footnote{
We expand on this in \ifbool{printApdx}{\cref{apx-mle}}{the appendix} where we also review the classic MLE setup \cite[][Ch. 9]{murphy2012machine}. 
}
between the model distribution $p$ with parameters $\btheta$ and the target distribution
$q$ with parameters $\btheta'$:
\begin{equation}
    \mleobj(\btheta, \btheta') \coloneqq - \mathbb{E}_{\exz\sim q(\bz; \btheta')} [\log p(\exz; \btheta)]
= 
\mathbb{E}_{\exz\sim q(\bz; \btheta')} [
    A(\btheta) - \langle \exz, \btheta \rangle
]
\end{equation}  

Now, exploiting the fact that $\nabla_{\btheta} A(\btheta) = \bmu(\btheta)$, we can compute the gradient of $\mleobj$ as
\begin{equation} \label{eq-grad-mleobj}
    \grad{\btheta}{\mleobj(\btheta, \btheta')} =
    \bmu(\btheta) - \mathbb{E}_{\exz\sim q(\bz; \btheta')} [\exz] 
    =
    \bmu(\btheta) - \bmu(\btheta'), 
\end{equation}
that is
, the difference between the marginals of the current distribution $p$ and the marginals of the  target distribution $q$,
also equivalent to the gradient of the KL divergence between $p$ and $q$.

\algrenewcommand\algorithmicindent{1em}
\algrenewcommand{\algorithmiccomment}[1]{\bgroup\hskip1em\textcolor{ourspecialtextcolor}{//~\textsl{#1}}\egroup}

\begin{algorithm}[t!]
\begin{multicols}{2}
\begin{algorithmic}
       \Function{ForwardPass}{$\btheta$}
       \\ \Comment{Sample from the noise distribution $\rho(\bepsilon)$}
       \State $\bepsilon \sim \rho(\bepsilon)$
       \\ \Comment{Compute a MAP state of perturbed $\btheta$}
       \State $\exz = \fMAP(\btheta + \bepsilon)$
       
       \Save $\btheta$, $\bepsilon$, and $\exz$ for the backward pass
       \Return $\exz$
       \EndFunction
\end{algorithmic}
\begin{algorithmic}
       \Function{BackwardPass}{$\grad{\bz}{\ell}(f_{\bu}(\bz), \exy), \lambda$}
       \Load $\btheta$, $\bepsilon$, and $\exz$ from the forward pass
       
       \\ \Comment{Compute target distribution parameters}
       \State $\btheta' = \btheta - \lambda \grad{\bz}{\ell}(f_{\bu}(\bz), \exy)$
       
       \\ \Comment{Single sample \imle gradient estimate}
       \State $\mlegradient(\extheta, \extheta') = \exz - \fMAP(\btheta' + \bepsilon)$
       \Return $\mlegradient(\extheta, \extheta')$
       \EndFunction
\end{algorithmic}
\end{multicols}
\caption{Instance of \imle with perturbation-based implicit differentiation.}
\label{algo:main}
\vspace{-4mm}
\end{algorithm}

We will not use \cref{eq-grad-mleobj} directly, as computing the marginals is, in general, a \#P-hard problem and scales poorly with the dimensionality $m$.
MAP states are typically less expensive to compute (\eg see \cref{ex-topk}) and are often used directly to approximate $\bmu(\btheta)$\footnote{This is known as the \emph{perceptron learning rule} in standard MLE.}
or to compute 
perturb-and-MAP approximations, 
where $\bmu(\btheta)\approx \mathbb{E}_{\bepsilon\sim \bnoisedist}\fMAP(\btheta + \bepsilon)$ where $\bepsilon\sim\bnoisedist$ is an appropriate \emph{noise distribution} with domain $\mathbb{R}^{m}$. 
In this work we follow -- and explore in more detail in \cref{section-perturb-and-map} -- the latter approach (also referred to as the Gumbel-max trick \citep[cf.][]{Papandreou:2011}),
a strategy that
retains most of the computational  advantages of the pure MAP approximation but may be less crude. 
Henceforth, we only assume access to an algorithm to compute MAP states (such as a standard ILP
solver in the case of \cref{ex:ILP}) and rephrase
\cref{eq:hybridm} as 
\begin{equation} \label{eq:hybridm2}
 \btheta = h_{\bv}(\bx), \quad 
 \bz 
 = 
 \fMAP(\btheta + \bepsilon) 
 \; \text{with} \; \bepsilon \sim p(\bepsilon), \quad  \by=f_{\bu}(\bz). 
\end{equation}
With \cref{eq:hybridm2} in place,  
the general expression for the I-MLE estimator is $\widehat{\nabla}_{\bv} L(\bx, \by; \bomega)=    \diff{\bv}{h_{\bv}(\exx)^\intercal}
    \mlegradient(\btheta, \btheta')$ 
with $\btheta=h_{\bv}(\exx)$ where, for $S\in\mathbb{N}^+$:
\begin{equation}
\label{eqn-perturb-map-approx}
    \mlegradient(\btheta, \btheta') = 
    \frac{1}{S}\sum_{i=1}^S
    [
        \fMAP(\btheta + \bepsilon_i) -
        \fMAP(\btheta' + \bepsilon_i)
    ],  \;  \text{with} \;
    \bepsilon_i \sim \bnoisedist \; \text{for} \; i\in\{1, \dots, S\}.
\end{equation}
If the states of both the distributions $p$ and $q$ are binary vectors, %
$\mlegradient(\btheta, \btheta') \in [-1, 1]^{m}$ and when $S=1$ $\mlegradient(\btheta, \btheta') \in \{-1, 0, 1\}^{m}$.
In the following, we discuss the problem of constructing families of target distributions $q$. We will also analyze under what assumptions the inequality of \cref{eq-inequiality-pq} 
holds. 

\subsection{Target Distributions via Perturbation-based Implicit Differentiation}
\label{subsection-implicit-differentiation}

The efficacy of the I-MLE estimator hinges on a 
proper 
choice of $q$, a hyperparameter of our framework.  
In this section we derive and motivate a class of general-purpose target distributions, rooted in perturbation-based implicit differentiation (PID): 
\begin{equation}
    \label{eq-pid-q}
    q(\bz; \btheta') = p(\bz; \btheta - \lambda \grad{\bz}{\ell}(f_{\bu}(\overline{\bz}), \exy)) 
    \; \text{with} \; 
    \overline{\bz} = \fMAP(\btheta + \bepsilon) \; \text{and} \;
    \bepsilon\sim \bnoisedist,
\end{equation}
where $\btheta=h_{\bv}(\exx)$, $(\exx, \exy)\in\mathcal{D}$ is a data point, and $\lambda>0$ is a hyperparameter that controls the perturbation intensity.

To motivate \cref{eq-pid-q}, consider the setting where the inputs to $f$ are the marginals of $\latentprobdist$ (rather than discrete perturb-and-MAP samples as in \cref{eq:hybridm2}), that is, $\by = f_{\bu}(\bmu(\btheta))$ with $\btheta=h_{\bv}(\exx)$, and redefine the training error $L$ of \cref{eq:loss_cs} accordingly.
A seminal result by \citet{domke:2010} shows that, in this case, we can obtain  $\grad{\btheta}{L}$ by perturbation-based differentiation as:
\begin{equation}
\label{eqn-domke}
\grad{\btheta}{L
(\exx, \exy; \bomega)} = \lim_{\lambda \rightarrow 
0}
\left\lbrace\frac{1}{\lambda}
 \left[ \bmu(\btheta) - \bmu\left(\btheta - \lambda  \grad{\bmu}{
L(\exx, \exy; \bomega)
} \right)\right]\right\rbrace, 
\end{equation}
where 
$\grad{\bmu}{L}
=
\diff{\bmu}{f_{\bu}(\bmu)}^\intercal \grad{\by}{\ell(\by, \exy)}
$. 
The expression inside the limit may be interpreted as the gradient of an implicit MLE objective 
(see \cref{eq-grad-mleobj})  between the distribution $p$ with (current) parameters $\btheta$  and 
$p$ 
with parameters perturbed in the negative direction of the downstream gradient $\grad{\bmu}{L}$. 
Now, we can adapt \eqref{eqn-domke} to our setting of \cref{eq:hybridm2} by resorting to the straight-through estimator (STE) assumption \citep{bengio2013estimating}. 
Here, the STE assumption translates into reparameterizing $\bz$ as a function of $\bmu$ and approximating $\diff{\bmu}{\bz}\approx \bm{I}$. 
Then, $\grad{\bmu}{L}=\diff{\bmu}{\bz}^\intercal \grad{\bz}{L}\approx \grad{\bz}{L}$ and we approximate \cref{eqn-domke} as:
\begin{equation}
\label{eqn-approx-domke-stright-through}
\grad{\btheta}{L
(\exx, \exy; \bomega)} \approx 
\frac{1}{\lambda} 
\left[ \bmu(\btheta) - \bmu\left(\btheta - \lambda \grad{\bz}{L}(\exx, \exy; \bomega)\right)\right] 
 = \frac{1}{\lambda}\nabla_{\btheta} \mleobj(\btheta, \btheta - \lambda \grad{\bz}{L(\exx, \exy; \bomega)}),
\end{equation}
for some $\lambda>0$. 
From \cref{eqn-approx-domke-stright-through} we derive \eqref{eq-pid-q} by taking a single sample estimator of $\grad{\bz}{L}$
(with perturb-and-MAP sampling) 
and by incorporating the constant $1/\lambda$ into a global learning rate. 
I-MLE with PID target distributions may be seen as a way to generalize the STE to more complex distributions.
Instead of using the gradients $\grad{\bz}{L}$ to backpropagate directly, 
\imle uses them to construct a target distribution $q$.
With that, it defines an implicit maximum likelihood objective, whose gradient (estimator) propagates the supervisory signal upstream, 
critically, taking the constraints into account.
When using \cref{eq-pid-q} with $\bnoisedist=\delta_{0}(\bepsilon)$\footnote{$\delta_{0}$ is the Dirac delta centered around $0$ -- this is equivalent to approximating the marginals with $\fMAP$.}, 
the I-MLE estimator also recovers a recently proposed gradient estimation rule to differentiate through black-box combinatorial optimization problems~\citep{poganvcic2019differentiation}. \imle unifies existing gradient estimation rules in one framework. 
\cref{algo:main} shows the pseudo-code of the algorithm implementing \cref{eqn-perturb-map-approx} for $S=1$, using the PID target distribution of \cref{eq-pid-q}.
The simplicity of the code also  demonstrates that instances of  \imle can easily be implemented as a layer.

We will resume the discussion about target distributions in \cref{section-optimal-q-co}, where we analyze more closely the setup of \cref{ex:ILP}. Next, we focus on the perturb-and-MAP strategies and derive a class of noise distributions that is particularly apt to the settings we consider in this work.

\subsection{A Novel Family of Perturb-and-MAP Noise Distributions}
\label{section-perturb-and-map}

When $p$ is a complex high-dimensional distribution,  obtaining Monte Carlo estimates of the gradient in \cref{eq-grad-mleobj} requires approximate sampling.
In this paper, we rely on perturbation-based sampling, also known as perturb and MAP~\citep{Papandreou:2011}. 
In this Section we propose a novel way to design tailored noise perturbations. While the proposed family of noise distributions works with \textsc{I-MLE}, the results of this section are of independent interest and can also be used in other (relaxed) perturb-and-MAP based gradient estimators~\citep[e.g.][]{paulus2020gradient}. 
First, we start by revisiting a classic result by \citet{Papandreou:2011} which theoretically motivates the perturb-and-MAP approach (also known as the Gumbel-max trick), which we generalize here to consider also the temperature parameter $\tau$.

\begin{proposition}
\label{prob-full-potential}
Let $p(\bz; \btheta)$ be a discrete exponential family distribution with integer polytope $\mathcal{C}$ and temperature $\tau$, and let $\langle\bz,\btheta\rangle$ be the unnormalized weight of each $\bz \in \mathcal{C}$. Moreover, let $\tilde{\btheta}$ be such that, for all $\bz \in \mathcal{C}$, $\langle\bz,\tilde{\btheta}\rangle =  \langle\bz,\btheta\rangle + \epsilon(\bz)$ with each
$\epsilon(\bz)$ sampled i.i.d. from $\GumbelDist(0,\tau)$. Then we have that  $\Pr(\fMAP(\tilde{\btheta})=\bz) = p(\bz; \btheta).$
\end{proposition}

\ifbool{proofsIMLE}{
    \begin{proof}
    Let $\epsilon_i \sim \GumbelDist(0,\tau)$ i.i.d. and $\ttheta_i = \theta_i + \epsilon_i$. Following a derivation similar to one made in \citet{Papandreou:2011}, we have:
    \begin{align*}
        & \Pr\{\argmax(\ttheta_1, \ldots, \ttheta_m) = n\}  = \\
        = & \Pr\{ \ttheta_n \geq \max_{j\neq n} \{\ttheta_j\} \}  \\
        = & \int_{-\infty}^{+\infty} g(t; \theta_n) \prod_{j\neq n} G(t; \theta_j) \,dt \\
        = & \int_{-\infty}^{+\infty} \frac{1}{\tau} \exp\left(\frac{\theta_n - t}{\tau} - e^{\frac{\theta_n - t}{\tau}}\right) \prod_{j\neq n} \exp\left( -e^{\frac{\theta_j - t}{\tau}}\right) \,dt \\
        = &  \int_{-\infty}^{+\infty} \frac{1}{\tau} e^{\frac{\theta_n - t}{\tau}}\exp\left(-e^{\frac{\theta_n-t}{\tau}}\right) \prod_{j\neq n} \exp\left(-e^{\frac{\theta_j - t}{\tau}}\right) \,dt \\
        = & \int_{0}^{1} \prod_{j \neq n} z^{\exp\left( \frac{\theta_j - \theta_n}{\tau}  \right)} \,dz \qquad \text{with } z := \exp\left( -e^{\frac{\theta_n - t}{\tau}} \right) \\
        = & \frac{1}{1 + \sum_{j \neq n} e^{\frac{\theta_j - \theta_n}{\tau}} } \\
        = & \frac{e^{\frac{\theta_n}{\tau}}}{\sum_{j=1}^{m} e^{\frac{\theta_j}{\tau}}, }
      \end{align*}
      where $g$ and $G$ are respectively the Gumbel probability density function and the Gumbel cumulative density function.
      The proposition now follows from arguments made in \citet{Papandreou:2011} using the maximal equivalent re-parameterization of $p(\bz; \btheta)$ where we specify a parameter $\langle \btheta,\bz\rangle$ for each $\bz \in \mathcal{C}$ and perturb these parameters. 
    \end{proof}
}{
All proofs can be found in  
\ifbool{printApdx}{\cref{proofs-imle}.}{the appendix.}
}
The proposition states that if we can perturb the weights $\langle \bz,\btheta \rangle$ of each $\bz \in \mathcal{C}$ with independent $\GumbelDist(0, \tau)$ noise, then obtaining MAP states from the perturbed model is equivalent to sampling from $p(\bz; \btheta)$ at temperature\footnote{
Note that the temperature here is different to the temperature of the Gumbel softmax trick~\citep{jang2016categorical}
which scales \emph{both} the sum of the logits \emph{and} the samples from $\GumbelDist(0, 1)$.} 
 $\tau$.
For complex exponential distributions,  perturbing the weights $\langle\bz,\btheta\rangle$ for each state $\bz \in \mathcal{C}$ is at least as expensive as computing the marginals exactly. 
Hence, one usually resorts to 
\emph{local perturbations} of each
$[\btheta]_i$ (the $i$-th entry of the vector $\btheta$) with Gumbel noise. 
Fortunately, we can prove that,  for a large class of distributions, it is possible to design more suitable \emph{local} perturbations. 
First, we show that, for any $\kappa \in \mathbb{N}^+$, a Gumbel distribution can be written as a finite sum of $\kappa$ 
i.i.d. (implicitly defined) random variables.
\begin{lemma}
\label{lemma-gamma-gumbel}
Let $X \sim \GumbelDist(0 ,\tau)$ and let $\kappa \in \mathbb{N}^+$. 
Define the Sum-of-Gamma distribution as 
\begin{equation}
    \label{eqn-sum-of-gammas}
    \SoGDist(\kappa, \tau, s) \coloneqq \frac{\tau}{\kappa} 
    \left\lbrace\sum_{i=1}^{s} \left\{\GammaDist(1/\kappa, \kappa/i)\right\} - \log(s)\right\rbrace,
\end{equation}
where $s\in\mathbb{N}^+$ and $\GammaDist(\alpha, \beta)$ is the Gamma distribution with shape $\alpha$ and scale $\beta$, and let $\SoGDist(\kappa, \tau)\coloneqq\lim_{s \rightarrow \infty} \SoGDist(\kappa, \tau, s)$.
Then we have that $X\sim \sum_{j=1}^{\kappa} \bepsilon_j$, with $\bepsilon_j\sim \SoGDist(\kappa, \tau).$
\end{lemma}

\ifbool{proofsIMLE}{
    \begin{proof}
    Let $\kappa \in \mathbb{N}^+$ and let $X \sim \GumbelDist(0, \tau)$. Its moment generating function has the form 
    \begin{equation}
    \label{eq:m-gen-f-gumb}
    \mathbb{E}[\exp(tX)] = \Gamma(1 - \tau t).
    \end{equation}
    As mentioned in \citet[p. 443,][]{johnson1998advances} we know that we can write the Gamma function as 
    \begin{equation}
    \label{eq-gamma-book}
    \Gamma(1 - \tau t) = e^{\gamma \tau t} \prod_{i=1}^{\infty} \left(1 - \frac{\tau t}{i}\right)^{-1} e^{\frac{-\tau t}{i}}
    \end{equation}
    where $\gamma$ is the Euler-Mascheroni constant.
    We have that 
    $$\left(1 - \frac{\tau t}{i}\right)^{-1} = \frac{i}{i - \tau t} = \frac{\frac{i}{\tau}}{\frac{i - \tau t}{\tau}} =  \frac{\frac{i}{\tau}}{\frac{i}{\tau} - \frac{\tau t}{\tau}} = 
    \frac{\frac{i}{\tau}}{\frac{i}{\tau} - t}.$$ 
    The last term is the moment generating function of an exponential distribution with scale $\frac{\tau}{i}$. 
    We can now take the logarithm on both sides of \cref{eq-gamma-book} and obtain
    \begin{equation*}
      tX = \gamma\tau t + \lim_{s\rightarrow \infty}\sum_{i=1}^{s} \left( t \ExpDist(\tau/i) - \frac{\tau t}{i} \right),
    \end{equation*}
    where $\ExpDist(\alpha)$ is the exponential distribution with scale $\alpha$.
    Hence, 
    \begin{align*}
     X \sim & \lim_{s\rightarrow \infty}\sum_{i=1}^{s}  \left( \ExpDist(\tau/i) - \frac{\tau}{i} \right) +  \gamma\tau \\
      = & \lim_{s\rightarrow \infty}\sum_{i=1}^{s}  \left( \ExpDist(\tau/i) - \frac{\tau}{i} \right) + \tau \lim_{s\rightarrow \infty}\sum_{i=1}^{s} \frac{1}{i} - \log(s) \\
      = & \lim_{s\rightarrow \infty}\sum_{i=1}^{s}  \left( \ExpDist(\tau/i) - \frac{\tau}{i} + \frac{\tau}{i} \right) - \tau\log(s)   \\
        = & \lim_{s\rightarrow \infty}\sum_{i=1}^{s}  \ExpDist(\tau/i) - \tau\log(s)
    \end{align*}
     
    Since $\ExpDist(\alpha) \sim \GammaDist(1, \alpha)$, and due to the scaling and summation properties of the Gamma distribution (with shape-scale parameterization), we can write for all $r>1$:
    $$\ExpDist(\alpha) \sim \sum_{j=1}^{r} \GammaDist(1/r, \alpha r)/r.$$
    Hence, picking $r=\kappa$ from the hypothesis, we have 
    
    \begin{align*}
        X \sim & \lim_{s\rightarrow \infty} \left\lbrace 
            \sum_{i=1}^{s} \sum_{j=1}^{\kappa} \GammaDist(1/\kappa, \tau \kappa/i)/\kappa
        \right\rbrace - \tau \log(s) \\
        = & 
        \lim_{s\rightarrow \infty} \left\lbrace 
            \sum_{j=1}^{\kappa} \frac{\tau}{k}\sum_{i=1}^{s} \GammaDist(1/\kappa, \kappa/i)
        \right\rbrace - \sum_{j=1}^\kappa \frac{\tau}{\kappa} \log(s) \\
        = & 
        \lim_{s\rightarrow \infty} 
        \sum_{j=1}^{\kappa} \frac{\tau}{\kappa} 
        \left\lbrace
            \left[
                \sum_{i=1}^{s} \GammaDist(1/\kappa, \kappa/i)
            \right] - \log(s)
        \right\rbrace \\
        = & 
        \sum_{j=1}^{\kappa} 
        \frac{\tau}{\kappa}
        \left\lbrace
            \lim_{s\rightarrow \infty} 
            \left[
                \sum_{i=1}^{s} \GammaDist(1/\kappa, \kappa/i)
            \right] - \log(s)
        \right\rbrace 
    \end{align*}
    This concludes the proof.
    Parts of the proof are inspired by a post on stackexchange~\cite{214875}.
    \end{proof}
}

Based on \cref{lemma-gamma-gumbel}, we can show that for exponential family distributions where every $\bz \in \mathcal{C}$ has exactly $k$ non-zero entries
we can design perturbations of $\langle\bz,\btheta\rangle$ following a  Gumbel distribution.  

\begin{theorem}
\label{thm-perturb-distribution}
Let $p(\bz; \btheta)$ be a discrete exponential family distribution with integer polytope $\mathcal{C}$ and temperature $\tau$.
Assume that if $\bz \in \mathcal{C}$ then $\langle\bz, \mathbf{1}\rangle = k$ for some constant $k \in \mathbb{N}^+$.
Let $\tilde{\btheta}$ be the perturbation obtained by $[\tilde{\btheta}]_j = [\btheta]_j + \bepsilon_j$ with $\bepsilon_j \sim \SoGDist(k, \tau)$ from \cref{eqn-sum-of-gammas}.
Then, 
$\forall\bz\in\mathcal{C}$ we have that $\langle \bz, \tilde{\btheta}\rangle = \langle \bz, \btheta\rangle + \epsilon(\bz)$, with $\epsilon(\bz) \sim \GumbelDist(0, \tau)$.
\end{theorem}

\ifbool{proofsIMLE}{
    \begin{proof}
    
    Since we perturb each $\theta_i$ by $\varepsilon_i$ 
     we have, by assumption, that $\langle\btheta, \mathbf{1}\rangle = k$, for every $\bz \in \mathcal{C}$, that
    \begin{equation}
        \langle \bz, \tilde{\btheta} \rangle = \langle  \bz, \btheta \rangle + \sum_{j=1}^k \varepsilon_j.  %
    \end{equation}
    Since by \cref{lemma-gamma-gumbel} we know that $\sum^{k}_{j=1} \varepsilon_i \sim \GumbelDist(0, \tau)$, the statement of the theorem follows.
    \end{proof}
}

\begin{wrapfigure}[10]{r}{0.489\textwidth}
\centering
\vspace{-4.0mm}
\subfigure{\includegraphics[scale=0.24]{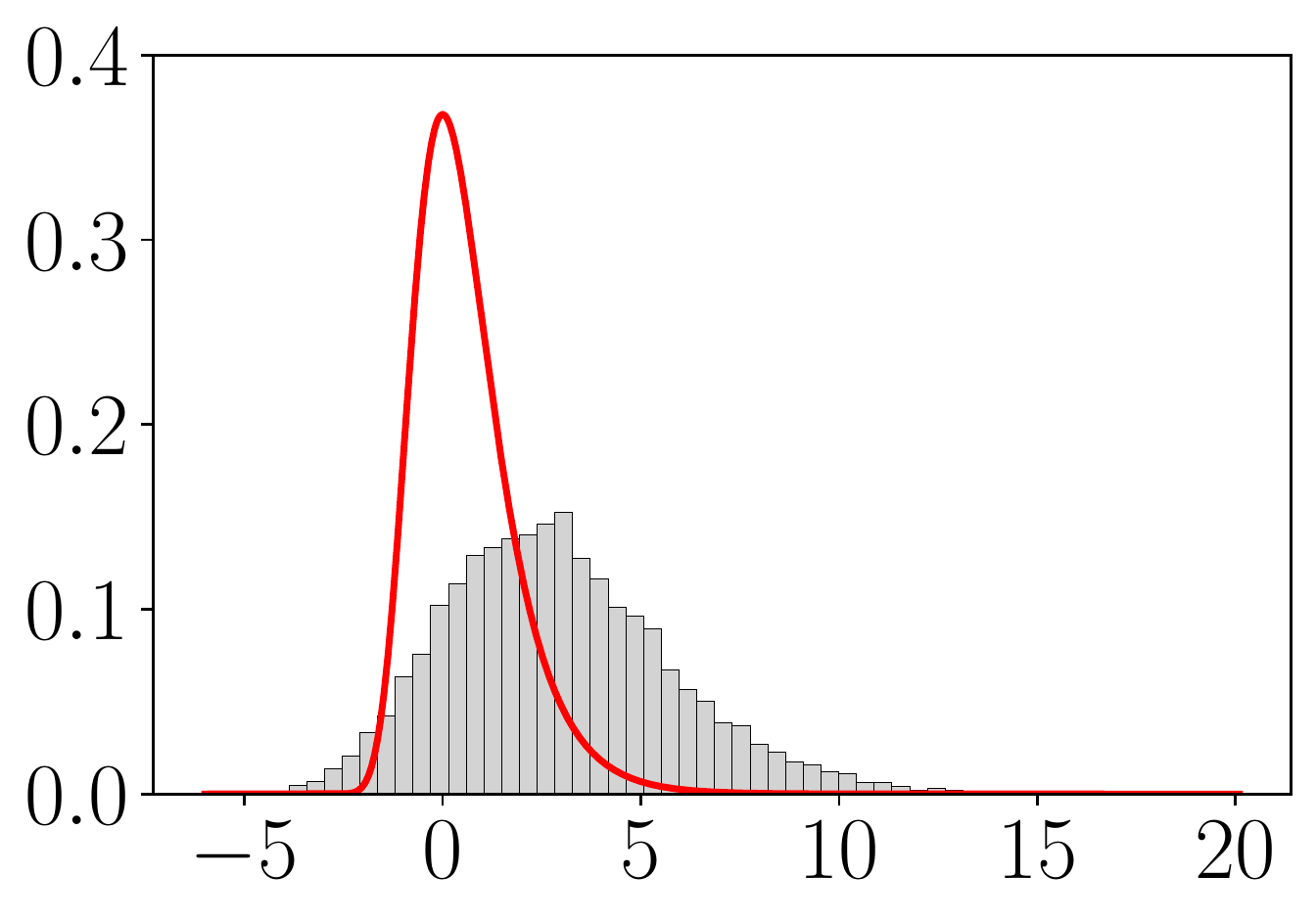}}
\subfigure{\includegraphics[scale=0.24]{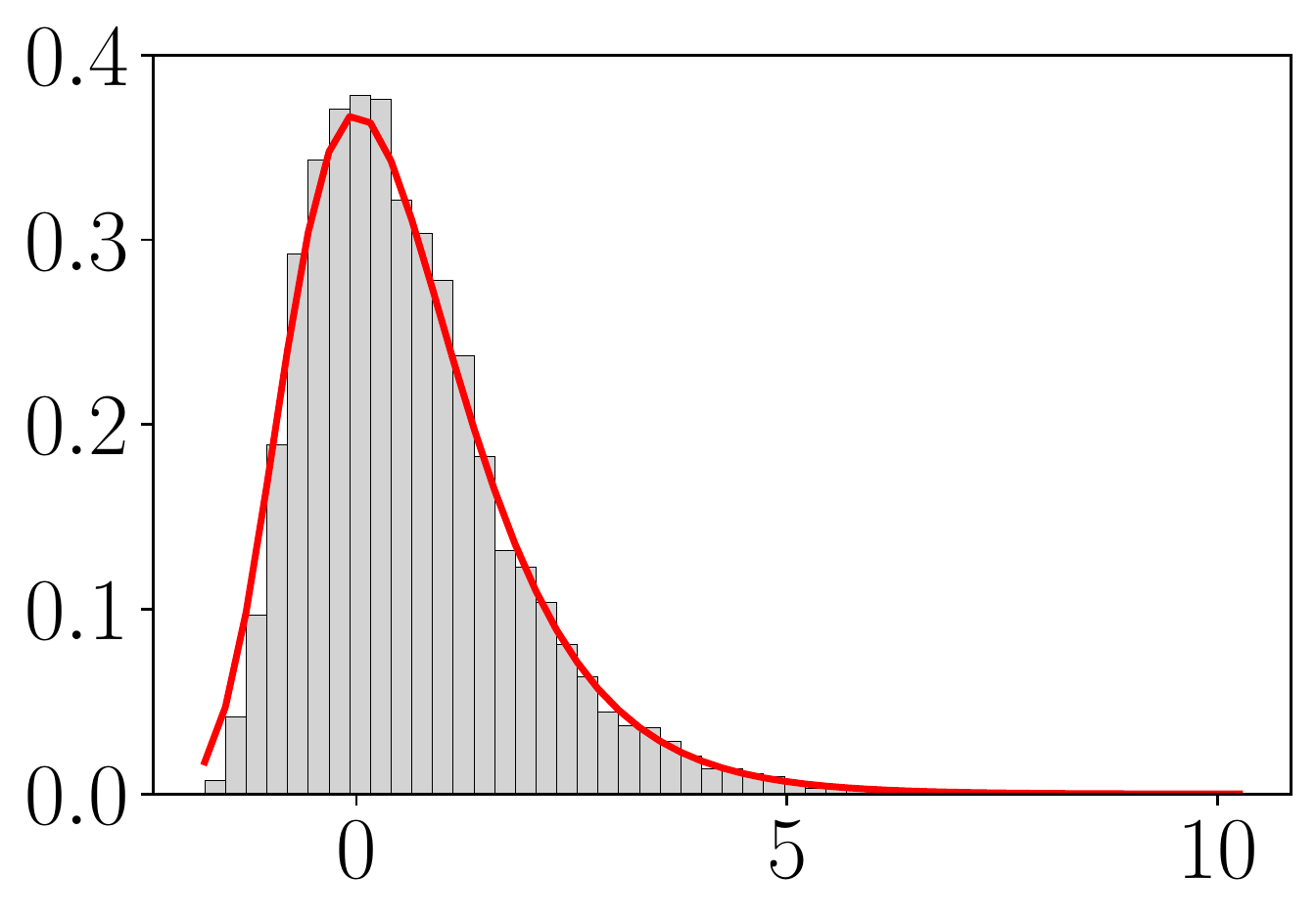}}
\vspace{-4mm}
\caption{\small \label{figure-dist-plots} Histograms for $10$k samples where each sample is (left) the sum of $5$ 
$\epsilon_j\sim\GumbelDist(0, 1)$ or (right) the sum of $5$ $\epsilon_j\sim\SoGDist(5, 1, 10)$.
}
\end{wrapfigure}
Many problems such as $k$-subset selection, traveling salesman, spanning tree, and graph matching strictly satisfy the assumption of \cref{thm-perturb-distribution}.
We can, however, also apply the strategy in cases where the variance of 
$\langle \mathbf{Z}, \mathbf{1} \rangle$
is small (\eg shortest weighted path).
The Sum-of-Gamma perturbations provide a more fine-grained approach to  noise perturbations. For $\tau=\kappa=1$, we obtain the standard Gumbel perturbations. 
In contrast to the standard $\GumbelDist(0, 1)$ noise, the proposed local Sum-of-Gamma 
perturbations result in 
weights' perturbations 
that follow the Gumbel distribution. \cref{figure-dist-plots} shows histograms of $10$k samples, where each sample is either the sum of $5$ samples from $\GumbelDist(0, 1)$ (the standard approach) or the sum of $k=5$ samples from $\SoGDist(5, 1, 10) = \frac{1}{5}\sum_{i=1}^{10} \{\GammaDist(1/5, 5/i) - \log(10)\}$.
While we still cannot sample faithfully from $p(\bz; \btheta)$ as the perturbations are \emph{not} independent, we can counteract the problem of partially dependent perturbations by increasing the temperature $\tau$ and, therefore, the variance of 
the noise distribution.
We explore and verify the importance of tuning $\tau$ empirically.
In the appendix, we also show that the infinite series from \cref{lemma-gamma-gumbel} can be well approximated by a finite sum using convergence results for the Euler-Mascheroni series~\citep{MORTICI2010}.

\section{Target Distributions for Combinatorial Optimization Problems}
\label{section-optimal-q-co}

In this section, we explore the setting where the discrete computational component arises from a combinatorial optimization (CO) problem, specifically an integer linear program (ILP).
Many authors have recently  considered
the setup where the CO component occupies the last layer of the model defined by \cref{eq:hybridm} (where $f_{\bu}$ is the identity) and the supervision is available  in terms of examples of either optimal solutions~\citep[e.g.][]{poganvcic2019differentiation} or optimal cost coefficients (conditioned on the inputs)~\citep[e.g.][]{elmachtoub2020smart}.
We have seen in \cref{ex:ILP} that we can naturally associate to each ILP a probability distribution  (see \cref{def-constrained-exp-family}) with $\btheta$ given by the negative  cost coefficients $\bc$ of the ILP and $\mathcal{C}$ the integral polytope. Letting $\tau \rightarrow 0$ is equivalent to taking the MAP in the forward pass. Furthermore, in \cref{subsection-implicit-differentiation} we showed that the I-MLE framework subsumes a recently propose method by  \citet{poganvcic2019differentiation}. 
Here, instead, we show that, for a certain choice of the target distribution, \imle estimates the gradient of an explicit maximum likelihood learning 
loss $\mleobj$ where the data distribution is ascribed to either (examples of) optimal solutions or optimal cost coefficients.

Let $q(\bz; \btheta')$ be the distribution $p(\bz; \btheta')$, with parameters
\begin{equation}
\label{eqn-target-q-optimal}
[{\btheta'}]_i :=\left\lbrace
\begin{array}{ll}
     [\btheta]_i  & \mbox{ if } [\grad{\bz}{L}]_i = 0  \\
     - [\grad{\bz}{L}]_i & \mbox{ otherwise}. 
\end{array}
\right.
\end{equation}
In the first CO setting, we observe training data $\mathcal{D}=\{(\exx_j, \exy_j)\}_{j=1}^N$ where $\exy_j\in\mathcal{C}$ and the loss $\ell$ measures a distance between a discrete  $\exz_j \sim p(\bz; \btheta_j)$ with $\btheta_j = h_{v}(\exx)$ and 
a given optimal solution of the ILP 
$\exy_j$. 
An example is the Hamming loss $\ell_H$~\citep{poganvcic2019differentiation} defined as $\ell_H(\bz, \by) = \bz \circ (\mathbf{1} - \by) + \by \circ (\mathbf{1} - \bz)$, where $\circ$ denotes the Hadamard (or entry-wise) product. 

\begin{fact}
\label{thm-optimal-target-distribution-y}
If one uses
$\ell_H$, then \imle with the target distribution 
of \cref{eqn-target-q-optimal}
and $\bnoisedist=\delta_0$
is equivalent to 
the perceptron-rule estimator of the MLE objective between $p(\bz; h_{\bv}(\exx_j))$ and $\exy_j$.
\end{fact}

\ifbool{proofsIMLE}{
\begin{proof}
Rewriting the definition of the Hamming loss gives us 
$$\ell_H(\bz, \by) = \frac{1}{m}\sum_{i=1}^{m} \left( \bz_i + \by_i - 2 \bz_i\by_i \right).$$
Hence, we have that $$\grad{\bz_i}{\ell_H} = \frac{1}{m}\left(1 - 2\by_i\right).$$ 
Therefore, $\grad{\bz_i}{\ell_H} = -\frac{1}{m}$ if $\by_i = 1$ and $\grad{\bz_i}{H} = \frac{1}{m}$ if $\by_i = 0$.   
Since, by definition $\by \in \mathcal{C}$, we have that
$$\fMAP\left(-\grad{\bz}{\ell_H}\right) = \by.$$ 
Now, when using \imle with $S=1$ and $\bnoisedist=\delta_0$ we approximate the gradients as  
\[\mlegradient(\btheta, \btheta') = \fMAP(\btheta) - \fMAP(\btheta') = \fMAP(\btheta) - \fMAP\left(-\grad{\bz}{\ell_H}\right) = \fMAP(\btheta)  - \by.\]
This concludes the proof. 
\end{proof}
}
It follows that the method by \citet{poganvcic2019differentiation} returns, for a large enough $\lambda$, the maximum-likelihood gradients  (scaled by $1/\lambda$) approximated by the perceptron rule. %
The proofs are given in \ifbool{printApdx}{\cref{proofs-imle}.}{the Appendix.}

In the second CO setting, we observe training data $\mathcal{D}=\{(\exx_j, \hat{\bc}_j)\}_{j=1}^N$, where $\hat{\bc}_j$ is the optimal cost conditioned on input $\exx_j$. Here, 
various authors \citep[e.g.][]{elmachtoub2020smart,Mandi_Guns:2020,DBLP:conf/nips/MandiG20} use as point-wise loss 
the regret $\ell_R(\btheta, \bc) = \bc^{\top} \left(\bz(\btheta) - \exz^{*}(\bc)\right)$ where $\bz(\btheta)$ is a state  sampled from $p(\bz; \btheta)$ (possibly with temperature $\tau\to0$, that is, a MAP state) and $\exz^{*}(\bc)\in \bz^*(\bc)$ is an optimal state for $\bc$.

\begin{fact}
\label{thm-optimal-target-distribution-c}
If one uses $\ell_R$ 
then \imle with the target distribution 
of \cref{eqn-target-q-optimal} is equivalent to the perturb-and-MAP estimator of the MLE objective between $p(\bz; h_{\bv}(\exx_j))$ and $p(\bz; - \exc_j)$. 
\end{fact}

\ifbool{proofsIMLE}{
\begin{proof}
We have that $\grad{\bz_i}{\ell_R} = \bc_i$ for all $i$.
Now, when using \imle with target distribution $\hat{q}(\bz; \btheta')$ of \cref{eqn-target-q-optimal} (and without loss of generality, for $S=1$) we have that $\hat{q}(\bz; \btheta') = p(\bz; -\bc)$, and we approximate the gradients as  
\[ \mlegradient(\btheta, \btheta') =  \fMAP(\btheta + \bepsilon_i) - \fMAP(\btheta' + \bepsilon_i) = \fMAP(\btheta + \bepsilon_i) - \fMAP(-\bc + \bepsilon_i),  \;  \text{where} \;
    \bepsilon_i \sim \bnoisedist. \]
Hence, \imle approximates the gradients of the maximum likelihood estimation problem between $p(\bz; h_{\bv}(\exx_j))$ and $p(\bz; - \exc_j)$ using perturb-and-MAP. This concludes the proof. 
\end{proof}
}

This last result also implies that when using the target distribution $q$ from \eqref{eqn-target-q-optimal} in conjunction with the regret, \imle performs maximum-likelihood learning minimizing the KL divergence between the current distribution and the distribution whose parameters are the optimal cost.

Moreover, both facts imply that, when sampling from the MAP states of the distribution $q$ defined by \cref{eqn-target-q-optimal}, 
we have that $\ell(\exz, \exy) = 0$ for $\exz \in \mathtt{MAP}(\btheta')$. 
Therefore, $\ell(\exz, \exy) = 0 \leq \mathbb{E}_{\exz\sim p(\bz; \btheta)}\left[\ell((\exz, \exy)\right]$, meaning that the inequality of \cref{eq-inequiality-pq} is satisfied for $\tau\to 0$.

\section{Related Work} 
\label{sec:rw}

Several papers address the gradient estimation problem for discrete r.v.s, many resorting to relaxations.
\citet{maddison2016concrete, jang2016categorical} propose the Gumbel-softmax distribution to relax categorical r.v.s;
\citet{paulus2020gradient} study extensions to more complex probability distributions.
The concrete distribution (the Gumbel-softmax distribution) is only directly applicable to categorical variables. For more complex distributions, one has to come up with tailor-made relaxations or use the straight-through or score function estimators (see for instance \cite{kim2016exact,grover2019stochastic}).
In our experiments, we compare with the Gumbel-softmax estimator in Figure 4 (left and right). We show that the $k$-subset VAE trained with \imle achieves loss values that are similar to those of the categorical ($1$-subset) VAE trained with the Gumbel-softmax gradient estimator. 
\citet{tucker2017rebar, grathwohl2017backpropagation} develop parameterized control variates (the former was named REBAR) based on continuous relaxations for the score-function estimator. %
In contrast, we focus explicitly on problems where \emph{only} discrete samples are used during training. Moreover, REBAR is tailored to categorical distributions. \imle is intended for models with complex distributions (e.g. those with with many constraints). %

Approaches that do not rely on relaxations are specific to certain distributions~\citep{bengio2013estimating, franceschi2019learning, liu2019rao} or assume knowledge of $\mathcal{C}$~\citep{kool2020estimating}.  %
We provide a general-purpose framework that does not require access to the linear constraints and the corresponding integer polytope $\mathcal{C}$.  
Experiments in the next section show that while \imle only requires a MAP solver, it is competitive and sometimes outperforms tailor-made relaxations. 
SparseMAP~\citep{Niculae2018SparseMAPDS} is an approach to structured prediction and latent variables, replacing the exponential distribution (specifically, the softmax) with a sparser distribution. Similar to our work, it only presupposes the availability of a MAP oracle.
LP-SparseMAP~\citep{Niculae2020LPSparseMAPDR} is an extension that uses a relaxation of the optimization problem rather than a MAP solver. Sparsity can also be exploited for efficient marginal inference in latent variable models~\citep{correia2020efficient}.

A series of works 
about differentiating through CO problems 
\citep{wilder2019melding, elmachtoub2020smart, ferber2020mipaal, DBLP:conf/nips/MandiG20} relax ILPs by adding $L^1$, $L^2$ or log-barrier regularization terms and differentiate through the KKT conditions deriving from the application of the cutting plane or the interior-point methods. 
These approaches are conceptually linked to techniques for differentiating through smooth programs \citep{amos2017optnet, donti2017task, agrawal2019differentiable, chen2020understanding, domke2012generic, franceschi2018bilevel} that arise not only in modelling but also in hyperparameter optimization and meta-learning.  \citet{poganvcic2019differentiation, rolinek2020deep, berthet2020learning} propose methods that are not tied to a specific ILP solver. 
As we saw 
above,
the former two, originally derived from a continuous interpolation argument, may be interpreted as special instantiations of \imle. The latter addresses the theory of perturbed optimizers and discusses perturb and MAP in the context of the Fenchel-Young loss. 
All the CO-related works assume that either optimal costs or solutions are given as training data, while \imle may be also applied in the absence of such supervision by making use of implicitly generated target distributions. 
Other authors focus on devising differentiable relaxations for specific CO problems such as SAT~\citep{evans2018learning} or MaxSAT~\citep{wang2019satnet}. 
Machine learning intersects with CO also in other contexts, \eg in learning heuristics to improve the performances of CO solvers or  differentiable models such as GNNs to \textquote{replace} them; see \citet{bengio2020machine} and references therein.

Direct Loss Minimization~\citep[DLM,][]{mcallester2010direct, song2016training} is also related to our work, but the assumption there is that examples of optimal states $\exz$ are given. \citet{lorberbom:2019} extend the DLM framework to discrete VAEs using coupled perturbations. 
Their approach is tailored to VAEs and not general-purpose.
Under a methodological viewpoint, 
\imle inherits from classical MLE~\citep{wainwright2008graphical} and perturb-and-MAP~\citep{Papandreou:2011}.
The theory of perturb-and-MAP was used to derive general-purpose upper bounds for log-partition functions~\citep{hazan2012partition,shpakova:2016}.

\section{Experiments}
\label{sec-exps}

The set of experiments can be divided into three parts. First, we analyze and compare the behavior of \imle with (i) the score function and (ii) the straight-through estimator using a toy problem. Second, we explore the latent variable setting where both $h_{\bv}$ and $f_{\bu}$ in \cref{eq:hybridm} are neural networks and the optimal structure is \emph{not} available during training. %
Finally, we address the problem of differentiating through black-box combinatorial optimization problems, where we use the target distribution derived in \cref{section-optimal-q-co}. More experimental details for available in the  appendix.

\paragraph{Synthetic Experiments.}

We conducted a series of experiments with a tractable $5$-subset distribution (see \cref{ex-topk}) where $\mathbf{z}\in\{0,1\}^{10}$.
We set the loss to $L(\bm{\theta}) = \mathbb{E}_{\exz \sim p(\mathbf{z}; \bm{\theta})} [ \norm{\exz - \mathbf{b}}^2 ]$, where $\mathbf{b}$ is a fixed vector sampled from $\mathcal{N}(0,\mathbf{I})$.
In \cref{fig:toy_experiments} (Top), we plot optimization curves with means and standard deviations, comparing the proposed estimator with the straight-through (STE) and the score function (SFE) estimators.~\footnote{Hyperparameters are optimized against $L$ for all methods independently. Statistics are over 100 runs. We found STE slightly better with Gumbel rather than SoG noise. SFE failed with all tested PaM strategies.}
For STE and I-MLE, we use Perturb-and-MAP (PaM) with Gumbel and $\SoGDist(1, 5, 10)$ noise, respectively.
The SFE uses faithful samples and exact marginals (which is feasible only when $m$ is very small) and converges much more slowly than the other methods, while the STE converges to worse solutions than those found using \imle.
\cref{fig:toy_experiments} (Bottom) shows the benefits of using SoG rather than Gumbel perturbations with \imle. 
While the best configurations for both are comparable, SoG noise achieves in average (over 100 runs) strictly better final values of $L$ for more than $50\%$ of the tested  configurations (varying $\lambda$ from \cref{eq-pid-q} and the learning rate) and exhibit smaller variance (see \ifbool{printApdx}{\cref{fig:toy_exps_apx_1}}{Appendix}). Additional details and  results in 
\ifbool{printApdx}{\cref{ax-toy-topk}.}{the Appendix.}

\begin{table}[t!]
  \begin{minipage}[b]{0.37\linewidth}
  \centering
     \includegraphics[width=0.78\columnwidth]{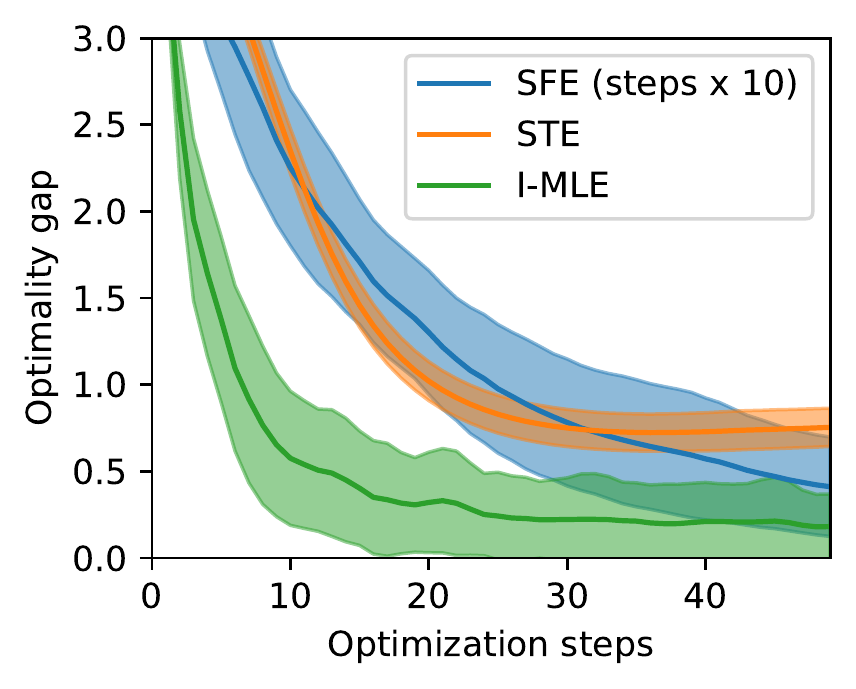}
    \\
    \includegraphics[width=0.9\columnwidth]{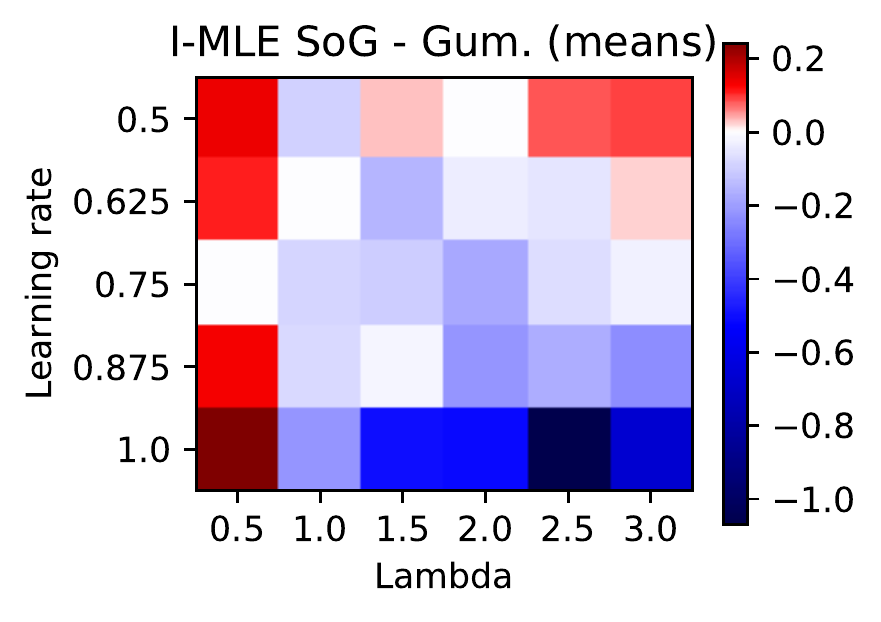}
    \captionof{figure}{\small \label{fig:toy_experiments}Top:
    Gradient-based optimization of $L$ with various
    estimators. 
    Bottom: Mean difference of the final value of $L$ between I-MLE with SoG or Gumbel $\noisedist$, varying $\lambda$ and the learning rate (blue $=$ better SoG). 
    } 
  \end{minipage}
  \hspace{2mm}
  \begin{varwidth}[b]{0.57\linewidth}
  \small
    \begin{tabular}{rrrrr}
    \toprule
    \multicolumn{1}{c}{\multirow{3}{*}{\bf Method}} & \multicolumn{2}{c}{\bf Test MSE}  & \multicolumn{2}{c}{\bf Subset Precision}  \\
    \cmidrule(lr){2-3} \cmidrule(lr){4-5}
    & \multicolumn{1}{c}{\bf Mean} & \multicolumn{1}{c}{\bf Std. Dev.} &  {\bf Mean} &  {\bf Std. Dev.} \\
    \midrule
    \multicolumn{5}{c}{$k=10$} \\
    \midrule
    L2X ($t=0.1$) & 6.68 & 1.08 & 26.65 & 9.39 \\
    SoftSub ($t=0.5$) & \textbf{2.67} & 0.14 & 44.44 & 2.27 \\
    STE  ($\tau=30$) & 4.44 & 0.09 & 38.93 & 0.14 \\
    \imle $\fMAP$ & 4.08 & 0.91 & 14.55 & 0.04 \\
    \imle $\GumbelDist$ & \textbf{2.68} & 0.10 & 39.28 & 2.62 \\
    \imle ($\tau=30$) & \textbf{2.71} & 0.10 & \textbf{47.98} & 2.26 \\
    \midrule
    \multicolumn{5}{c}{$k=5$} \\
    \midrule
    L2X ($t=0.1$) & 5.75 & 0.30 & 33.63 & 6.91 \\
    SoftSub ($t=0.5$) & \textbf{2.57} & 0.12 & \textbf{54.06} & 6.29 \\
    \imle ($\tau=5$) & \textbf{2.62} & 0.05 & \textbf{54.76} & 2.50 \\
    \midrule
    \multicolumn{5}{c}{$k=15$} \\
    \midrule
    L2X ($t=0.1$) & 7.71 & 0.64 & 23.49 & 10.93 \\
    SoftSub ($t=0.5$) & \textbf{2.52} & 0.07 & 37.78 & 1.71 \\
    \imle ($\tau=30$) & 2.91 & 0.18 & \textbf{39.56} & 2.07 \\
    \bottomrule
    \end{tabular}
    \vspace{4mm}
    \caption{\small \label{tab:l2x-aroma}Detailed results for the aspect \textsc{Aroma}. Test MSE and subset precision, both $\times 100$, for $k \in \{5, 10, 15\}$.}
  \end{varwidth}%
\vspace{-7mm}
\end{table}

\paragraph{Learning to Explain.}
\label{section-experiments-l2x}

The \textsc{BeerAdvocate} dataset \citep{McAuley2012LearningAA} consists of free-text reviews and ratings for $4$ different aspects of beer: appearance, aroma, palate, and taste. Each sentence in the test set has annotations providing the words that best describe the various aspects. Following the experimental setup of recent work~\citep{paulus2020gradient}, we address the problem introduced by the L2X paper~\citep{chen2018learning} of learning a distribution over $k$-subsets of words that best explain a given aspect rating. The complexity of the MAP problem  for the $k$-subset distribution is linear in $k$.  s%
The training set has 80k reviews for the aspect \textsc{Appearance} and 70k reviews for all other aspects.  Since the original dataset~\citep{McAuley2012LearningAA} did not provide separate validation and test sets, we compute $10$ different evenly sized validation/test splits of the 10k held out set and compute mean and standard deviation over $10$ models, each trained on one split. %
Subset precision was computed using a subset of 993 annotated reviews.
We use pre-trained word embeddings from~\cite{Lei2016RationalizingNP}.
Prior work used non-standard neural networks for which an implementation is not available~\citep{paulus2020gradient}.
Instead, we used the neural network from the L2X paper with $4$ convolutional and one dense layer. %
This neural network outputs the parameters $\btheta$ of the distribution $p(\bz; \btheta)$ over $k$-hot binary latent masks with $k \in \{5, 10, 15\}$. 
We compare to relaxation-based baselines L2X~\citep{chen2018learning} and SoftSub~\citep{Xie2019ReparameterizableSS}.
We also compare the straight-through estimator (STE) with Sum-of-Gamma (SoG) perturbations.
We used the standard hyperparameter settings of \citet{chen2018learning} and choose the temperature parameter $t \in \{0.1, 0.5, 1.0, 2.0\}$. For \imle we choose $\lambda \in \{10^1, 10^2, 10^3\}$, while for both \imle and STE we choose $\tau \in \{k, 2k, 3k\}$ based on the validation MSE.
We used the standard Adam settings.
We trained separate models for each aspect using MSE as point-wise loss $\ell$.

\cref{tab:l2x-aroma} lists detailed results for the aspect \textsc{Aroma}.
\imle's MSE values are competitive with those of the best baseline, and its subset precision is significantly higher than all other methods (for $\tau=30$).
Using only MAP as the approximation of the marginals leads to poor results. This shows that using the tailored perturbations with tuned temperature is crucial to achieve state of the art results. The Sum-of-Gamma perturbation introduced in this paper outperforms the standard local Gumbel perturbations. %
More details and results can be found in the appendix.

\paragraph{Discrete Variational Auto-Encoder.}
\label{section-experiment-vae}

\begin{figure*}[t!]%
\centering
\subfigure{%
\label{fig:r1}%
\includegraphics[width=0.332\textwidth]{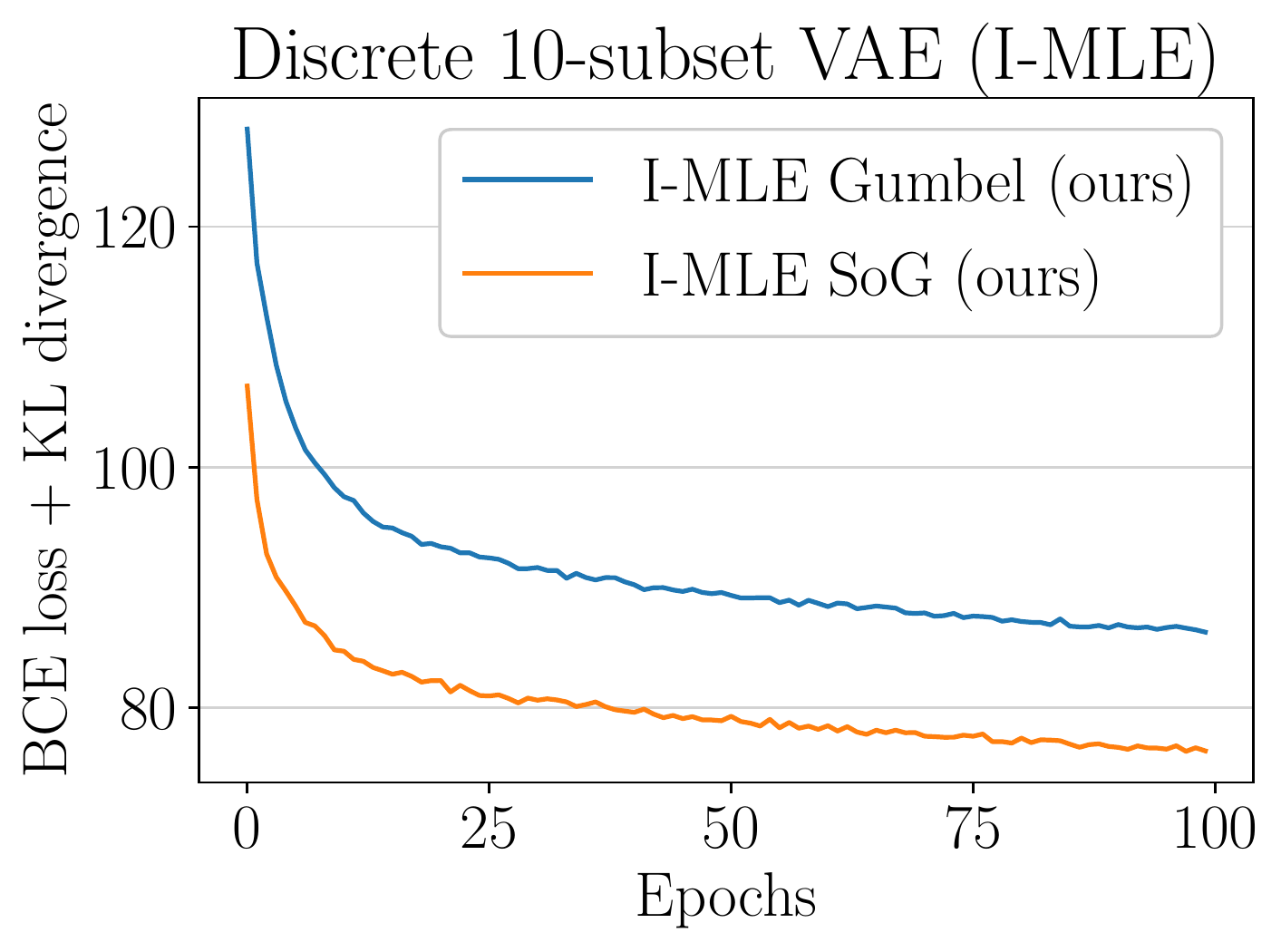}}%
\subfigure{%
\label{fig:r10}%
\includegraphics[width=0.332\textwidth]{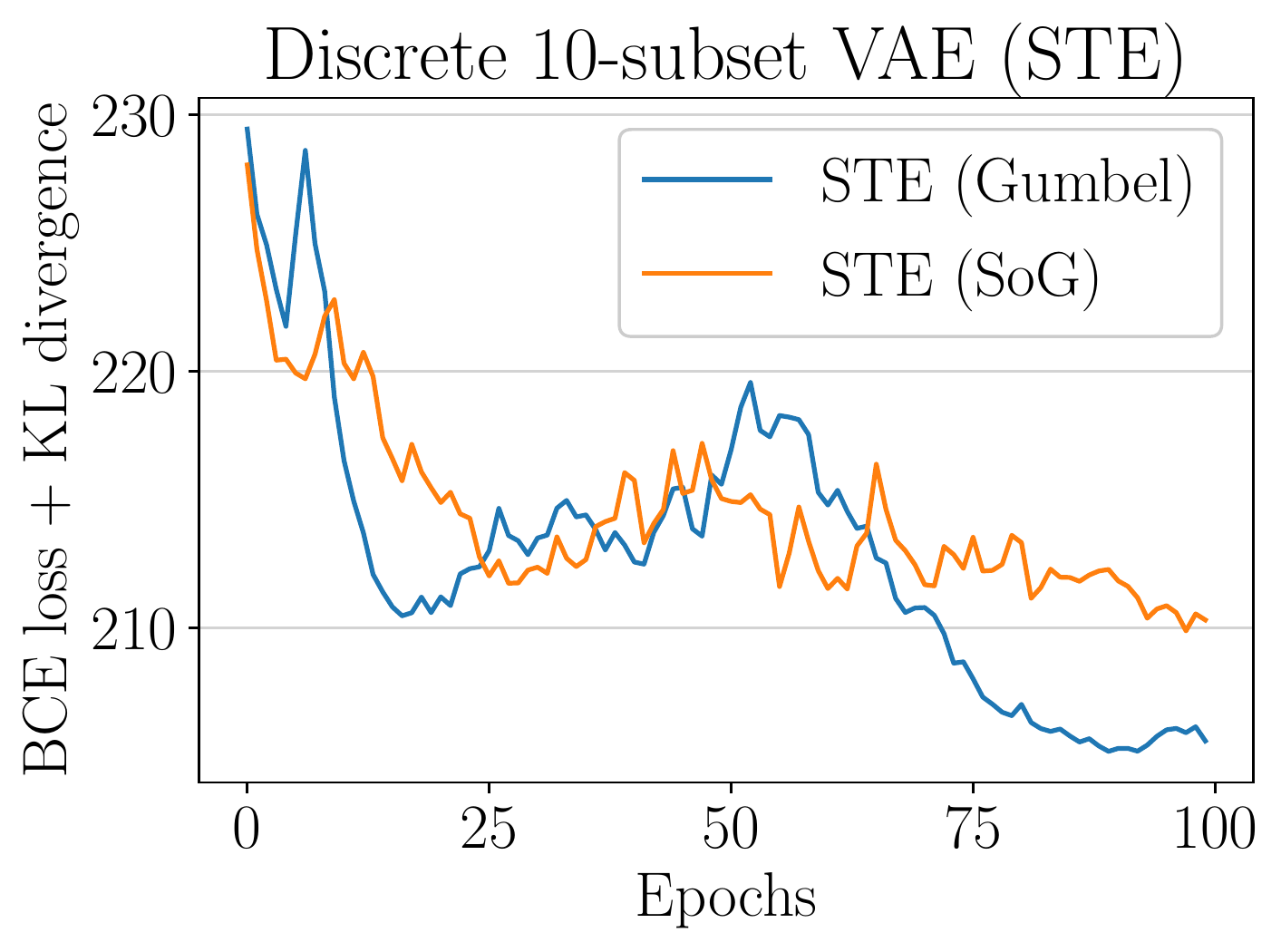}}%
\subfigure{%
\label{fig:softmax}%
\includegraphics[width=0.332\textwidth]{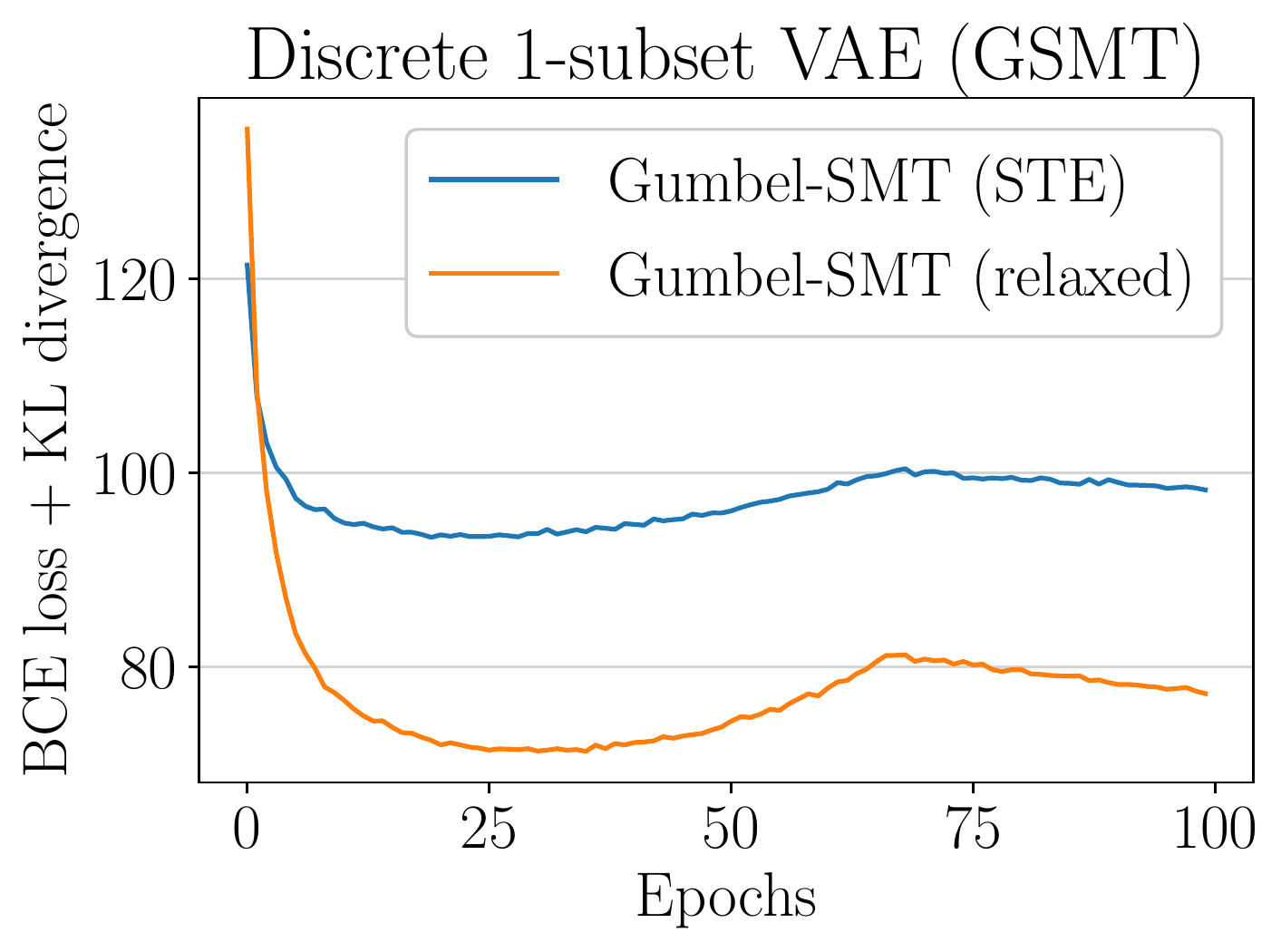}}%
\caption{\label{fig-autoencoder-plots} \small Plots of the sum of the binary reconstruction loss and the KL divergence as a function of the number of epochs (lower is better). (Left) Discrete $10$-subset VAE trained with \imle with $\lambda=10$ (\imle). (Center) Discrete $10$-subset VAE trained with the straight-through estimator (STE). (Right) Discrete $1$-subset VAE using the Gumbel softmax trick (GSMT). The down-up-down artifact is due to temperature annealing. Sum-of-Gamma (SoG) perturbations have the lowest test loss for the $10$-subset VAEs. For $\lambda=10$ and SoG perturbations, the test loss is similar to that of the categorical ($1$-subset) VAE trained with the Gumbel softmax trick. %
}
\vspace{-5mm}
\end{figure*}

We evaluate various perturbation strategies for a discrete $k$-subset Variational Auto-Encoder (VAE) and compare them to the straight-through estimator (STE) and the Gumbel-softmax trick. The latent variables model a probability distribution over $k$-subsets of (or top-$k$ assignments too) binary vectors of length $20$.  The special case of $k=1$ is equivalent to a categorical variable with $20$  categories. For $k>1$, we use \imle using the class of PID target distributions of \cref{eq-pid-q} and compare various perturb-and-MAP noise sampling strategies.
The experimental setup is similar to those used in prior work on the Gumbel softmax tricks~\citep{jang2016categorical}. The loss is the sum of the reconstruction losses (binary cross-entropy loss on output pixels) and the KL divergence between the marginals of the variables and the uniform distribution. The encoding and decoding functions of the VAE consist of three dense layers (encoding: 512-256-20x20; decoding: 256-512-784). We do not use temperature annealing. Using \cref{eqn-perturb-map-approx} with $S=1$, we use either $\GumbelDist(0, 1)$ perturbations (the standard approach)\footnote{Increasing the temperature $\tau$ of $\GumbelDist(0, \tau)$ samples increased the test loss.} or Sum-of-Gamma (SoG) perturbations at a temperature of $\tau=10$. We run $100$ epochs and record the loss on the test data. The difference in training time is negligible. 
\cref{fig-autoencoder-plots} shows that using the SoG noise distribution is beneficial. %
The test loss using the SoG perturbations is lower despite the perturbations having higher variance and, therefore, samples of the model being more diverse. 
This shows that using perturbations of the weights that follow a
proper Gumbel distribution is indeed beneficial. 
\imle significantly outperforms the STE, which does not work in this setting and is competitive with the Gumbel-Softmax trick for the $1$-subset (categorical) distribution where marginals can be computed in closed form. 

\paragraph{Differentiating through Combinatorial Solvers.} \label{section-experiment-blackbox}

\begin{wraptable}[10]{r}{0.64\textwidth}
\vspace{-4.4mm}
\small
	\caption{\small Results for the Warcraft shortest path task. Reported is the accuracy, \ie percentage of paths with the optimal costs. Standard deviations are over five runs.} \label{tab:sp-results}
	\centering
	\begin{tabular}{cccccc}
	\toprule
	$K$ & {\bf \textsc{I-Mle} ($\bmu$-$\bmu$)} & {\bf \textsc{I-Mle} ($\mathtt{M}$-$\mathtt{M}$)} & {\bf BB} & {\bf DPO} \\ %
	\midrule
	12 & ${\bf 97.2}\std{0.5}$ & $95.2\std{0.3}$ & $95.2\std{0.7}$ & $94.8\std{0.3}$ \\
	18 & ${\bf 95.8}\std{0.7}$ & $94.4\std{0.5}$ & $94.7\std{0.4}$ & $92.3\std{0.8}$ \\
	24 & ${\bf 94.3}\std{1.0}$ & $93.2\std{0.2}$ & $93.8\std{0.3}$ & $91.5\std{0.4}$ \\
	30 & ${93.6}\std{0.4}$ & ${\bf 93.7}\std{0.6}$ & $93.6\std{0.5}$ & $91.5\std{0.8}$ \\
	\bottomrule
	\end{tabular}
\end{wraptable}
In these experiments, proposed by \citet{poganvcic2019differentiation}, the training datasets consists of 10,000 examples of randomly generated images of terrain maps from the Warcraft II tile set~\citep{warcraft_map_editor}.
Each example has an underlying $K \times K$ grid whose cells represent terrains with a fixed cost. %
The shortest (minimum cost) path between the top-left and bottom-right cell in the grid is encoded as an indicator matrix and serves as the target output.
An image of the terrain map is presented to a CNN, which produces a $K \times K$ matrix of vertex costs.
These costs are then given to \emph{Dijkstra's algorithm} (the MAP solver) to compute the  shortest path.
We closely follow the evaluation protocol of \citet{poganvcic2019differentiation}. 
We considered two instantiations of \imle: one  derived from \cref{thm-optimal-target-distribution-y} ($\mathtt{M}$-$\mathtt{M}$ in \cref{tab:sp-results}) using $\ell_H$ and one derived from \cref{thm-optimal-target-distribution-c}  ($\bmu$-$\bmu$) using $\ell_R$, with $\noisedist=\SoGDist(k, 1, 10)$ where $k$ is the empirical mean of the path lengths (different for each grid size $K$).
We compare with the method proposed by \citet{poganvcic2019differentiation} (BB\footnote{
Note that this is the same as using \imle with PID target distribution form \cref{eq-pid-q} and $\noisedist=\delta_0$.}) and \citet{berthet2020learning} (DPO).
The results are listed in \cref{tab:sp-results}.
\imle obtains results comparable to (BB) with  $\mathtt{M}$-$\mathtt{M}$
and is more accurate with  $\bmu$-$\bmu$.
We believe that the $\bmu$-$\bmu$ advantage may be partially due to an implicit form of data augmentation since we know from \cref{thm-optimal-target-distribution-c} that, by using \imle, we obtain samples from the distribution whose parameters are the optimal cost.
Training dynamics, showing faster convergence of \imle ($\bmu$-$\bmu$), and additional details are available in \ifbool{printApdx}{\cref{app:warcraft}}{the Appendix}.

\section{Conclusions}

\imle is an efficient, simple-to-implement, and general-purpose framework for learning  
hybrid 
models. %
\imle is competitive with relaxation-based approaches for discrete latent-variable models and with approaches to backpropagate through CO solvers. Moreover, we showed empirically that \imle outperforms the straight-through estimator. A limitation of the work is its dependency on computing MAP states which is, in general, an NP-hard problem (although for many interesting cases there are efficient algorithms). 
Future work includes devising target distributions when $\grad{\bz}{L}$ is not available, studying 
the properties (including the bias) of the proposed estimator, developing adaptive strategies for $\tau$ and $\lambda$, and integrating and testing \imle in several challenging application domains.

\bibliography{ge}

\begin{thebibliography}{54}
\providecommand{\natexlab}[1]{#1}
\providecommand{\url}[1]{\texttt{#1}}
\expandafter\ifx\csname urlstyle\endcsname\relax
  \providecommand{\doi}[1]{doi: #1}\else
  \providecommand{\doi}{doi: \begingroup \urlstyle{rm}\Url}\fi

\bibitem[Agrawal et~al.(2019)Agrawal, Amos, Barratt, Boyd, Diamond, and
  Kolter]{agrawal2019differentiable}
A.~Agrawal, B.~Amos, S.~Barratt, S.~Boyd, S.~Diamond, and Z.~Kolter.
\newblock Differentiable convex optimization layers.
\newblock \emph{arXiv preprint arXiv:1910.12430}, 2019.

\bibitem[Amos and Kolter(2017)]{amos2017optnet}
B.~Amos and J.~Z. Kolter.
\newblock Optnet: Differentiable optimization as a layer in neural networks.
\newblock In \emph{International Conference on Machine Learning}, pages
  136--145. PMLR, 2017.

\bibitem[Bengio et~al.(2013)Bengio, L{\'e}onard, and
  Courville]{bengio2013estimating}
Y.~Bengio, N.~L{\'e}onard, and A.~Courville.
\newblock Estimating or propagating gradients through stochastic neurons for
  conditional computation.
\newblock \emph{arXiv preprint arXiv:1308.3432}, 2013.

\bibitem[Bengio et~al.(2020)Bengio, Lodi, and Prouvost]{bengio2020machine}
Y.~Bengio, A.~Lodi, and A.~Prouvost.
\newblock Machine learning for combinatorial optimization: a methodological
  tour d’horizon.
\newblock \emph{European Journal of Operational Research}, 2020.

\bibitem[Berthet et~al.(2020)Berthet, Blondel, Teboul, Cuturi, Vert, and
  Bach]{berthet2020learning}
Q.~Berthet, M.~Blondel, O.~Teboul, M.~Cuturi, J.~Vert, and F.~R. Bach.
\newblock Learning with differentiable pertubed optimizers.
\newblock In \emph{NeurIPS}, 2020.

\bibitem[Chen et~al.(2018)Chen, Song, Wainwright, and Jordan]{chen2018learning}
J.~Chen, L.~Song, M.~Wainwright, and M.~Jordan.
\newblock Learning to explain: An information-theoretic perspective on model
  interpretation.
\newblock In \emph{International Conference on Machine Learning}, pages
  883--892. PMLR, 2018.

\bibitem[Chen et~al.(2020)Chen, Zhang, Reisinger, and
  Song]{chen2020understanding}
X.~Chen, Y.~Zhang, C.~Reisinger, and L.~Song.
\newblock Understanding deep architecture with reasoning layer.
\newblock \emph{Advances in Neural Information Processing Systems}, 33, 2020.

\bibitem[Correia et~al.(2020)Correia, Niculae, Aziz, and
  Martins]{correia2020efficient}
G.~M. Correia, V.~Niculae, W.~Aziz, and A.~F. Martins.
\newblock Efficient marginalization of discrete and structured latent variables
  via sparsity.
\newblock \emph{Advances in Neural Information Processing Systems}, 2020.

\bibitem[Domke(2010)]{domke:2010}
J.~Domke.
\newblock Implicit differentiation by perturbation.
\newblock In \emph{Advances in Neural Information Processing Systems 23}, pages
  523--531. 2010.

\bibitem[Domke(2012)]{domke2012generic}
J.~Domke.
\newblock Generic methods for optimization-based modeling.
\newblock In \emph{Artificial Intelligence and Statistics}, pages 318--326.
  PMLR, 2012.

\bibitem[Donti et~al.(2017)Donti, Amos, and Kolter]{donti2017task}
P.~L. Donti, B.~Amos, and J.~Z. Kolter.
\newblock Task-based end-to-end model learning in stochastic optimization.
\newblock \emph{Advances in Neural Information Processing Systems}, 2017.

\bibitem[Elmachtoub and Grigas(2020)]{elmachtoub2020smart}
A.~N. Elmachtoub and P.~Grigas.
\newblock Smart ``predict, then optimize", 2020.

\bibitem[Evans and Grefenstette(2018)]{evans2018learning}
R.~Evans and E.~Grefenstette.
\newblock Learning explanatory rules from noisy data.
\newblock \emph{Journal of Artificial Intelligence Research}, 61:\penalty0
  1--64, 2018.

\bibitem[Ferber et~al.(2020)Ferber, Wilder, Dilkina, and
  Tambe]{ferber2020mipaal}
A.~Ferber, B.~Wilder, B.~Dilkina, and M.~Tambe.
\newblock Mipaal: Mixed integer program as a layer.
\newblock In \emph{Proceedings of the AAAI Conference on Artificial
  Intelligence}, volume~34, pages 1504--1511, 2020.

\bibitem[Franceschi et~al.(2018)Franceschi, Frasconi, Salzo, Grazzi, and
  Pontil]{franceschi2018bilevel}
L.~Franceschi, P.~Frasconi, S.~Salzo, R.~Grazzi, and M.~Pontil.
\newblock Bilevel programming for hyperparameter optimization and
  meta-learning.
\newblock In \emph{International Conference on Machine Learning}, pages
  1568--1577. PMLR, 2018.

\bibitem[Franceschi et~al.(2019)Franceschi, Niepert, Pontil, and
  He]{franceschi2019learning}
L.~Franceschi, M.~Niepert, M.~Pontil, and X.~He.
\newblock Learning discrete structures for graph neural networks.
\newblock In \emph{International conference on machine learning}, pages
  1972--1982. PMLR, 2019.

\bibitem[Grathwohl et~al.(2018)Grathwohl, Choi, Wu, Roeder, and
  Duvenaud]{grathwohl2017backpropagation}
W.~Grathwohl, D.~Choi, Y.~Wu, G.~Roeder, and D.~Duvenaud.
\newblock Backpropagation through the void: Optimizing control variates for
  black-box gradient estimation.
\newblock \emph{ICLR}, 2018.

\bibitem[Grover et~al.(2019)Grover, Wang, Zweig, and
  Ermon]{grover2019stochastic}
A.~Grover, E.~Wang, A.~Zweig, and S.~Ermon.
\newblock Stochastic optimization of sorting networks via continuous
  relaxations.
\newblock \emph{arXiv preprint arXiv:1903.08850}, 2019.

\bibitem[Guyomarch(2017)]{warcraft_map_editor}
J.~Guyomarch.
\newblock {Warcraft II Open-Source Map Editor}.
\newblock \url{http://github.com/war2/war2edit}, 2017.

\bibitem[Hafner et~al.(2020)Hafner, Lillicrap, Norouzi, and
  Ba]{hafner2020mastering}
D.~Hafner, T.~Lillicrap, M.~Norouzi, and J.~Ba.
\newblock Mastering atari with discrete world models.
\newblock \emph{arXiv preprint arXiv:2010.02193}, 2020.

\bibitem[Hazan and Jaakkola(2012)]{hazan2012partition}
T.~Hazan and T.~Jaakkola.
\newblock On the partition function and random maximum a-posteriori
  perturbations.
\newblock In \emph{Proceedings of the 29th International Coference on
  International Conference on Machine Learning}, pages 1667--1674, 2012.

\bibitem[He et~al.(2016)He, Zhang, Ren, and Sun]{DBLP:conf/cvpr/HeZRS16}
K.~He, X.~Zhang, S.~Ren, and J.~Sun.
\newblock Deep residual learning for image recognition.
\newblock In \emph{{CVPR}}, pages 770--778. {IEEE} Computer Society, 2016.

\bibitem[Jang et~al.(2017)Jang, Gu, and Poole]{jang2016categorical}
E.~Jang, S.~Gu, and B.~Poole.
\newblock Categorical reparameterization with gumbel-softmax.
\newblock \emph{ICLR}, 2017.

\bibitem[Johnson and Balakrishnan(1998)]{johnson1998advances}
N.~L. Johnson and N.~Balakrishnan.
\newblock Advances in the theory and practice of statistics, 1998.

\bibitem[Kim et~al.(2016)Kim, Sabharwal, and Ermon]{kim2016exact}
C.~Kim, A.~Sabharwal, and S.~Ermon.
\newblock Exact sampling with integer linear programs and random perturbations.
\newblock In \emph{Thirtieth AAAI Conference on Artificial Intelligence}, 2016.

\bibitem[Kool et~al.(2020)Kool, van Hoof, and Welling]{kool2020estimating}
W.~Kool, H.~van Hoof, and M.~Welling.
\newblock Estimating gradients for discrete random variables by sampling
  without replacement.
\newblock \emph{ICLR}, 2020.

\bibitem[LeCun et~al.(2006)LeCun, Chopra, Hadsell, Ranzato, and
  Huang]{lecun2006tutorial}
Y.~LeCun, S.~Chopra, R.~Hadsell, M.~Ranzato, and F.~Huang.
\newblock A tutorial on energy-based learning.
\newblock \emph{Predicting structured data}, 1\penalty0 (0), 2006.

\bibitem[Lei et~al.(2016)Lei, Barzilay, and Jaakkola]{Lei2016RationalizingNP}
T.~Lei, R.~Barzilay, and T.~Jaakkola.
\newblock Rationalizing neural predictions.
\newblock In \emph{EMNLP}, 2016.

\bibitem[Liu et~al.(2019)Liu, Regier, Tripuraneni, Jordan, and
  Mcauliffe]{liu2019rao}
R.~Liu, J.~Regier, N.~Tripuraneni, M.~Jordan, and J.~Mcauliffe.
\newblock Rao-blackwellized stochastic gradients for discrete distributions.
\newblock In \emph{International Conference on Machine Learning}, pages
  4023--4031. PMLR, 2019.

\bibitem[Lorberbom et~al.(2019)Lorberbom, Gane, Jaakkola, and
  Hazan]{lorberbom:2019}
G.~Lorberbom, A.~Gane, T.~Jaakkola, and T.~Hazan.
\newblock Direct optimization through argmax for discrete variational
  auto-encoder.
\newblock In \emph{Advances in Neural Information Processing Systems}, pages
  6203--6214, 2019.

\bibitem[Maddison et~al.(2017)Maddison, Mnih, and Teh]{maddison2016concrete}
C.~J. Maddison, A.~Mnih, and Y.~W. Teh.
\newblock The concrete distribution: A continuous relaxation of discrete random
  variables.
\newblock \emph{ICLR}, 2017.

\bibitem[Mandi and Guns(2020)]{DBLP:conf/nips/MandiG20}
J.~Mandi and T.~Guns.
\newblock Interior point solving for lp-based prediction+optimisation.
\newblock In \emph{Advances in Neural Information Processing Systems}, 2020.

\bibitem[Mandi et~al.(2020)Mandi, Demirovi{\'c}, Stuckey, and
  Guns]{Mandi_Guns:2020}
J.~Mandi, E.~Demirovi{\'c}, P.~J. Stuckey, and T.~Guns.
\newblock Smart predict-and-optimize for hard combinatorial optimization
  problems.
\newblock In \emph{Proceedings of the AAAI Conference on Artificial
  Intelligence}, volume~34, pages 1603--1610, 2020.

\bibitem[McAllester et~al.(2010)McAllester, Hazan, and
  Keshet]{mcallester2010direct}
D.~A. McAllester, T.~Hazan, and J.~Keshet.
\newblock Direct loss minimization for structured prediction.
\newblock In \emph{Advances in Neural Information Processing Systems},
  volume~1, page~3, 2010.

\bibitem[McAuley et~al.(2012)McAuley, Leskovec, and
  Jurafsky]{McAuley2012LearningAA}
J.~McAuley, J.~Leskovec, and D.~Jurafsky.
\newblock Learning attitudes and attributes from multi-aspect reviews.
\newblock \emph{2012 IEEE 12th International Conference on Data Mining}, pages
  1020--1025, 2012.

\bibitem[Mi{\v{s}}i{\'c} and Perakis(2020)]{mivsic2020data}
V.~V. Mi{\v{s}}i{\'c} and G.~Perakis.
\newblock Data analytics in operations management: A review.
\newblock \emph{Manufacturing \& Service Operations Management}, 22\penalty0
  (1):\penalty0 158--169, 2020.

\bibitem[Mortici(2010)]{MORTICI2010}
C.~Mortici.
\newblock {Fast convergences towards Euler-Mascheroni constant}.
\newblock \emph{{Computational \& Applied Mathematics}}, 29, 00 2010.

\bibitem[Murphy(2012)]{murphy2012machine}
K.~P. Murphy.
\newblock \emph{Machine learning: a probabilistic perspective}.
\newblock MIT press, 2012.

\bibitem[Niculae and Martins(2020)]{Niculae2020LPSparseMAPDR}
V.~Niculae and A.~F.~T. Martins.
\newblock Lp-sparsemap: Differentiable relaxed optimization for sparse
  structured prediction.
\newblock In \emph{ICML}, 2020.

\bibitem[Niculae et~al.(2018)Niculae, Martins, Blondel, and
  Cardie]{Niculae2018SparseMAPDS}
V.~Niculae, A.~F.~T. Martins, M.~Blondel, and C.~Cardie.
\newblock Sparsemap: Differentiable sparse structured inference.
\newblock In \emph{ICML}, 2018.

\bibitem[{Papandreou} and {Yuille}(2011)]{Papandreou:2011}
G.~{Papandreou} and A.~L. {Yuille}.
\newblock Perturb-and-map random fields: Using discrete optimization to learn
  and sample from energy models.
\newblock In \emph{2011 International Conference on Computer Vision}, pages
  193--200, 2011.

\bibitem[Paulus et~al.(2020)Paulus, Choi, Tarlow, Krause, and
  Maddison]{paulus2020gradient}
M.~B. Paulus, D.~Choi, D.~Tarlow, A.~Krause, and C.~J. Maddison.
\newblock Gradient estimation with stochastic softmax tricks.
\newblock \emph{arXiv preprint arXiv:2006.08063}, 2020.

\bibitem[Pogan{\v{c}}i{\'c} et~al.(2019)Pogan{\v{c}}i{\'c}, Paulus, Musil,
  Martius, and Rolinek]{poganvcic2019differentiation}
M.~V. Pogan{\v{c}}i{\'c}, A.~Paulus, V.~Musil, G.~Martius, and M.~Rolinek.
\newblock Differentiation of blackbox combinatorial solvers.
\newblock In \emph{International Conference on Learning Representations}, 2019.

\bibitem[Raedt et~al.(2016)Raedt, Kersting, Natarajan, and
  Poole]{raedt2016statistical}
L.~D. Raedt, K.~Kersting, S.~Natarajan, and D.~Poole.
\newblock Statistical relational artificial intelligence: Logic, probability,
  and computation.
\newblock \emph{Synthesis Lectures on Artificial Intelligence and Machine
  Learning}, 10\penalty0 (2):\penalty0 1--189, 2016.

\bibitem[Rol{\'\i}nek et~al.(2020)Rol{\'\i}nek, Swoboda, Zietlow, Paulus,
  Musil, and Martius]{rolinek2020deep}
M.~Rol{\'\i}nek, P.~Swoboda, D.~Zietlow, A.~Paulus, V.~Musil, and G.~Martius.
\newblock Deep graph matching via blackbox differentiation of combinatorial
  solvers.
\newblock In \emph{ECCV}, 2020.

\bibitem[Schulman et~al.(2015)Schulman, Heess, Weber, and
  Abbeel]{schulman2015gradient}
J.~Schulman, N.~Heess, T.~Weber, and P.~Abbeel.
\newblock Gradient estimation using stochastic computation graphs.
\newblock In \emph{Proceedings of the 28th International Conference on Neural
  Information Processing Systems-Volume 2}, pages 3528--3536, 2015.

\bibitem[Shpakova and Bach(2016)]{shpakova:2016}
T.~Shpakova and F.~Bach.
\newblock Parameter learning for log-supermodular distributions.
\newblock In \emph{Advances in Neural Information Processing Systems},
  volume~29, pages 3234--3242, 2016.

\bibitem[Song et~al.(2016)Song, Schwing, Urtasun, et~al.]{song2016training}
Y.~Song, A.~Schwing, R.~Urtasun, et~al.
\newblock Training deep neural networks via direct loss minimization.
\newblock In \emph{International Conference on Machine Learning}, pages
  2169--2177, 2016.

\bibitem[Tucker et~al.(2017)Tucker, Mnih, Maddison, Lawson, and
  Sohl-Dickstein]{tucker2017rebar}
G.~Tucker, A.~Mnih, C.~J. Maddison, D.~Lawson, and J.~Sohl-Dickstein.
\newblock Rebar: Low-variance, unbiased gradient estimates for discrete latent
  variable models.
\newblock \emph{Advances in Neural Information Processing Systems}, 2017.

\bibitem[Wainwright and Jordan(2008)]{wainwright2008graphical}
M.~J. Wainwright and M.~I. Jordan.
\newblock \emph{Graphical models, exponential families, and variational
  inference}.
\newblock Now Publishers Inc, 2008.

\bibitem[Wang et~al.(2019)Wang, Donti, Wilder, and Kolter]{wang2019satnet}
P.-W. Wang, P.~Donti, B.~Wilder, and Z.~Kolter.
\newblock Satnet: Bridging deep learning and logical reasoning using a
  differentiable satisfiability solver.
\newblock In \emph{International Conference on Machine Learning}, pages
  6545--6554. PMLR, 2019.

\bibitem[Wilder et~al.(2019)Wilder, Dilkina, and Tambe]{wilder2019melding}
B.~Wilder, B.~Dilkina, and M.~Tambe.
\newblock Melding the data-decisions pipeline: Decision-focused learning for
  combinatorial optimization.
\newblock In \emph{Proceedings of the AAAI Conference on Artificial
  Intelligence}, volume~33, pages 1658--1665, 2019.

\bibitem[Xi'an(2016)]{214875}
Xi'an.
\newblock Which pdf of x leads to a gumbel distribution of the finite-size
  average of x?
\newblock Cross Validated, 2016.
\newblock URL \url{https://stats.stackexchange.com/q/214875}.
\newblock URL:https://stats.stackexchange.com/q/214875 (version: 2016-05-27).

\bibitem[Xie and Ermon(2019)]{Xie2019ReparameterizableSS}
S.~M. Xie and S.~Ermon.
\newblock Reparameterizable subset sampling via continuous relaxations.
\newblock In \emph{IJCAI}, 2019.

\end{thebibliography}
\bibliographystyle{abbrvnat}

\ifbool{printChecklist}{\include{checklist}}

\ifbool{printApdx}{\appendix

\section{Standard Maximum Likelihood Estimation and Links to \imle}

\label{apx-mle}

In the standard MLE setting \citep[see, e.g., ][Ch. 9]{murphy2012machine} we are interested in learning the parameters of a probability distribution, here assumed to be from the (constrained) exponential family (see \cref{def-constrained-exp-family}), given a set of example states. 
More specifically,  
given training data  $\mathcal{D} = \{\hbx_j\}_{j=1}^N$, with $\hbx_j\in \mathcal{C}\subseteq \{0, 1\}^m$, maximum-likelihood learning aims to minimize the empirical risk 
\begin{equation}
\label{eq-apx-std-MLE-loss}
\mleobj(\btheta, q_\mathcal{D}) 
= \mathbb{E}_{\exz\sim q_{\mathcal{D}}}[- \log p(\hbx; \btheta) ]
=
\frac{1}{N} \sum_{j=1}^{N} -\log p(\hbx_j; \btheta) = 
\frac{1}{N} \sum_{j=1}^{N} \left( A(\btheta) 
- \langle\hbx_j,\btheta\rangle
\right)
\end{equation}
with respect to $\btheta$, where $q_{\mathcal{D}}(\bz)=\sum_j \delta_{\exz_j}(\bz) /N $ 
is the empirical data distribution and $\delta_{\exz}$ is the Dirac delta centered in $\exz$.
In \cref{eq-apx-std-MLE-loss}, the point-wise loss $\ell$ is the negative log likelihood $- \log p(\exz, \btheta)$, and $q_{\mathcal{D}}$ may be seen as a (data/empirical) \emph{target distribution}. 
Note that in the main paper, as we assumed $q$ to be from the exponential family with parameters $\btheta'$, we used the notation $\mathcal{L}(\btheta, \btheta')$ to indicate the MLE objective rather than $\mathcal{L}(\btheta, q)$. 
These two definitions are, however,  essentially equivalent.

\cref{eq-apx-std-MLE-loss} is a smooth objective that can be optimized with a (stochastic) gradient descent procedure. 
For a data point $\hbx$, the gradient of the point-wise loss is given by $\grad{\btheta}{\ell} = \bmu(\btheta) - \hbx$, since $\grad{\btheta}{A}=\bmu$.
For the entire dataset, one has 
\begin{equation}
    \label{eq-apx-grad_smle}
    \grad{\btheta}{\mleobj(\btheta, q_{\mathcal{D}})} = \bmu(\btheta) - \frac{1}{N} \sum_{j=1}^N \exz_j =
    \bmu(\btheta) - \mathbb{E}_{\exz\sim q_{\mathcal{D}}}[ \exz_j ]
\end{equation}
which is (cf.  \cref{eq-grad-mleobj}) the difference between the marginals of $\latentprobdist$ (the mean of $\mathbf{Z}$) and the empirical mean of $\mathcal{D}$,
$
\bmu(q_{\mathcal{D}}) =
\mathbb{E}_{\exz\sim\mathcal{D}}[\exz]
$. 
As mentioned in the main paper, the main computational challenge when evaluating \cref{eq-apx-grad_smle} is to compute the marginals (of $\latentprobdist$). 
There are many approximate schemes, one of which is the so-called \emph{perceptron rule}, which approximate \cref{eq-apx-grad_smle} as
\begin{equation*}
    \mlegradient(\btheta, q_{\mathcal{D}} ) = \fMAP(\btheta) - \frac{1}{N} \sum_{j=1}^N \exz_j 
\end{equation*}
and it is frequently employed in a stochastic manner by sampling one or more points from $\mathcal{D}$, rather than computing the full dataset mean.

We may interpret the standard MLE setting described in this section from the perspective of the problem setting we presented in \cref{sec-ps}.
The first indeed 
amounts to the special case of the latter where there are no inputs ($\mathcal{X}=\emptyset$), the  (target) output space coincides with the state space of the distribution ($\mathcal{Y}=\mathcal{Z}$), $f$ is the identity mapping, $\ell$ is the negative log-likelihood and the model's parameter coincide with the distribution parameters, that is $\bomega=\btheta$.

\section{Proofs of Section \ref{section-perturb-and-map} and Section~\ref{section-optimal-q-co}}

\label{proofs-imle}

This section contains the proofs of the results relative to the perturb and map section (\cref{section-perturb-and-map}) and the section on optimal target distributions for typical loss functions when backpropagating through combinatorial optimization problems (section~\ref{section-optimal-q-co}).
We repeat the statements here, for the convenience of the reader.

\begin{mirrorproposition}
Let $p(\bz; \btheta)$ be a discrete exponential family distribution with constraints $\mathcal{C}$ and temperature $\tau$, and let $\langle\btheta,\bz\rangle$ be the unnormalized weight of each $\bz$ with $\bz \in \mathcal{C}$. Moreover, let $\tilde{\btheta}$ be such that, for all $\bz \in \mathcal{C}$,
$$\langle\bz,\tilde{\btheta}\rangle =  \langle\bz,\btheta\rangle + \epsilon(\bz)$$ with each
$\epsilon(\bz)$ i.i.d. samples from $\GumbelDist(0,\tau)$. Then, $$\Pr\left(\mathtt{MAP}(\tilde{\btheta})=\bz\right) = p(\bz; \btheta).$$
\end{mirrorproposition}

\begin{proof}
Let $\epsilon_i \sim \GumbelDist(0,\tau)$ i.i.d. and $\ttheta_i = \theta_i + \epsilon_i$. Following a derivation similar to one made in \citet{Papandreou:2011}, we have:
\begin{align*}
    & \Pr\{\argmax(\ttheta_1, \ldots, \ttheta_m) = n\}  = \\
    = & \Pr\{ \ttheta_n \geq \max_{j\neq n} \{\ttheta_j\} \}  \\
    = & \int_{-\infty}^{+\infty} g(t; \theta_n) \prod_{j\neq n} G(t; \theta_j) \,dt \\
    = & \int_{-\infty}^{+\infty} \frac{1}{\tau} \exp\left(\frac{\theta_n - t}{\tau} - e^{\frac{\theta_n - t}{\tau}}\right) \prod_{j\neq n} \exp\left( -e^{\frac{\theta_j - t}{\tau}}\right) \,dt \\
    = &  \int_{-\infty}^{+\infty} \frac{1}{\tau} e^{\frac{\theta_n - t}{\tau}}\exp\left(-e^{\frac{\theta_n-t}{\tau}}\right) \prod_{j\neq n} \exp\left(-e^{\frac{\theta_j - t}{\tau}}\right) \,dt \\
    = & \int_{0}^{1} \prod_{j \neq n} z^{\exp\left( \frac{\theta_j - \theta_n}{\tau}  \right)} \,dz \qquad \text{with } z \triangleq \exp\left( -e^{\frac{\theta_n - t}{\tau}} \right) \\
    = & \frac{1}{1 + \sum_{j \neq n} e^{\frac{\theta_j - \theta_n}{\tau}} } \\
    = & \frac{e^{\frac{\theta_n}{\tau}}}{\sum_{j=1}^{m} e^{\frac{\theta_j}{\tau}}, }
  \end{align*}
  where $g$ and $G$ are respectively the Gumbel probability density function and the Gumbel cumulative density function.
  The proposition now follows from arguments made in \citet{Papandreou:2011} using the maximal equivalent re-parameterization of $p(\bz; \btheta)$ where we specify a parameter $\langle \btheta,\bz\rangle$ for each $\bz$ with $\mathcal{C}(\bz)$ and perturb these parameters. 
\end{proof}

\begin{mirrorlemma}
Let $X \sim \mbox{Gumbel}(0 ,\tau)$ and let $\kappa \in \mathbb{N} \setminus \{0\}$.  Then we can write 
\begin{equation*}
X \sim \sum_{j=1}^{\kappa} \frac{\tau}{\kappa} \left[ \lim_{s \rightarrow \infty}\sum_{i=1}^{s} \left\{\GammaDist(1/\kappa, \kappa/i)\right\} - \log(s)\right],
\end{equation*}
where $\GammaDist(\alpha, \beta)$ is the gamma distribution with shape $\alpha$ and scale $\beta$.
\end{mirrorlemma}

\begin{proof}
Let $\kappa \in \mathbb{N} \setminus \{0\}$ and let $X \sim \GumbelDist(0, \tau)$. Its moment generating function has the form 
\begin{equation}
\label{eq:m-gen-f-gumb}
\mathbb{E}[\exp(tX)] = \Gamma(1 - \tau t).
\end{equation}
As mentioned in \citet[p. 443,][]{johnson1998advances} we know that we can write the Gamma function as 
\begin{equation}
\label{eq-gamma-book}
\Gamma(1 - \tau t) = e^{\gamma \tau t} \prod_{i=1}^{\infty} \left(1 - \frac{\tau t}{i}\right)^{-1} e^{\frac{-\tau t}{i}}
\end{equation}
where $\gamma$ is the Euler-Mascheroni constant.
We have that 
$$\left(1 - \frac{\tau t}{i}\right)^{-1} = \frac{i}{i - \tau t} = \frac{\frac{i}{\tau}}{\frac{i - \tau t}{\tau}} =  \frac{\frac{i}{\tau}}{\frac{i}{\tau} - \frac{\tau t}{\tau}} = 
\frac{\frac{i}{\tau}}{\frac{i}{\tau} - t}.$$ 
The last term is the moment generating function of an exponential distribution with scale $\frac{\tau}{i}$. 
We can now take the logarithm on both sides of \cref{eq-gamma-book} and obtain
\begin{equation*}
  tX = \gamma\tau t + \lim_{s\rightarrow \infty}\sum_{i=1}^{s} \left( t \ExpDist(\tau/i) - \frac{\tau t}{i} \right),
\end{equation*}
where $\ExpDist(\alpha)$ is the exponential distribution with scale $\alpha$.
Hence, 
\begin{align*}
 X \sim & \lim_{s\rightarrow \infty}\sum_{i=1}^{s}  \left( \ExpDist(\tau/i) - \frac{\tau}{i} \right) +  \gamma\tau \\
  = & \lim_{s\rightarrow \infty}\sum_{i=1}^{s}  \left( \ExpDist(\tau/i) - \frac{\tau}{i} \right) + \tau \lim_{s\rightarrow \infty}\sum_{i=1}^{s} \frac{1}{i} - \log(s) \\
  = & \lim_{s\rightarrow \infty}\sum_{i=1}^{s}  \left( \ExpDist(\tau/i) - \frac{\tau}{i} + \frac{\tau}{i} \right) - \tau\log(s)   \\
    = & \lim_{s\rightarrow \infty}\sum_{i=1}^{s}  \ExpDist(\tau/i) - \tau\log(s)
\end{align*}
 
Since $\ExpDist(\alpha) \sim \GammaDist(1, \alpha)$, and due to the scaling and summation properties of the Gamma distribution (with shape-scale parameterization), we can write for all $r>1$:
$$\ExpDist(\alpha) \sim \sum_{j=1}^{r} \GammaDist(1/r, \alpha r)/r.$$
Hence, picking $r=\kappa$ from the hypothesis, we have 

\begin{align*}
    X \sim & \lim_{s\rightarrow \infty} \left\lbrace 
        \sum_{i=1}^{s} \sum_{j=1}^{\kappa} \GammaDist(1/\kappa, \tau \kappa/i)/\kappa
    \right\rbrace - \tau \log(s) \\
    = & 
    \lim_{s\rightarrow \infty} \left\lbrace 
        \sum_{j=1}^{\kappa} \frac{\tau}{k}\sum_{i=1}^{s} \GammaDist(1/\kappa, \kappa/i)
    \right\rbrace - \sum_{j=1}^\kappa \frac{\tau}{\kappa} \log(s) \\
    = & 
    \lim_{s\rightarrow \infty} 
    \sum_{j=1}^{\kappa} \frac{\tau}{\kappa} 
    \left\lbrace
        \left[
            \sum_{i=1}^{s} \GammaDist(1/\kappa, \kappa/i)
        \right] - \log(s)
    \right\rbrace \\
    = & 
    \sum_{j=1}^{\kappa} 
    \frac{\tau}{\kappa}
    \left\lbrace
        \lim_{s\rightarrow \infty} 
        \left[
            \sum_{i=1}^{s} \GammaDist(1/\kappa, \kappa/i)
        \right] - \log(s)
    \right\rbrace 
\end{align*}
This concludes the proof.
Parts of the proof are inspired by a post on stackexchange~\cite{214875}.
\end{proof}

\begin{mirrortheorem}
Let $p(\bz; \btheta)$ be a discrete exponential family distribution with constraints $\mathcal{C}$ and temperature $\tau$, and let $k \in \mathbb{N} \setminus \{0\}$. Let us assume that if $\mathcal{C}(\bz)$ then $\langle\bz, \mathbf{1}\rangle = k$.
Let $\tilde{\btheta}$ be the perturbation obtained by $\tilde{\btheta}_i = \btheta_i + \epsilon_i$
with 
\begin{equation}
    \epsilon_i \sim \frac{\tau}{k} \left[\lim_{s\to\infty} \sum_{i=1}^s \left\{\GammaDist(1/k,k/i)\right\} - \log(s)\right],
\end{equation}
where $\GammaDist(\alpha, \beta)$ is the gamma distribution with shape $\alpha$ and scale $\beta$.
Then, for every $\bz$ we have that $\langle \bz, \tilde{\btheta}\rangle = \langle \bz, \btheta\rangle + \epsilon(\bz)$ with  $\epsilon(\bz) \sim \GumbelDist(0, \tau)$.
\end{mirrortheorem}

\begin{proof}

Since we perturb each $\theta_i$ by $\varepsilon_i$ 
 we have, by assumption, that $\langle\btheta, \mathbf{1}\rangle = k$, for every $\bz$ with $\mathcal{C}(\bz)$, that
\begin{equation}
    \langle \bz, \tilde{\btheta} \rangle = \langle  \bz, \btheta \rangle + \sum_{j=1}^k \varepsilon_j.  %
\end{equation}
Since by \cref{lemma-gamma-gumbel} we know that $\sum^{k}_{j=1} \varepsilon_i \sim \GumbelDist(0, \tau)$, the statement of the theorem follows.
\end{proof}

The following  theorem shows that the infinite series from \cref{lemma-gamma-gumbel} can be well approximated by a finite sum using convergence results for the Euler-Mascheroni series.

\begin{theorem}
Let $X\sim \GumbelDist(0, \tau)$ and $\tilde{X}(m) \sim \sum_{j=1}^{\kappa}  \frac{\tau}{\kappa} \left[ \sum_{i=1}^{m} \{ \GammaDist(1/\kappa, \kappa/i)\} - \log(m) \right]$. Then 
$$\frac{\tau}{2(m+1)} < \mathbb{E}[\tilde{X}(m)] - \mathbb{E}[X]  < \frac{\tau}{2m}.$$
\end{theorem}

\begin{proof}
We have that 
\begin{align*}
& \mathbb{E}[\tilde{X}(m)] = \\
& \quad = \mathbb{E} \left[ \sum_{j=1}^{\kappa}  \frac{\tau}{\kappa} \left[ \sum_{i=1}^{m} \{ \GammaDist(1/\kappa, \kappa/i)\} - \log(m) \right]\right] \\
& \quad = \sum_{i=1}^{m}   \mathbb{E}\left[\GammaDist(1/\kappa, \tau \kappa/i)\right] - \tau \log(m) \\
& \quad = \left[ \sum_{i=1}^{m} \frac{1}{\kappa} \frac{\tau \kappa}{i} - \tau \log(m) \right] \\
& \quad = \tau \left[  \sum_{i=1}^{m} \frac{1}{i} - \log(m) \right].
\end{align*}
If $X\sim \GumbelDist(0, \tau)$, we know that $\mathbb{E}\left[X \right] = \tau \gamma$.
Hence,
\begin{align*}
  \mathbb{E}[\tilde{X}(m)] - \mathbb{E}[X] & = \tau \left[  \sum_{i=1}^{m} \frac{1}{i} - \log(m) \right] -   \tau \gamma\\
& = \tau \left[ \sum_{i=1}^{m} \frac{1}{i} - \log(m) - \gamma  \right].
\end{align*}
The theorem now follows from convergence results of the Euler-Mascheroni series~\cite{MORTICI2010}. 
\end{proof}

\begin{mirrorfact}
If one uses
$\ell_H$, then \imle with the target distribution 
of \cref{eqn-target-q-optimal}
and $\bnoisedist=\delta_0(\bepsilon)$
is equivalent to 
the perceptron-rule estimator of the MLE objective between $p(\bz; h_{\bv}(\exx_j))$ and $\exy_j$.
\end{mirrorfact}

\begin{proof}
Rewriting the definition of the Hamming loss gives us 
$$\ell_H(\bz, \by) = \frac{1}{m}\sum_{i=1}^{m} \left( \bz_i + \by_i - 2 \bz_i\by_i \right).$$
Hence, we have that $$\grad{\bz_i}{\ell_H} = \frac{1}{m}\left(1 - 2\by_i\right).$$ 
Therefore, $\grad{\bz_i}{\ell_H} = -\frac{1}{m}$ if $\by_i = 1$ and $\grad{\bz_i}{H} = \frac{1}{m}$ if $\by_i = 0$.   
Since, by definition $\by \in \mathcal{C}$, we have that
$$\fMAP\left(-\grad{\bz}{\ell_H}\right) = \by.$$ 
Now, when using \imle with $S=1$ and $\bnoisedist=\delta_0(\bepsilon)$ we approximate the gradients as  
\[\mlegradient(\btheta, \btheta') = \fMAP(\btheta) - \fMAP(\btheta') = \fMAP(\btheta) - \fMAP\left(-\grad{\bz}{\ell_H}\right) = \fMAP(\btheta)  - \by.\]
This concludes the proof. 
\end{proof}

\begin{mirrorfact}
If one uses $\ell_R$ 
then \imle with the target distribution 
of \cref{eqn-target-q-optimal} is equivalent to the perturb-and-MAP estimator of the MLE objective between $p(\bz; h_{\bv}(\exx_j))$ and $p(\bz; - \exc_j)$. 
\end{mirrorfact}

\begin{proof}
We have that $\grad{\bz_i}{\ell_R} = \bc_i$ for all $i$.
Now, when using \imle with target distribution $\hat{q}(\bz; \btheta')$ of \cref{eqn-target-q-optimal} (and without loss of generality, for $S=1$) we have that $\hat{q}(\bz; \btheta') = p(\bz; -\bc)$, and we approximate the gradients as  
\[ \mlegradient(\btheta, \btheta') =  \fMAP(\btheta + \bepsilon_i) - \fMAP(\btheta' + \bepsilon_i) = \fMAP(\btheta + \bepsilon_i) - \fMAP(-\bc + \bepsilon_i),  \;  \text{where} \;
    \bepsilon_i \sim \bnoisedist. \]
Hence, \imle approximates the gradients of the maximum likelihood estimation problem between $p(\bz; h_{\bv}(\exx_j))$ and $p(\bz; - \exc_j)$ using perturb-and-MAP. This concludes the proof. 
\end{proof}

\clearpage

\section{Experiments: Details and Additional Results}

\subsection{Synthetic Experiments}
\label{ax-toy-topk}

\begin{figure}[t]
    \centering
    \includegraphics[width=0.32\textwidth]{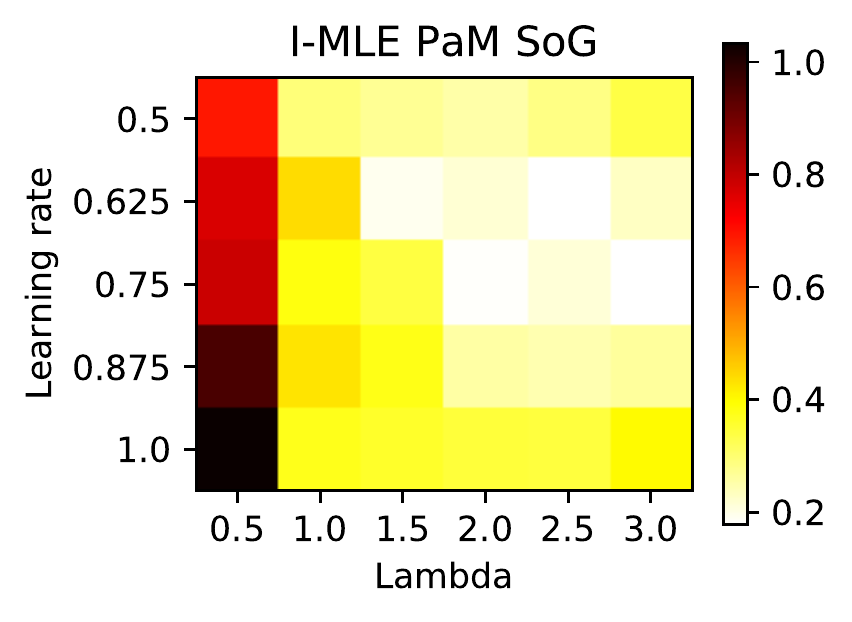}
    \includegraphics[width=0.32\textwidth]{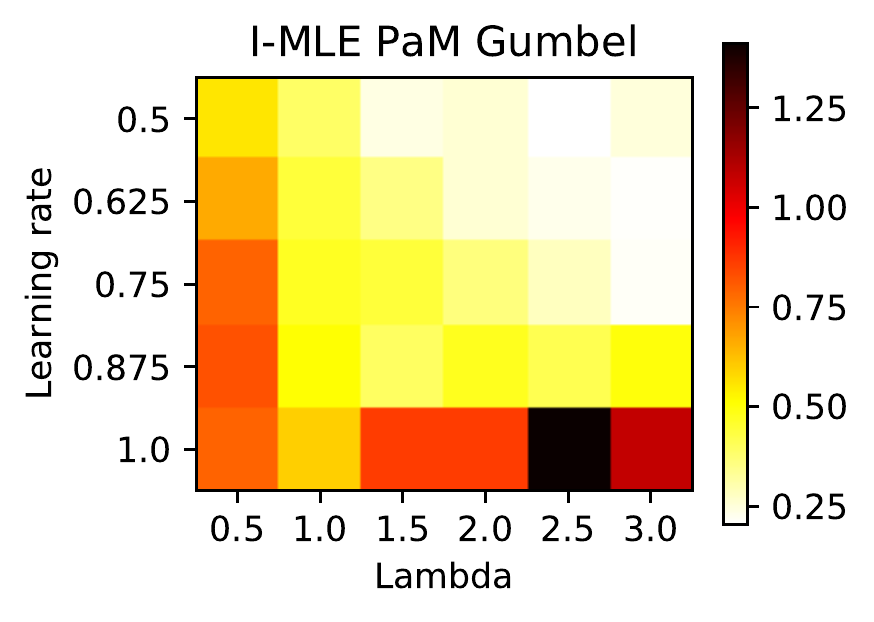}
    \includegraphics[width=0.32\textwidth]{toy_topk/cm_I-MLE-SoG-Gum-means.pdf}
    \caption{Average (over 100 runs) values of $L(\btheta)$ after 50 steps of stochastic gradient descent (with momentum) using single-sample \imle with $\SoGDist(1, 5, 10)$ noise (left) and $\GumbelDist(0, 1)$ noise (center) varying the perturbation intensity $\lambda$ (see \cref{eq-pid-q}) and learning rate. The rightmost heat-map depicts the (point-wise) difference between the two methods (blue = better SoG).}
    \label{fig:toy_exps_apx_means}
\end{figure}
\begin{figure}[t]
    \centering
    \includegraphics[width=0.32\textwidth]{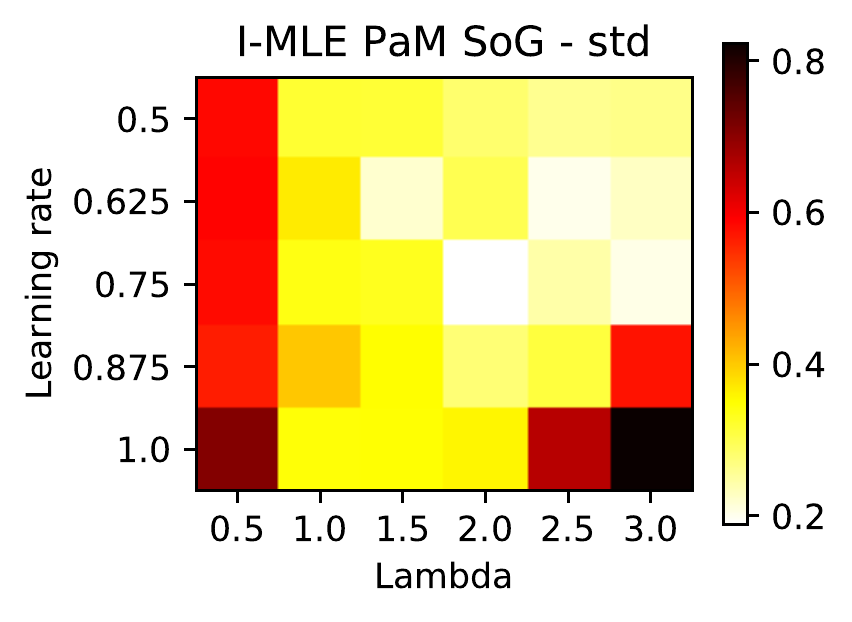}
    \includegraphics[width=0.32\textwidth]{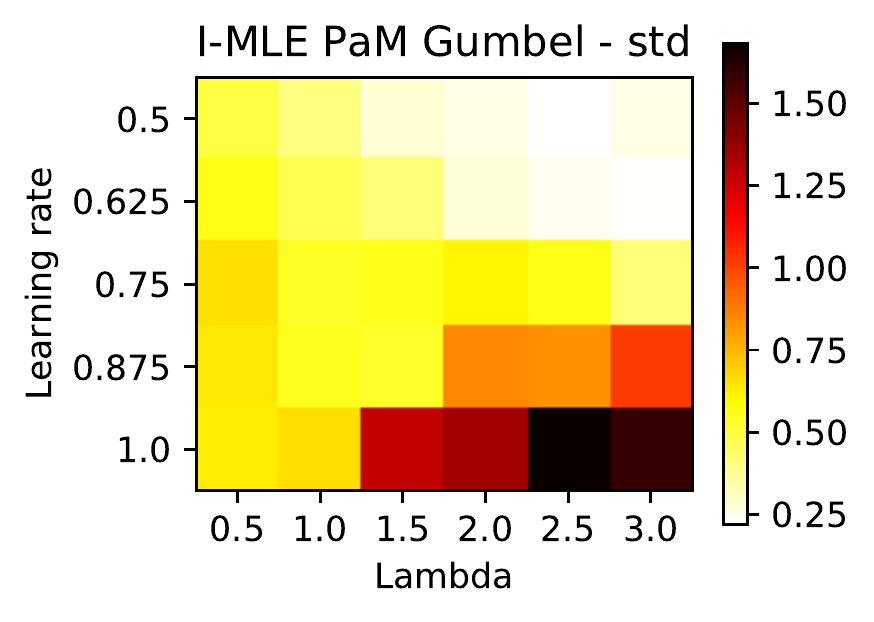}
    \includegraphics[width=0.32\textwidth]{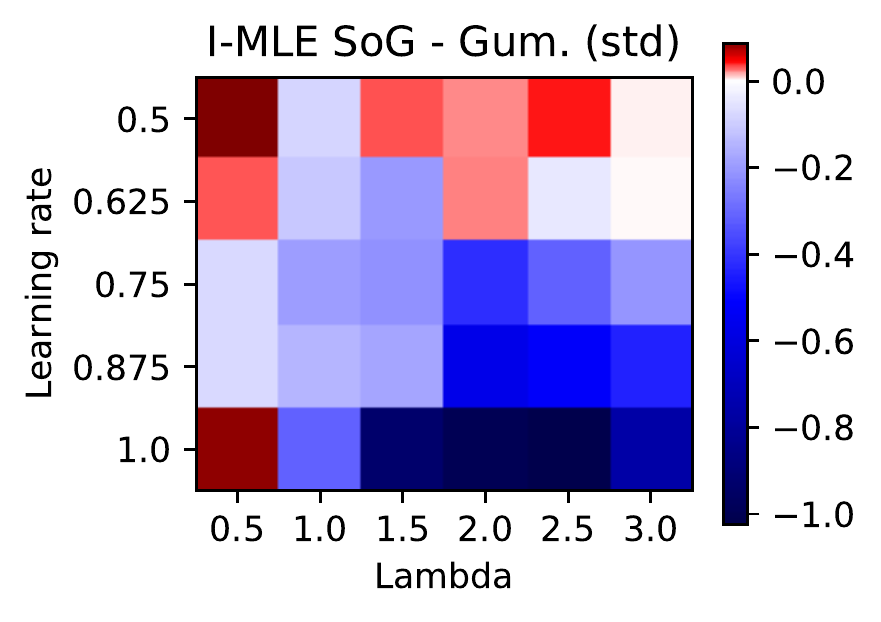}
    \caption{Same as above, but reporting standard deviations.}
    \label{fig:toy_exps_apx_1}
\end{figure}

In this series of experiments  we analyzed the behaviour of various discrete gradient estimators, comparing our proposed \imle with standard straight-trhough (STE) and score-function (SFE) estimators. 
We also study of the effect of using Sum-of-Gamma perturbations rather than standard Gumbel noise.
In order to be able to compute exactly (up to numerical precision) all the quantities involved, we chose a  tractable $5$-subset distribution (see Example 2) of size $m=10$. 
\begin{wrapfigure}[13]{r}{0.38\textwidth}
\centering
\vspace{-4mm}
\includegraphics[width=0.37\textwidth]{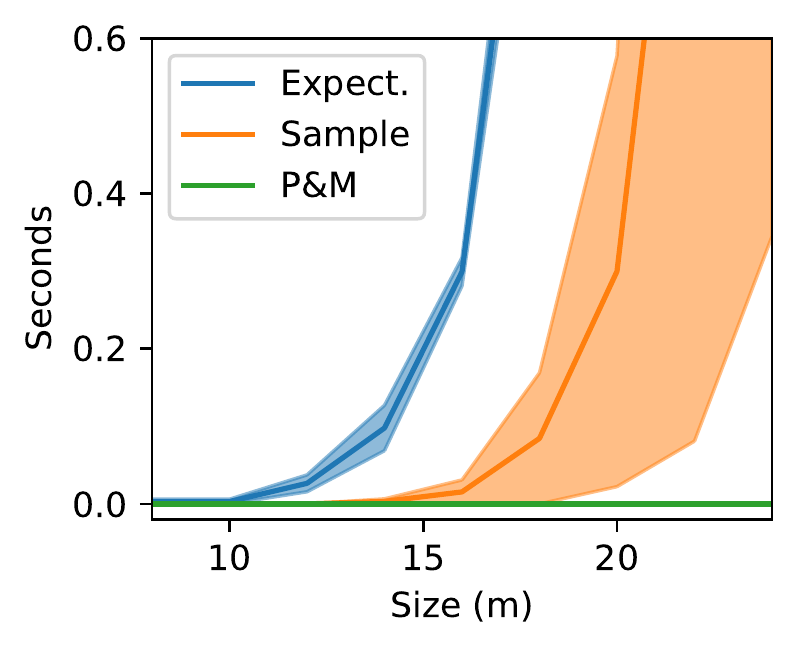}%
\vspace{-4mm}
\caption{\small \label{fig:time} Runtime (mean and standard deviation) for computing $L$ and samples of it, as the dimensionality of the $k$-subset distribution increases (with $k=m/2$). 
}
\end{wrapfigure}
We set the loss to $L(\bm{\theta}) = \mathbb{E}_{\exz \sim p(\mathbf{z}; \bm{\theta})} [ \norm{\exz - \mathbf{b}}^2 ]$, where $\mathbf{b}$ is a fixed vector sampled (only once) from $\mathcal{N}(0,\mathbf{I})$. 
This amounts to an unconditional setup where there are no input features (as in the standard MLE setting of \cref{apx-mle}), but where the point-wise loss $\ell(\bz)$ is the Euclidean distance between the distribution output and a fixed vector $\mathbf{b}$. 
In \cref{fig:time} we plot the runtime (mean and standard deviation over 10 evaluations) of the full objective $L$ (Expect., in the plot), of a (faithful) sample of $\ell$ and of a perturb-and-MAP sample with Sum-of-Gamma noise distribution (P\&M) for increasing size $m$, with $k=m/2$. 
As it is evident from the plot, the runtime for both expectation and faithful samples, which require computing all the states in $\mathcal{C}$, increases exponentially, while perturb and MAP remains almost constant.

Within this setting, the one-sample \imle  estimator is
\begin{equation*}
    \widehat{\nabla}_{\mathrm{I\mbox{-}MLE}} L(\btheta) = \fMAP(\btheta + \epsilon) - \fMAP(\btheta' + \epsilon), \; \text{with} \; \epsilon\sim \SoGDist(1, 5, 10)
\end{equation*}
where $\btheta'=\btheta - \lambda [2 ( \exz  - \mathbf{b})]$, where   $\exz=\fMAP(\btheta + \epsilon)$ is a (perturb-and-MAP) sample, while the one-sample straight through estimator is 
\begin{equation*}
    \widehat{\nabla}_{\mathrm{STE}}{L(\btheta)} =  2 ( \exz  - \mathbf{b}), \; \text{with} \; \exz = \fMAP(\btheta + \epsilon), \; \epsilon \sim \GumbelDist(0, 1).
\end{equation*}
For the score function estimator, we have used an expansive faithful sample/full marginal implementation  given by 
\begin{equation*}
     \widehat{\nabla}_{\mathrm{SFE}}{L(\btheta)} =   \norm{\exz - \mathbf{b}}^2  \grad{\btheta}{\log p(\exz; \btheta)} 
     = \norm{\exz - \mathbf{b}}^2 [\exz - \bmu(\btheta)], \; \text{with} \; \exz \sim \latentprobdist 
\end{equation*}
since, in preliminary experiments, we did not manage to obtain meaningful results with SFE using perturb-and-MAP for sampling and/or marginals approximation.
These equations give the formulae for the estimators which we used for the results plotted in \cref{fig:toy_experiments} (top) in the main paper.

In \cref{fig:toy_exps_apx_means} we plot the heat-maps for the sensitivity results comparing between \imle with SoG and \imle with Gumbel perturbations. 
The two leftmost heat-maps depict the average value (over 100 runs) of $L(\btheta)$ after 50 steps of stochastic gradient descent, for various choices of $\lambda$ and learning rates (momentum factor was fixed at 0.9 for all experiments).
The rightmost plot of \cref{fig:toy_exps_apx_means} is the same as the one in the main paper, and represents the difference between the first and the second heat-maps. 
\cref{fig:toy_exps_apx_1} refers to the same setting, but this time showing standard deviations. 
The rightmost plot of \cref{fig:toy_exps_apx_1} suggests that using SoG perturbations results also in reduced variance (of the final loss) for most of the tried hyperparameter combinations.
Finally, in \cref{fig:toy_exps_apx_ste_sfe} we show sensitivity plots for STE (both with SoG and Gumbel perturbations) and SFE, where we vary the learning rate.

\begin{figure}[t]
    \centering
    \includegraphics[width=0.19\textwidth]{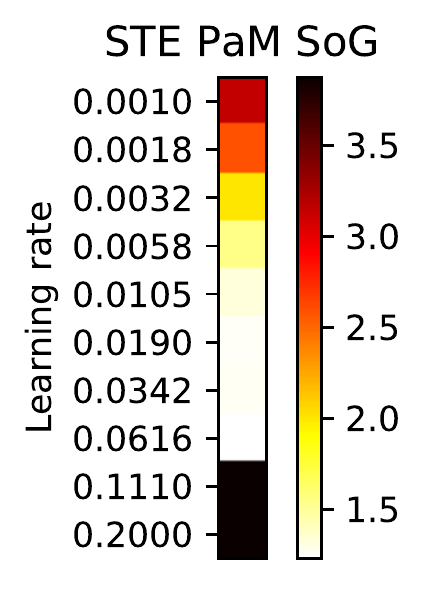}
    \includegraphics[width=0.19\textwidth]{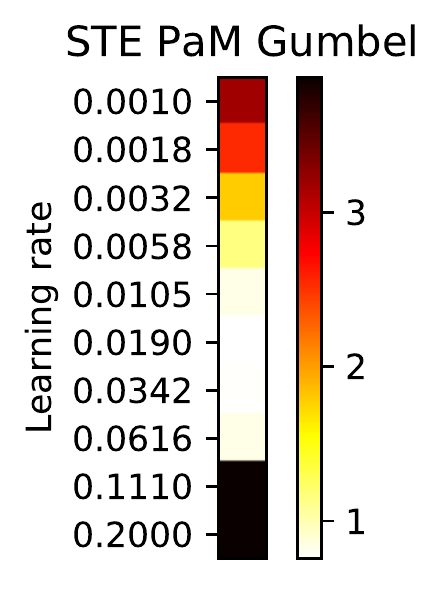}
    \includegraphics[width=0.19\textwidth]{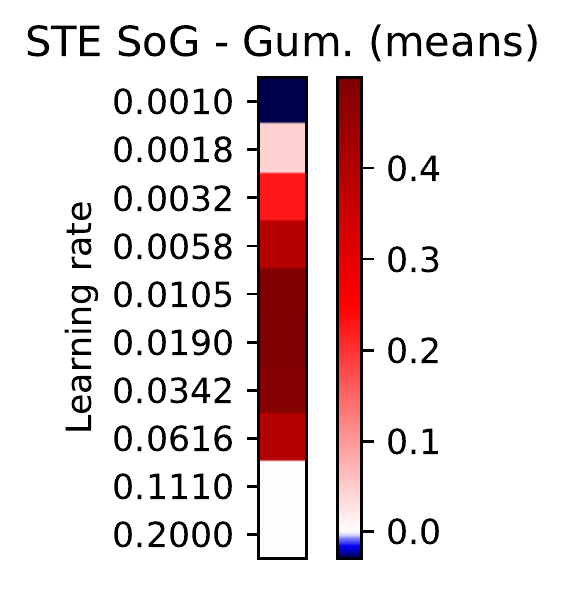}
    \includegraphics[width=0.19\textwidth]{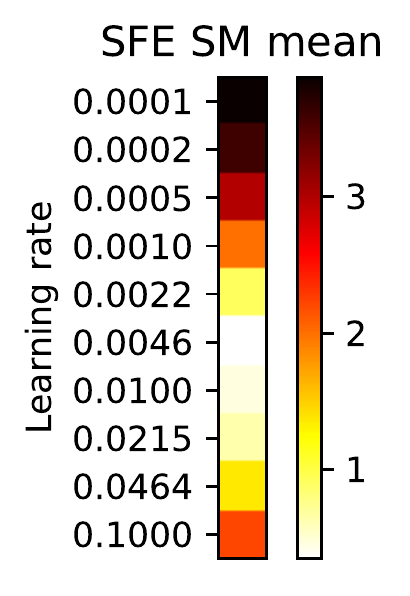}
    \includegraphics[width=0.19\textwidth]{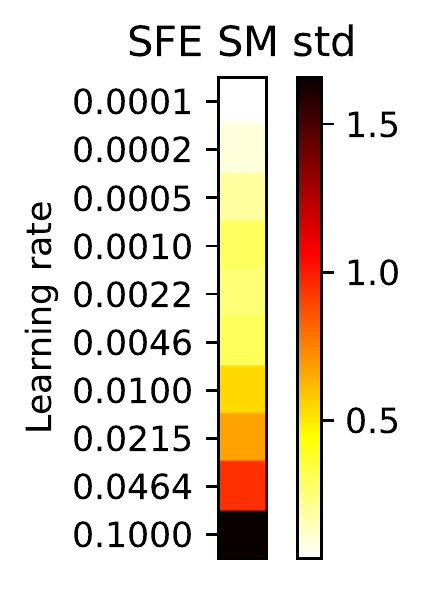}
    \caption{First three plots, from  left to right: average (over 100 runs) final values of $L(\btheta)$ after 50 steps of optimization using the straight-through estimator with SoG noise, varying the learning rate; same but using Gumbel noise; difference of averages between the first and the second heat-maps. Last two plots, from left to right: average (over 20 runs) final values of $L(\btheta)$ after 500 steps of optimization using the score function estimator (with faithful samples and exact marginals), varying the learning rate; standard deviation for the same setting.}
    \label{fig:toy_exps_apx_ste_sfe}
\end{figure}

\subsection{Learning to Explain}

\begin{table*}[t]
\small
    \centering
    \resizebox{\textwidth}{!}{
    \begin{tabular}{ccccccc}
    \toprule
    \multirow{2}{*}{\bf Method} & \multicolumn{2}{c}{\bf Appearance} & \multicolumn{2}{c}{\bf Palate} & \multicolumn{2}{c}{\bf Taste} \\
    \cmidrule(lr){2-3} \cmidrule(lr){4-5} \cmidrule(lr){6-7}
    & \multicolumn{1}{c}{\bf Test MSE}  & \multicolumn{1}{c}{\bf Subset precision} & \multicolumn{1}{c}{\bf Test MSE}  & \multicolumn{1}{c}{\bf Subset precision} & \multicolumn{1}{c}{\bf Test MSE}  & \multicolumn{1}{c}{\bf Subset precision}  \\
    \midrule
    L2X ($t=0.1$) & 10.70 $\pm$ 4.82 & 30.02 $\pm$ 15.82 & 6.70 $\pm$ 0.63 & 50.39 $\pm$ 13.58 & 6.92 $\pm$ 1.61 & 32.23 $\pm$ 4.92 \\
    SoftSub ($t=0.5$) & \ \textbf{2.48} $\pm$ 0.10 & 52.86 $\pm$ \ \ 7.08 &  \textbf{2.94} $\pm$ 0.08 &  39.17 $\pm$ \ \ 3.17 & \textbf{2.18} $\pm$ 0.10 & \textbf{41.98} $\pm$ 1.42 \\
    \textsc{I-Mle} ($\tau=30$) & \ \textbf{2.51} $\pm$ 0.05 & \textbf{65.47} $\pm$ \ \ 4.95 &  \textbf{2.96} $\pm$ 0.04 & 40.73 $\pm$ \ \ 3.15 & 2.38 $\pm$ 0.04 & \textbf{41.38} $\pm$ 1.55 \\
    \bottomrule
\end{tabular}
}
\caption{\label{table-all-aspects-k-10} Experimental results (mean $\pm$ std. dev.) for the learning to explain experiments for $k=10$ and various aspects.}
\end{table*}

Experiments were run on a server with Intel(R) Xeon(R) CPU E5-2637 v4 @ 3.50GHz, 4 GeForce GTX 1080 Ti, and 128 GB RAM. 

The pre-trained word embeddings and data set can be found here: \url{http://people.csail.mit.edu/taolei/beer/}.
Figure~\ref{fig-appendix-l2x} depicts the neural network architecture used for the experiments. As in prior work, we use a batch size of 40. The maximum review length is 350 tokens. We use the standard neural network architecture from prior work~\cite{chen2018learning,paulus2020gradient}. The dimensions of the token embeddings (of the embedding layers) are 200. All 1D convolutional layers have 250 filters with a kernel size of $3$. All dense layers have a dimension of 100. The dropout layer has a dropout rate of 0.2. The layer Multiply perform the multiplication between the token mask (output of I-MLE) and the embedding matrix. The Lambda layer computes the mean of the selected embedding vectors. The last dense layer has a sigmoid activation. IMLESubsetkLayer is the layer implementing \textsc{I-Mle}. We train for 20 epochs using the standard Adam settings in Tensorflow 2.4.1 (learning rate=0.001, beta1=0.9, beta2=0.999, epsilon=1e-07, amsgrad=False), and no learning rate schedule. The training time (for the 20 epochs) for \textsc{I-MLE},  with sum-of-Gamma perturbations, is $380$ seconds, for SoftSub $360$ seconds, and for L2X $340$ seconds. We always evaluate the model with the best validation MSE among the $20$ epochs. 

Implementations of \imle and all experiments will soon be made available. 
\cref{table-all-aspects-k-10} lists the results for L2X, SoftSub, and \textsc{I-Mle} for three additional aromas and $k=10$. 

\begin{table}[t!]
\centering
\resizebox{\columnwidth}{!}{
\setlength{\tabcolsep}{2pt}
\begin{tabular}{cccc}
\includegraphics[width=.25\linewidth]{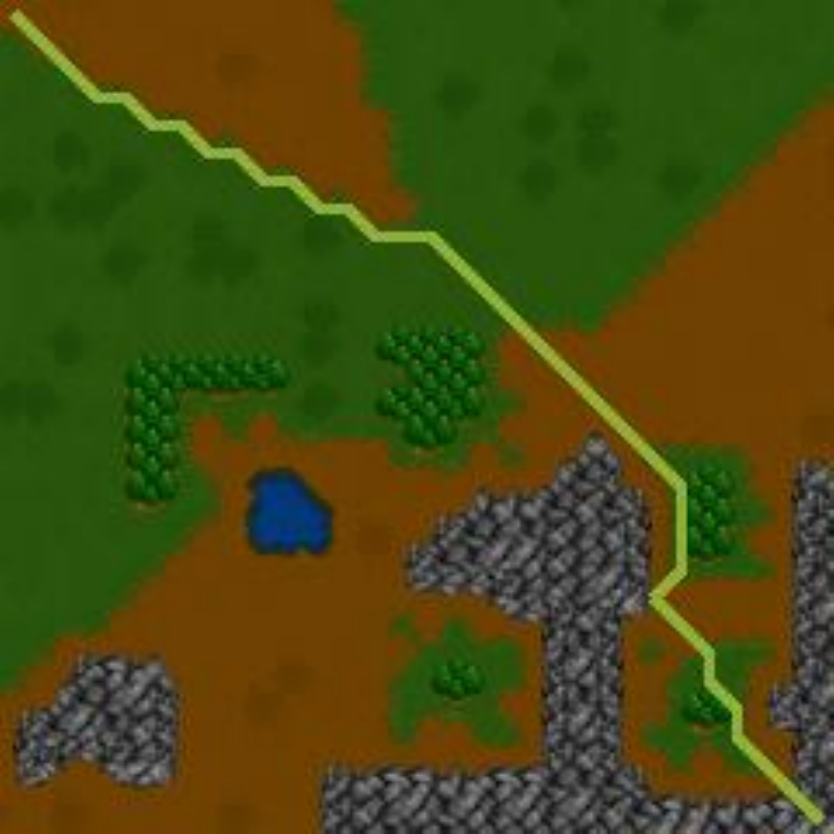} & \includegraphics[width=.25\linewidth]{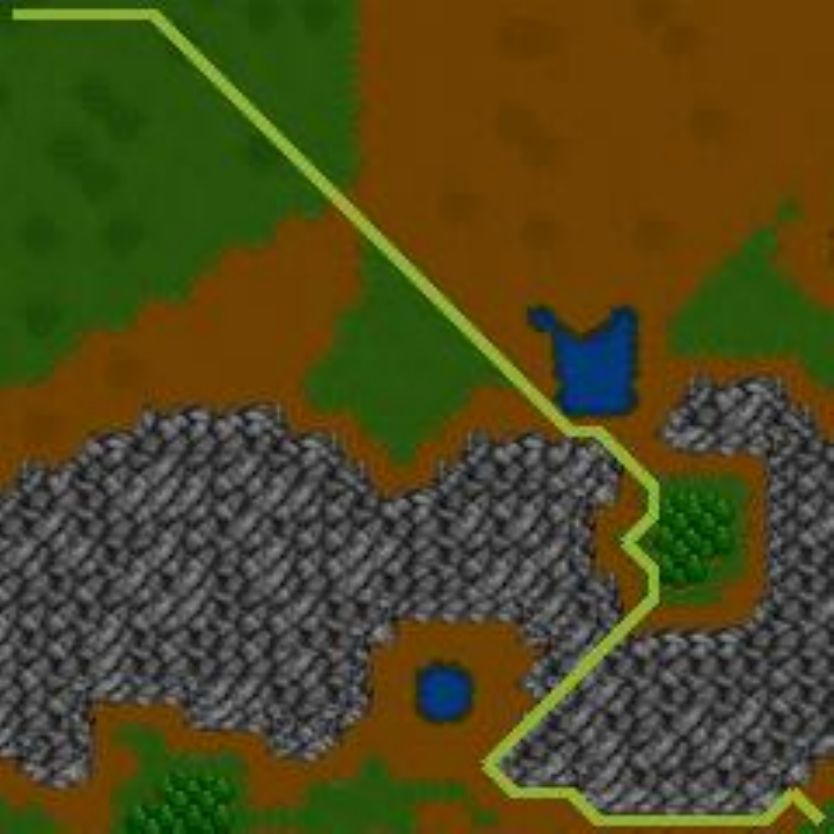} & \includegraphics[width=.25\linewidth]{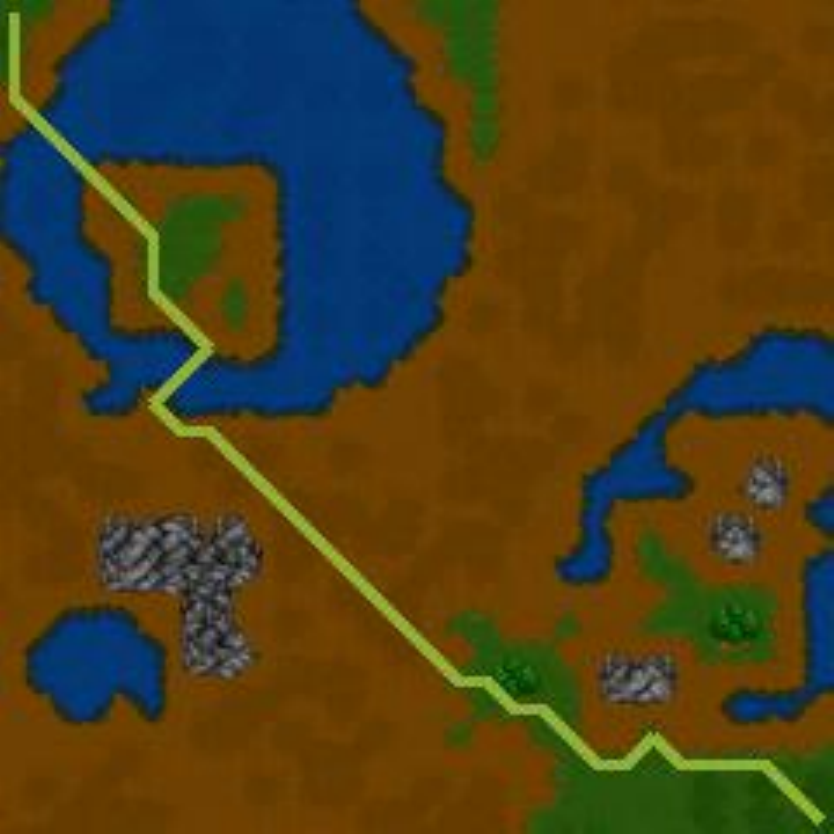} & \includegraphics[width=.25\linewidth]{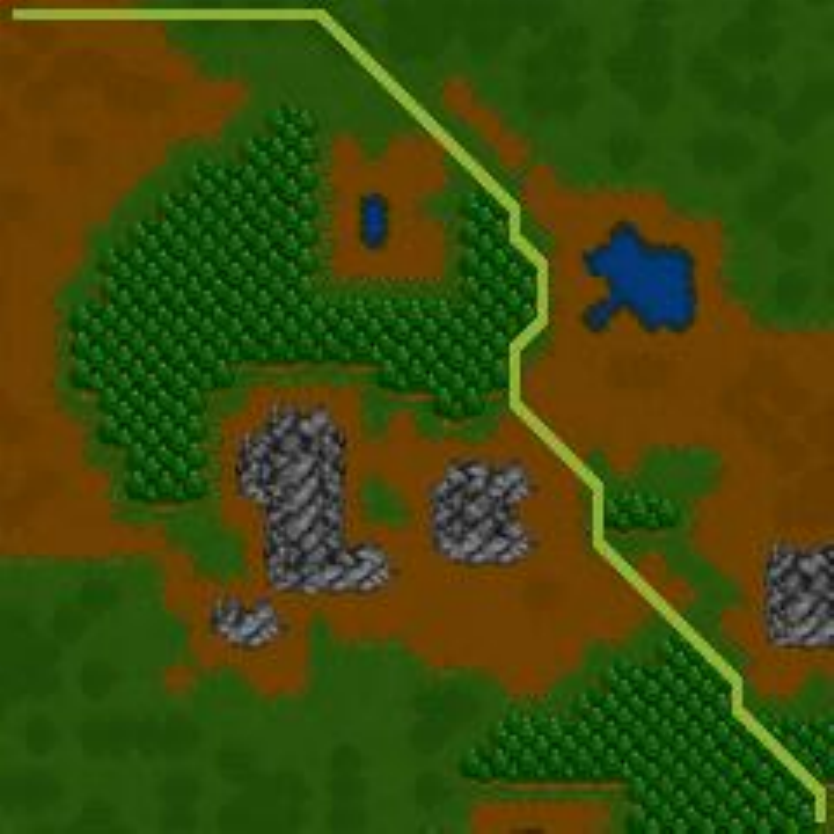}\\
\includegraphics[width=.25\linewidth]{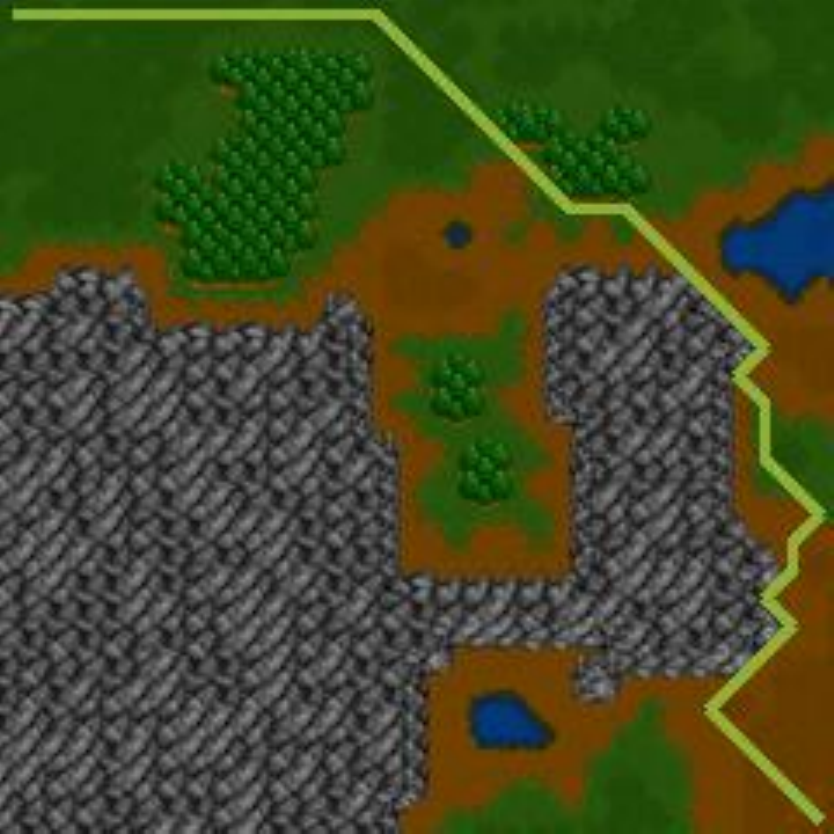} & \includegraphics[width=.25\linewidth]{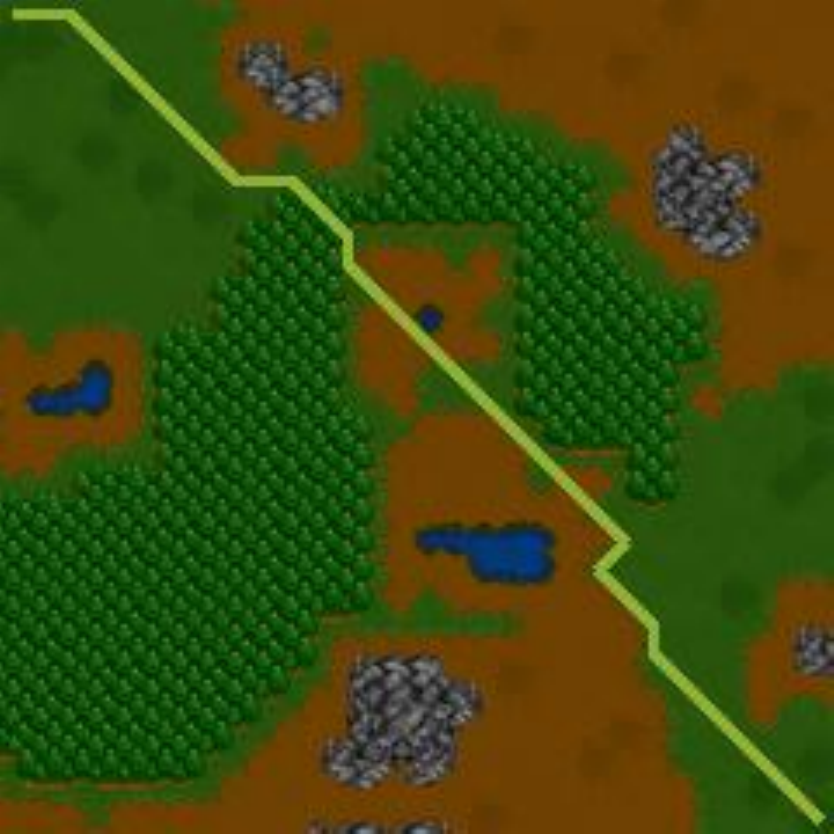} & \includegraphics[width=.25\linewidth]{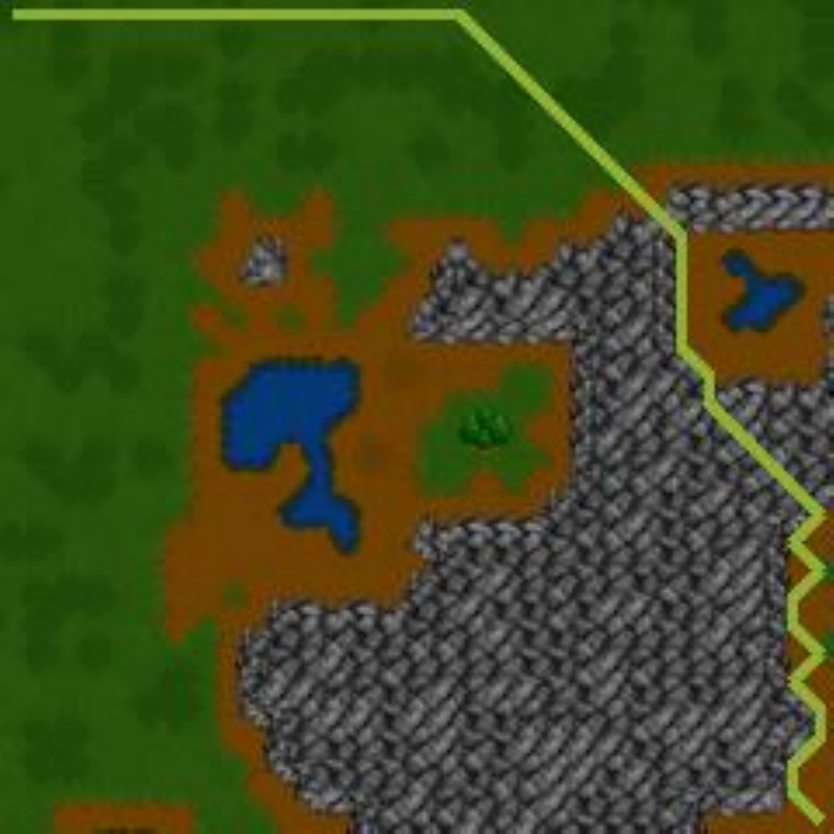} & \includegraphics[width=.25\linewidth]{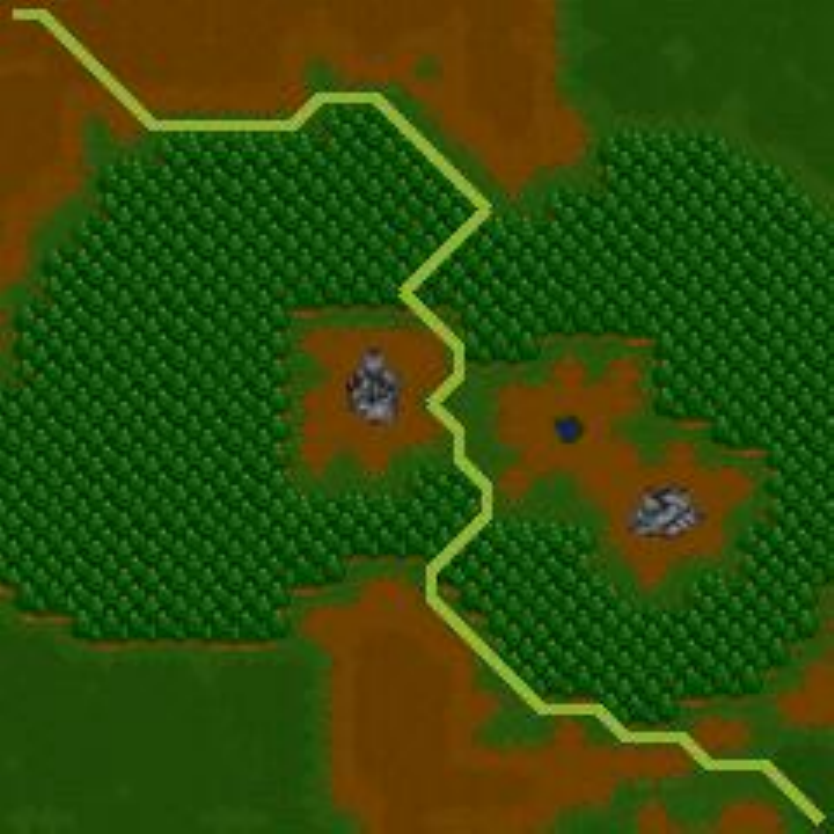}\\
\includegraphics[width=.25\linewidth]{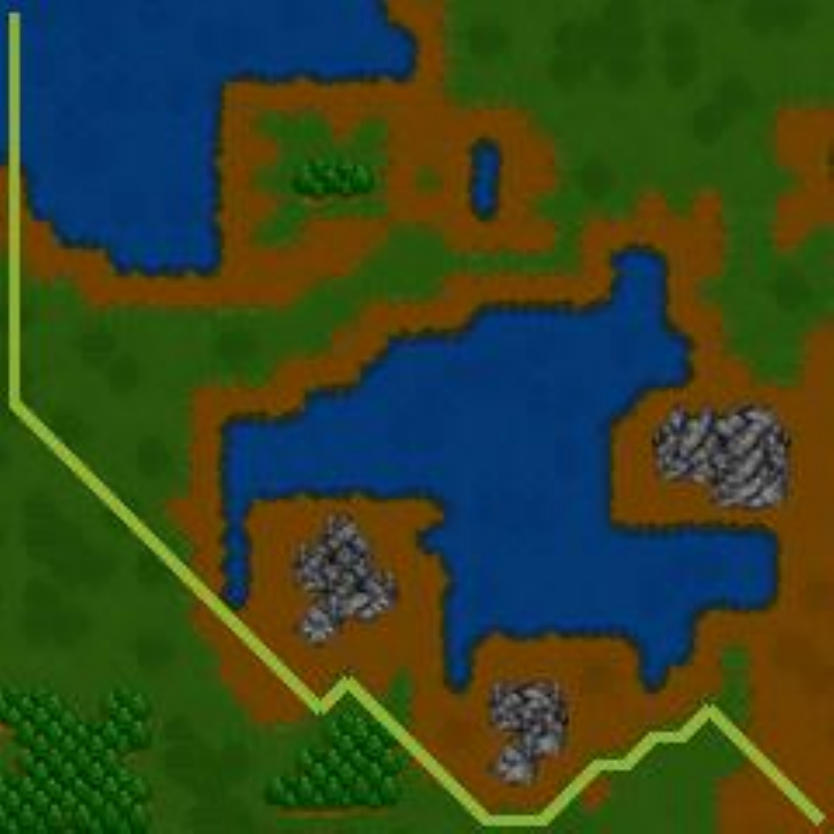} & \includegraphics[width=.25\linewidth]{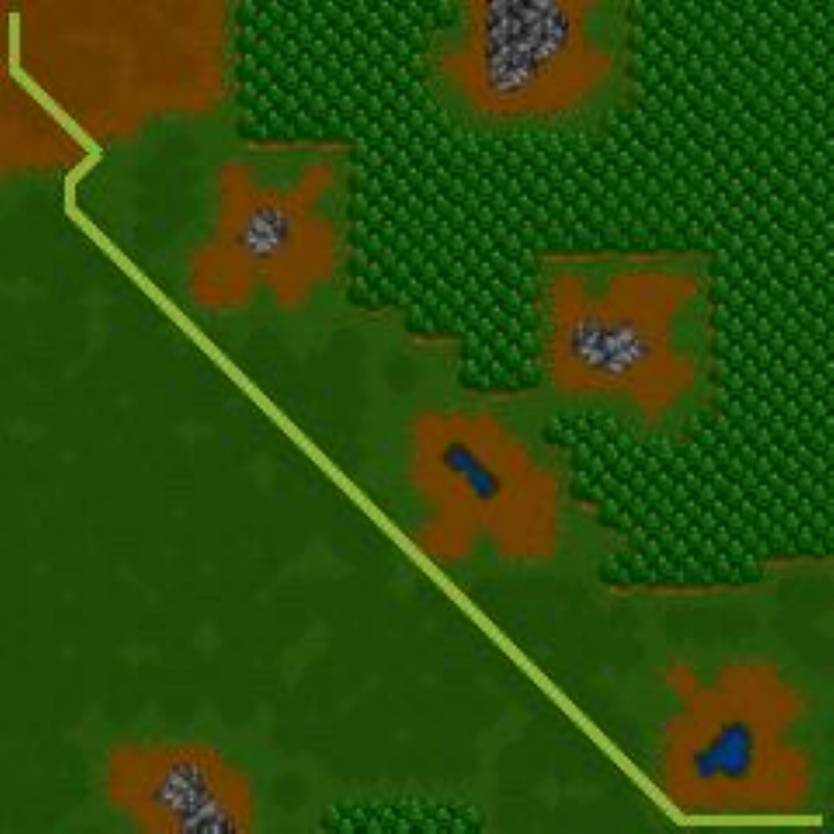} & \includegraphics[width=.25\linewidth]{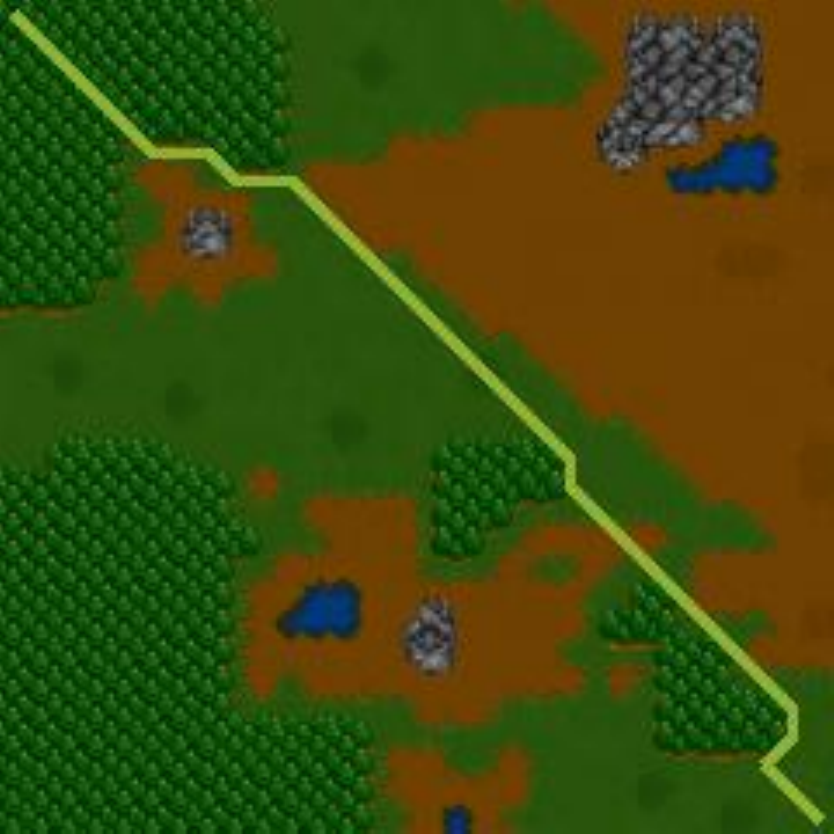} & \includegraphics[width=.25\linewidth]{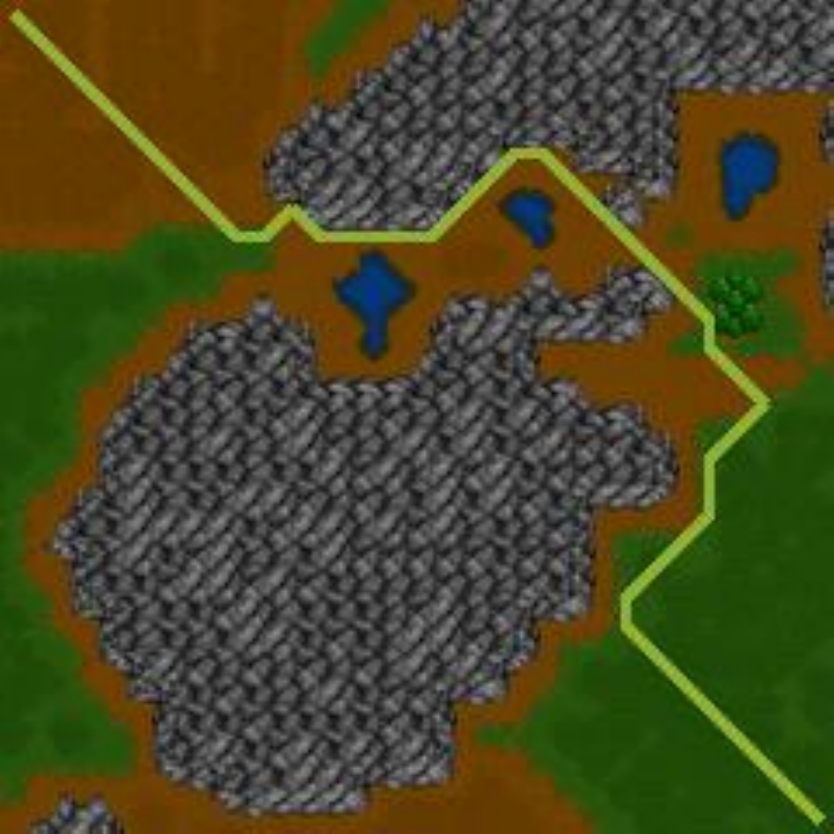}\\
\includegraphics[width=.25\linewidth]{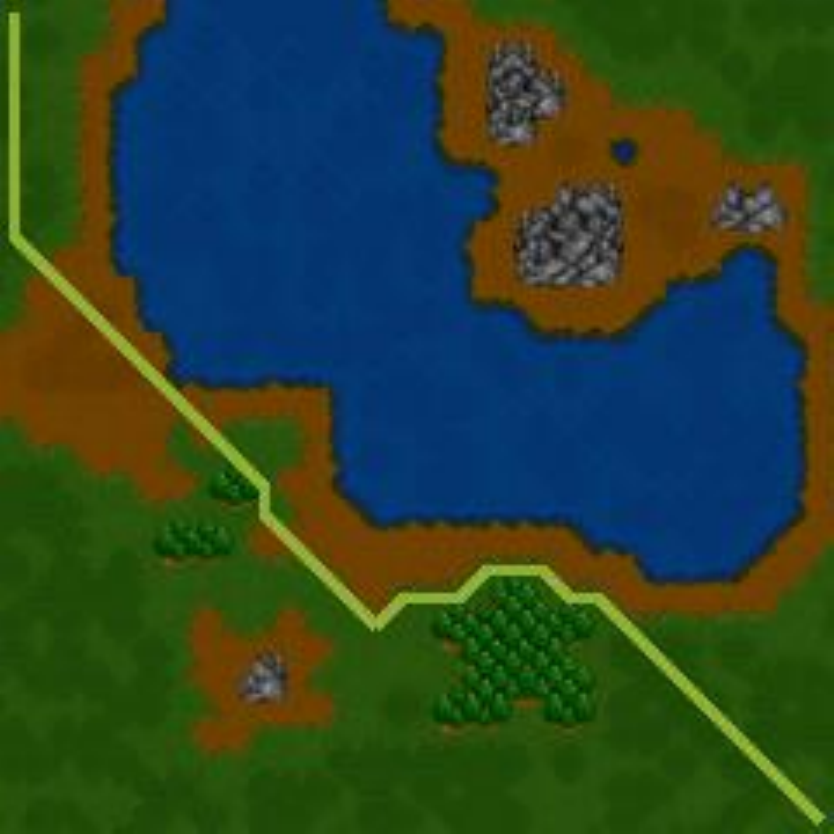} & \includegraphics[width=.25\linewidth]{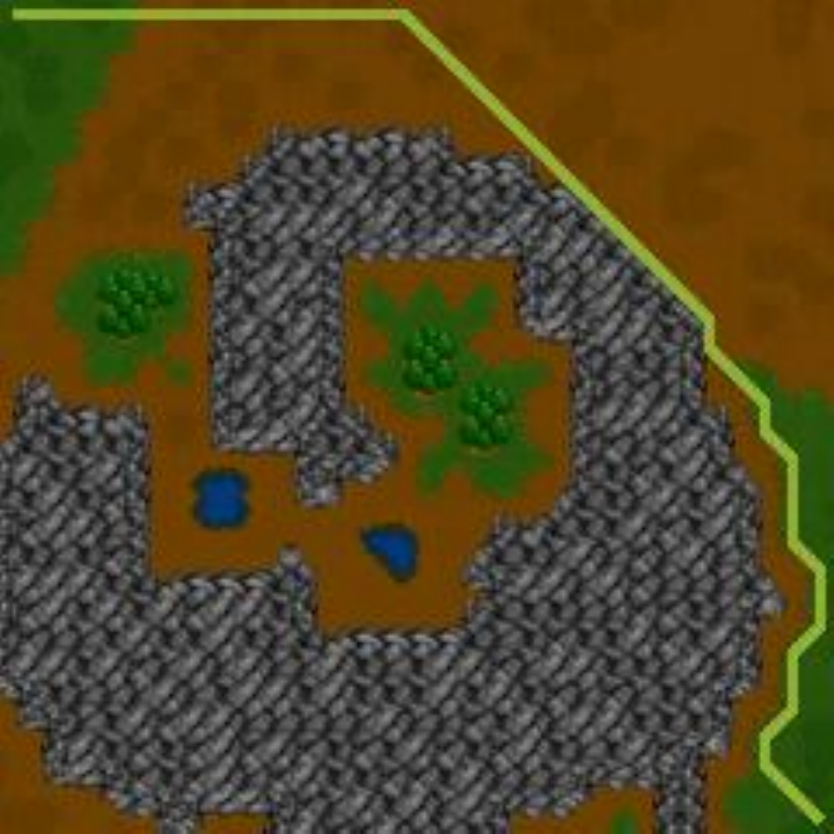} & \includegraphics[width=.25\linewidth]{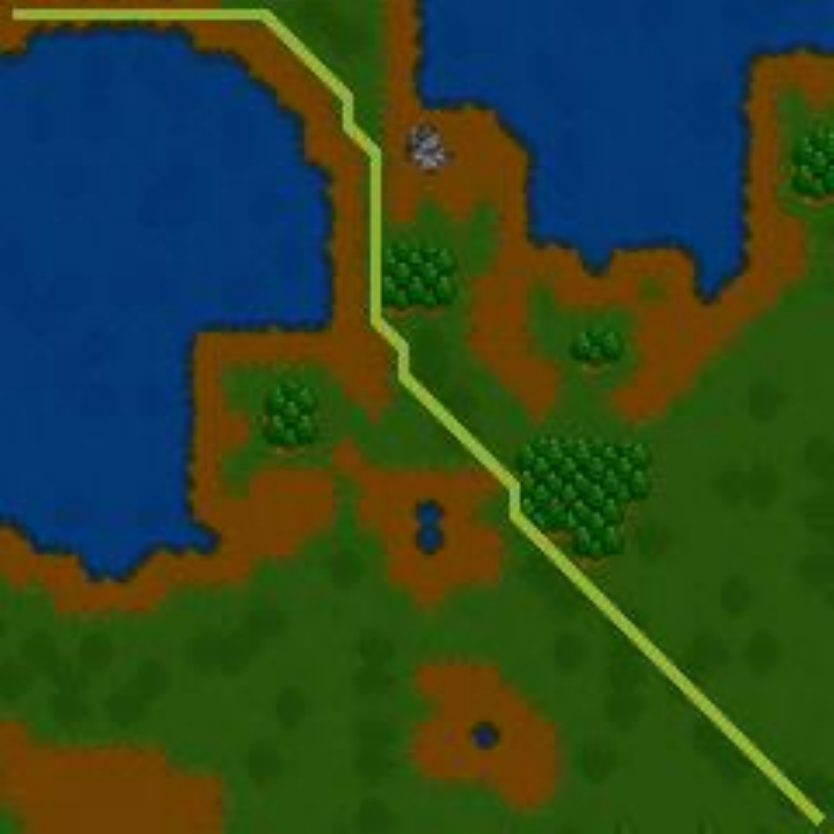} & \includegraphics[width=.25\linewidth]{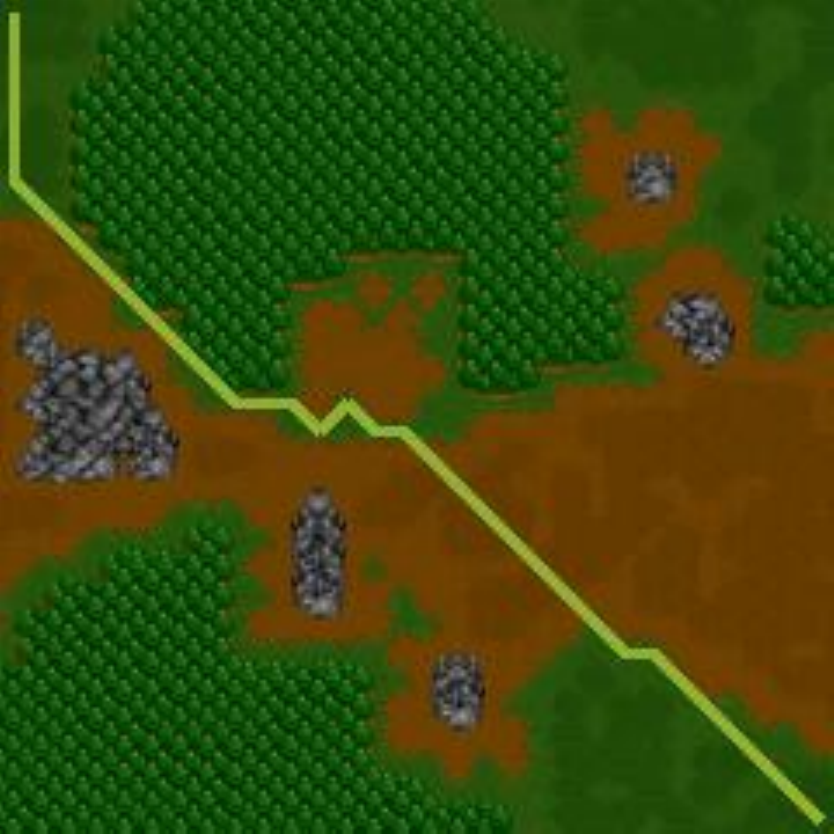}\\
\end{tabular}
}
\caption{\label{app:warcraft} Sample of Warcraft maps, and corresponding shortest paths from the upper left to the lower right corner of the map.} 
\end{table}

\subsection{Discrete Variational Auto-Encoder}

\begin{figure}[t]%
\centering
\subfigure{%
\label{fig:original-images}%
\includegraphics[width=0.3\textwidth]{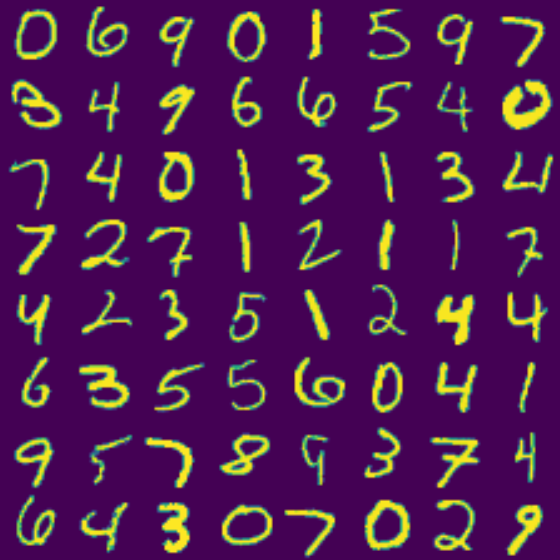}}%
\hspace{2.0mm}
\subfigure{%
\label{fig:reconstructed-images-gamma}%
\includegraphics[width=0.3\textwidth]{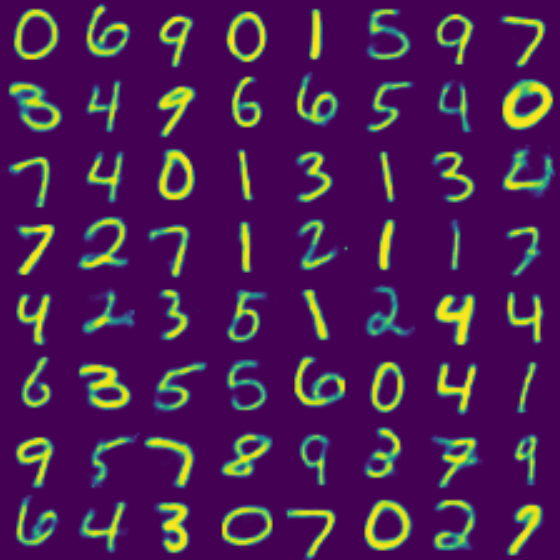}}%
\hspace{2.0mm}
\subfigure{%
\label{fig:reconstructed-images-lambda-1}%
\includegraphics[width=0.3\textwidth]{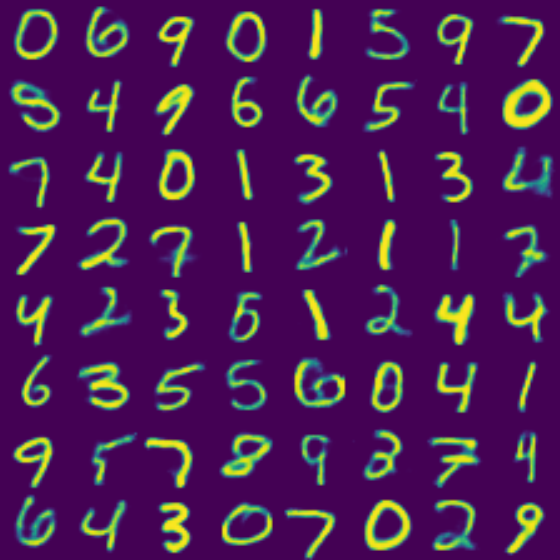}}%
\caption{\label{figure-dist-plots-mnist}Original MNIST digits from the test set and their reconstructions using the discrete $10$-subset VAE trained with Sum-of-Gamma perturbations for $\lambda=1$ (center) and $\lambda=10$ (right).}
\end{figure}

Experiments were run on a server with Intel(R) Xeon(R) CPU E5-2637 v4 @ 3.50GHz, 4 GeForce GTX 1080 Ti, and 128 GB RAM. 

The data set can be loaded in Tensorflow 2.x with \textit{tf.keras.datasets.mnist.load\_data()}. As in prior work, we use a  batch size of $100$ and train for $100$ epochs, plotting the test loss after each epoch. We use the standard Adam settings in Tensorflow 2.4.1 (learning rate=0.001, beta1=0.9, beta2=0.999, epsilon=1e-07, amsgrad=False), and no learning rate schedule. The MNIST dataset consists in black-and-white $28\times28$ pixels images of hand-written digits. The encoder network consists of an input layer with dimension $784$ (we flatten the images), a dense layer with dimension $512$ and ReLu activation, a dense layer with dimension $256$ and ReLu activation, and a dense layer with dimension $400$ ($20\times20$) which outputs the $\btheta$ and no non-linearity. 
The IMLESubsetkLayer takes $\btheta$ as input and outputs a discrete latent code of size $20\times20$. The decoder network, which takes this discrete latent code as input, consists of a dense layer with dimension $256$ and ReLu activation, a dense layer with dimension $512$ and ReLu activation, and finally a dense layer with dimension $784$ returning the logits for the output pixels. Sigmoids are applied to these logits and the binary cross-entropy loss is computed.  The training time (for the 100 epochs) was 21 minutes with the sum-of-Gamma perturbations and 18 minutes for the standard Gumbel perturbations. 

\begin{figure}[t!]
\centering
\includegraphics[width=1.0\textwidth]{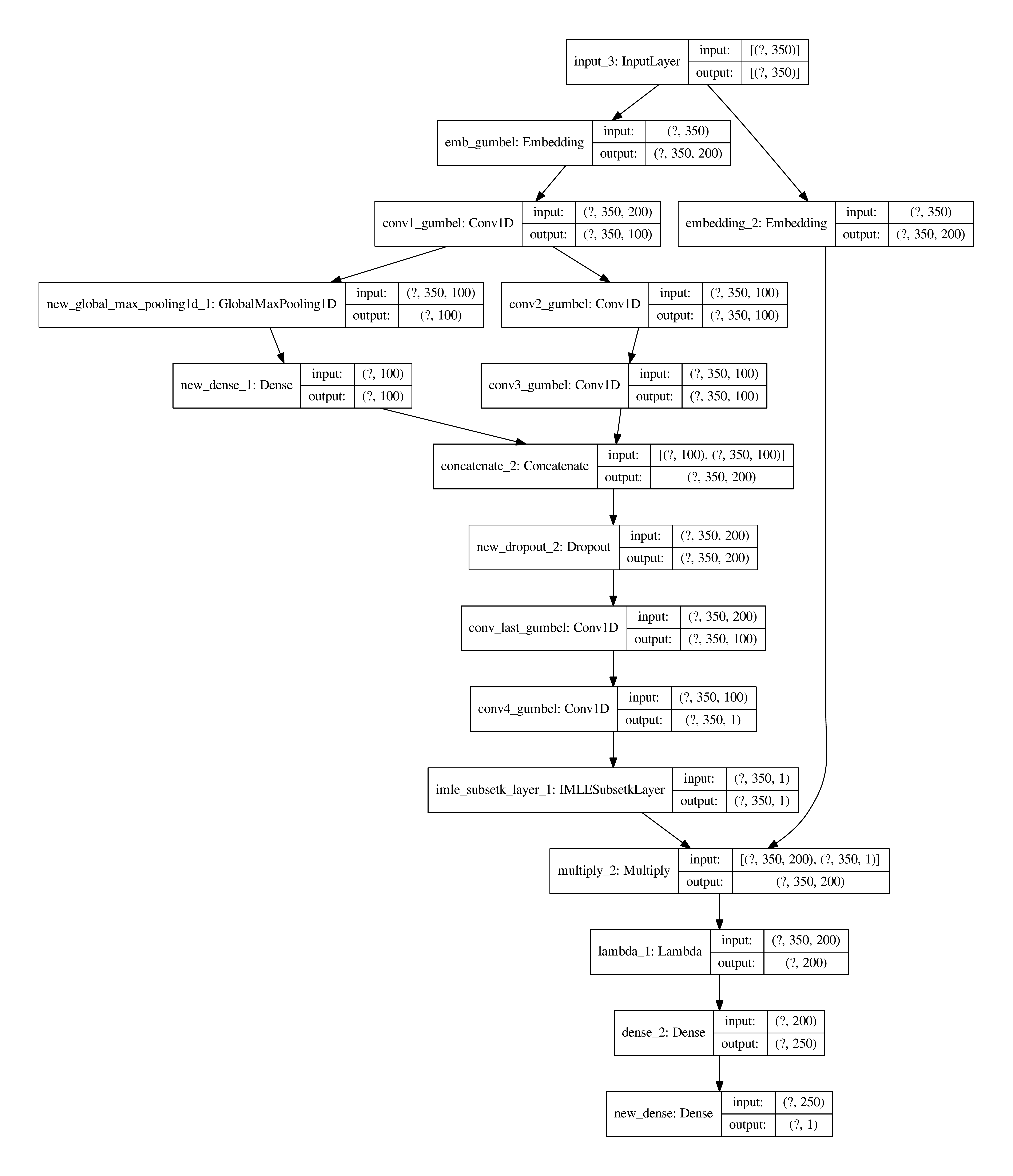}%
\caption{\label{fig-appendix-l2x}The neural network architecture for the learning to explain experiments. (Please zoom into the vector graphic for more details.) We use the standard architecture and settings from prior work~\cite{chen2018learning}. The maximum review length is 350 tokens. The dimensions of the token embeddings (of the embedding layers) are 200. All 1D convolutional layers have 250 filters with a kernel size of $3$. All dense layers have a dimension of 100. The dropout layer has a dropout rate of 0.2. The layer Multiply perform the multiplication between the token mask (output of I-MLE) and the embedding matrix. The Lambda layer computes the mean of the selected embedding vectors The last dense layer has a sigmoid activation. IMLESubsetkLayer is the layer implementing \textsc{I-Mle}. Code is available in the submission system.}
\end{figure}
\begin{figure}[t]
\centering
\includegraphics[width=0.9\columnwidth]{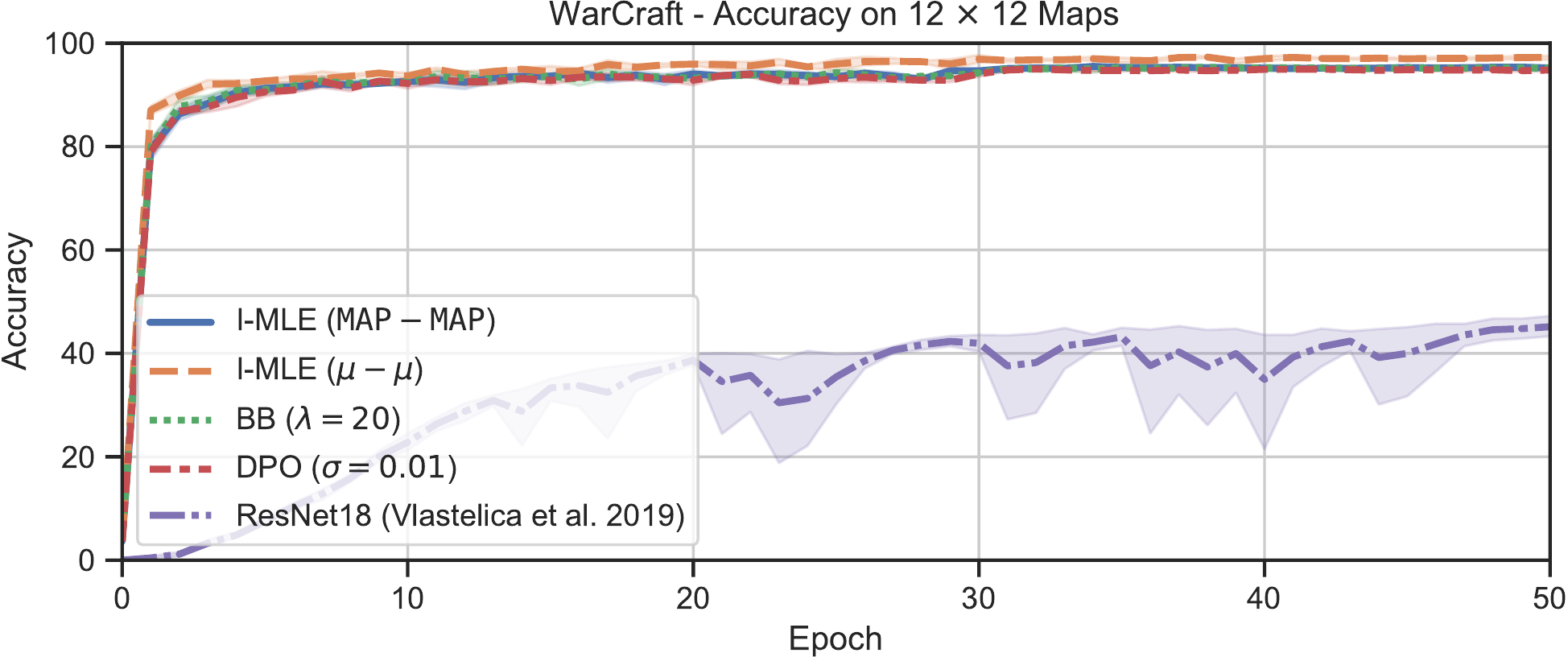}
\includegraphics[width=0.9\columnwidth]{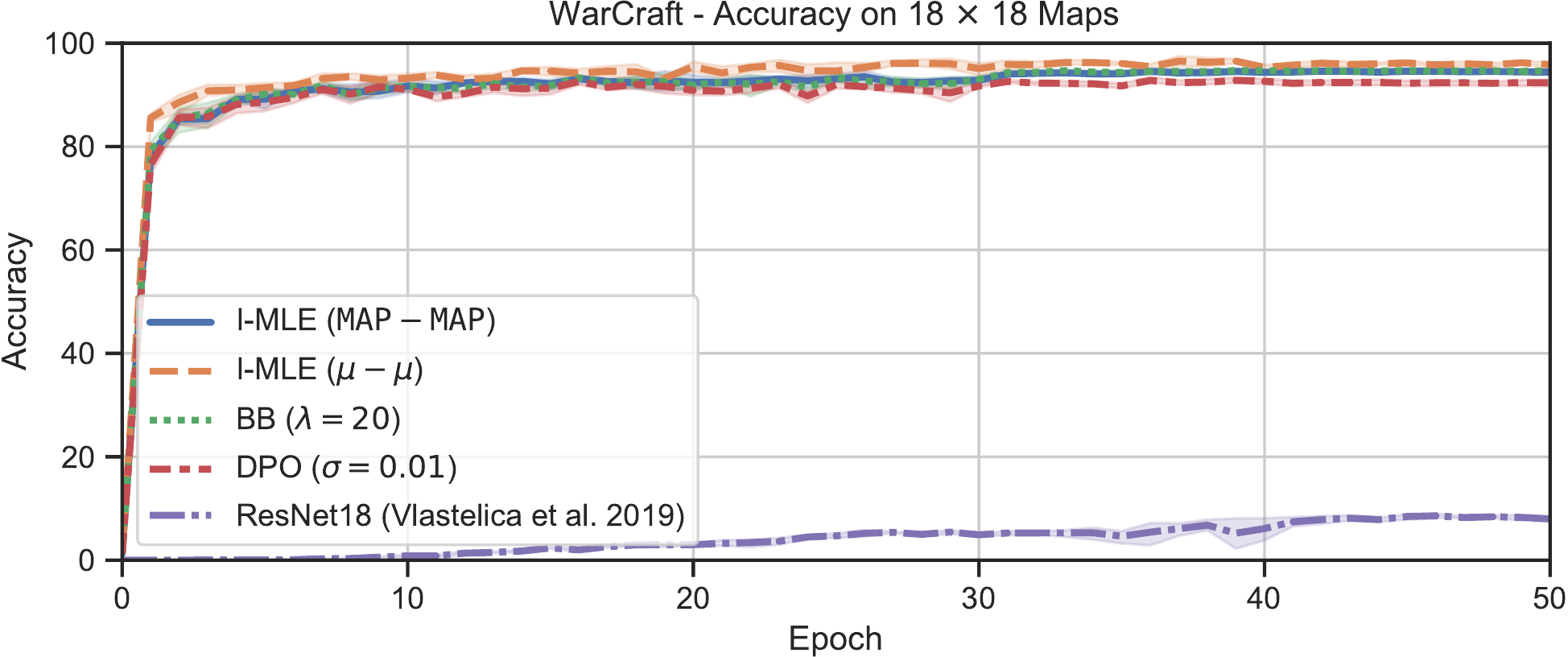}
\includegraphics[width=0.9\columnwidth]{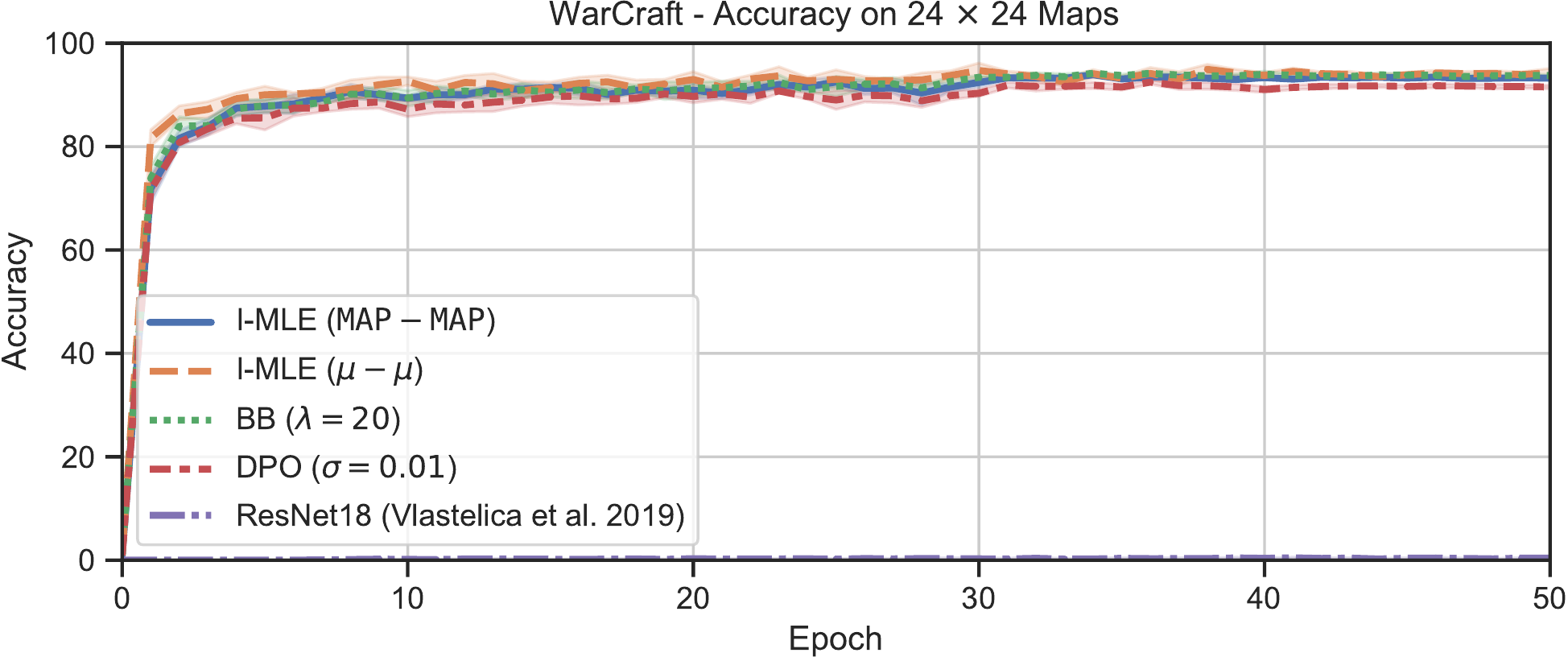}
\includegraphics[width=0.9\columnwidth]{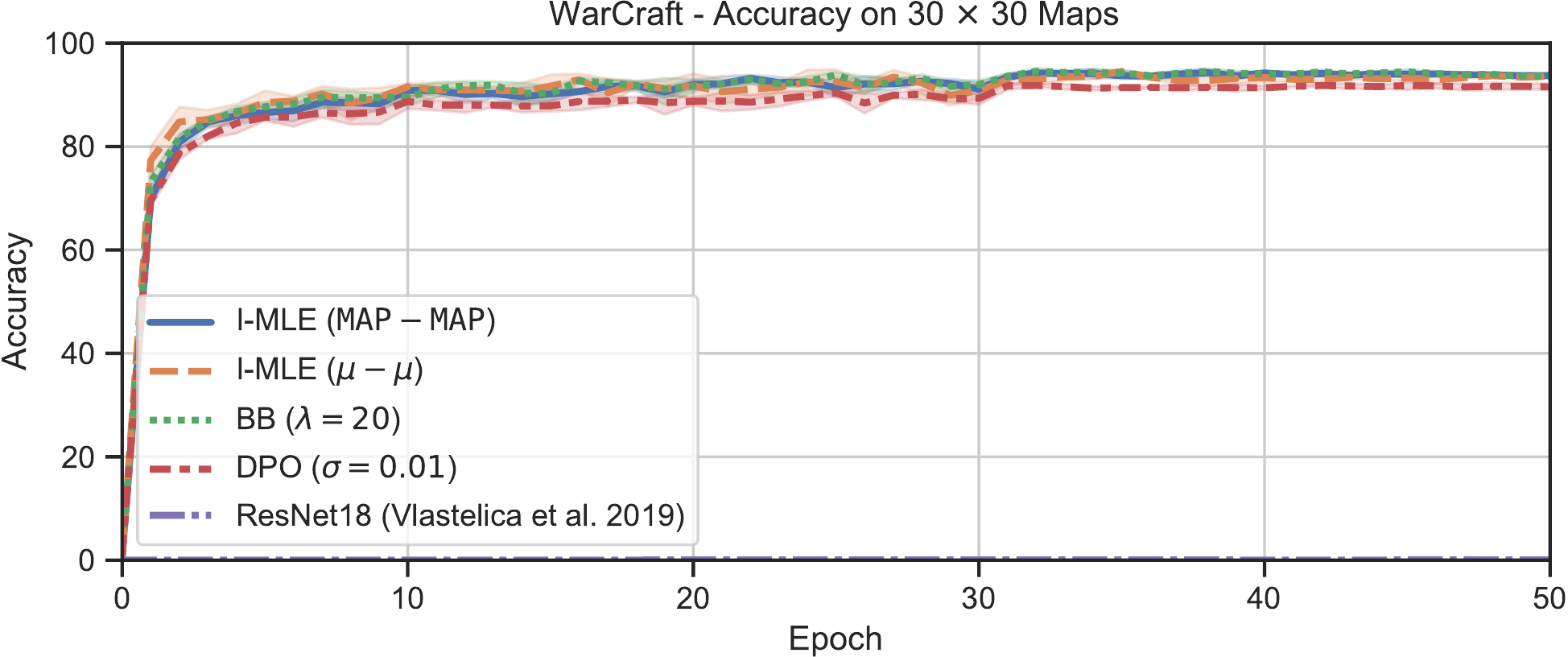}
\caption{Training dynamics for different models on $K \times K$ shortest path tasks on Warcraft maps, with $K \in \{ 12, 18, 24, 30 \}$.} \label{app:dynamics}
\end{figure}

\subsection{Differentiating through Combinatorial Solvers} \label{sec-axp:warcraft}

The experiments were run on a server with  Intel(R) Xeon(R) Silver 4208 CPU @ 2.10GHz CPUs, 4 NVIDIA Titan RTX GPUs, and 256 GB main memory.

\cref{app:warcraft} shows a set of $30 \times 30$ Warcraft maps, and the corresponding shortest paths from the upper left to the lower right corner of the map.
In these experiments, we follow the same experimental protocol of \citet{poganvcic2019differentiation}: optimisation was carried out via the Adam optimiser, with scheduled learning rate drops dividing the learning rate by 10 at epochs 30 and 40.
The initial learning rate was $5 \times 10^{-4}$, and the models were trained for 50 epochs using 70 as the batch size.
As in \citep{poganvcic2019differentiation}, the $K \times K$ weights matrix is produced by a subset of ResNet18~\citep{DBLP:conf/cvpr/HeZRS16}, 
whose weights are trained on the task.
For training BB, in all experimental results in \cref{section-experiment-blackbox} and \cref{app:warcraft}, the hyperparameter $\lambda$ was set to $\lambda = 20$.

\cref{app:dynamics} shows the training dynamics of different models, including the method proposed by \citet{poganvcic2019differentiation} (BB) with different choices of the $\lambda$ hyperparameter, the ResNet18 baseline proposed by \citet{poganvcic2019differentiation}, and \textsc{I-Mle}.

\begin{table}
\small
	\caption{Results for the Warcraft shortest path task using \textsc{I-Mle} with two target distributions, namely \cref{eq-pid-q} and \cref{eqn-target-q-optimal}. Reported is the accuracy, \ie percentage of paths with the optimal costs. Standard deviations are over five runs.} \label{tab:sp-appendix}
	\centering
	\resizebox{\columnwidth}{!}{
	\begin{tabular}{cccccc}
	\toprule
	$K$ & {$\bmu$-$\bmu$}, \cref{eqn-target-q-optimal} & {$\mathtt{M}$-$\mathtt{M}$}, \cref{eqn-target-q-optimal} & {$\lambda = 20$}, \cref{eq-pid-q} & {$\lambda = 20, \tau = 0.01$}, \cref{eq-pid-q} \\
	\midrule
	12 & ${\bf 97.2}\std{0.5}$ & $95.2\std{0.3}$ & $95.2\std{0.7}$ & $95.1\std{0.4}$ \\
	18 & ${\bf 95.8}\std{0.7}$ & $94.4\std{0.5}$ & $94.7\std{0.4}$ & $94.4\std{0.4}$ \\
	24 & ${\bf 94.3}\std{1.0}$ & $93.2\std{0.2}$ & $93.8\std{0.3}$ & $93.7\std{0.4}$ \\
	30 & $93.6\std{0.4}$ & $93.7\std{0.6}$ & $93.6\std{0.5}$ & ${\bf 93.8}\std{0.3}$ \\
	\bottomrule
	\end{tabular}
	}
\end{table}

Furthermore, we experimented with two different target distributions, namely \cref{eq-pid-q} and \cref{eqn-target-q-optimal}, where noise samples were drawn from a sum-of-Gamma distribution.
Results are summarised in \cref{tab:sp-appendix}.
In our experiments, the two target distributions yield very similar results for $\tau = 0.01$, and results tend to degrade for larger values of $\tau$.
This is to be expected since the target distribution in \cref{eqn-target-q-optimal} is meaningful in the context described in \cref{section-optimal-q-co}, where there are forms of explicit supervision over the discrete states.
Code and data for all the experiments described in this paper are available online, at \url{https://github.com/nec-research/tf-imle}.

}

\ifbool{printOld}{\include{old_stuff}}

\ifbool{extraspace}{\include{extraspace}}

\end{document}